\documentclass{article}

\usepackage{microtype}
\usepackage{graphicx}
\usepackage{subfigure}
\usepackage{booktabs} 
\usepackage{amsfonts}       
\usepackage{nicefrac}       
\usepackage{microtype}      
\usepackage{amssymb}
\usepackage{amsmath}
\usepackage{mathtools}
\usepackage{bbold}
\usepackage{subfigure}
\usepackage{wrapfig}
\usepackage{enumitem}

\usepackage{hyperref}

\usepackage{xcolor}

\usepackage[compact]{titlesec}

\usepackage{listings}
\usepackage{selectp}

\DeclareFixedFont{\ttb}{T1}{txtt}{bx}{n}{8} 
\DeclareFixedFont{\ttm}{T1}{txtt}{m}{n}{8}  

\usepackage{color}
\definecolor{deepblue}{rgb}{0,0,0.5}
\definecolor{deepred}{rgb}{0.6,0,0}
\definecolor{deepgreen}{rgb}{0,0.5,0}

\usepackage{listings}

\newcommand\pythonstyle{\lstset{
language=Python,
basicstyle=\ttm,
otherkeywords={self},             
keywordstyle=\ttb\color{deepblue},
emph={MyClass,__init__},          
emphstyle=\ttb\color{deepred},    
stringstyle=\color{deepgreen},
frame=tb,                         
showstringspaces=false            %
}}

\lstnewenvironment{python}[1][]
{
\pythonstyle
\lstset{#1}
}
{}

\newcommand\pythoninline[1]{{\pythonstyle\lstinline!#1!}}

\usepackage[preprint]{icml2021}

\icmltitlerunning{Discovery of Options via Meta-Learned Subgoals}

\begin{document}

\twocolumn[
\icmltitle{Discovery of Options via Meta-Learned Subgoals}



\icmlsetsymbol{equal}{*}

\begin{icmlauthorlist}
\icmlauthor{Vivek Veeriah}{umich}
\icmlauthor{Tom Zahavy}{dm}
\icmlauthor{Matteo Hessel}{dm}
\icmlauthor{Zhongwen Xu}{dm}
\icmlauthor{Junhyuk Oh}{dm}
\icmlauthor{Iurii Kemaev}{dm}
\icmlauthor{Hado van Hasselt}{dm}
\icmlauthor{David Silver}{dm}
\icmlauthor{Satinder Singh}{umich,dm}
\end{icmlauthorlist}

\icmlaffiliation{umich}{University of Michigan}
\icmlaffiliation{dm}{DeepMind}

\icmlcorrespondingauthor{Vivek Veeriah}{vveeriah@umich.edu}

\icmlkeywords{Machine Learning, ICML}

\vskip 0.3in
]



\printAffiliationsAndNotice{}  

\begin{abstract}
Temporal abstractions in the form of options have been shown to help reinforcement learning (RL) agents learn faster. However, despite prior work on this topic, the problem of discovering options through interaction with an environment remains a challenge. In this paper, we introduce a novel meta-gradient approach for discovering useful options in multi-task RL environments. Our approach is based on a manager-worker decomposition of the RL agent, in which a manager maximises rewards from the environment by learning a task-dependent policy over both a set of task-independent discovered-options and primitive actions. The option-reward and termination functions that define a subgoal for each option are parameterised as neural networks and trained via meta-gradients to maximise their usefulness. Empirical analysis on gridworld and DeepMind Lab tasks show that: (1) our approach can discover meaningful and diverse temporally-extended options in multi-task RL domains, (2) the discovered options are frequently used by the agent while learning to solve the training tasks, and (3) that the discovered options help a randomly initialised manager learn faster in completely new tasks. 
\end{abstract}

\section{Introduction}
\label{sec:introduction}

Reinforcement learning (RL) problems involve 
learning about temporally-extended 
actions at multiple time scales. In RL, the \emph{options} framework~\citep{sutton1999between} provides a well-defined formalisation for the notion of temporally-extended actions. Options that achieve specific subgoals can be useful in reinforcement learning (RL) in at least two ways: in model-based RL, they provide faster rates of convergence through longer-backups of value functions within planning updates~\citep{Silver2012CompositionalPU,Mann14,Brunskill2014}, while in model-free RL, temporally-extended actions commit agents to intentional multi-step behaviours, which can translate into better exploration~\citep{machado2016learning,Nachum2019, Osband2019}. 

We consider a scenario where an agent learns to solve a distribution over tasks.
In such cases having carefully designed temporal abstractions can greatly reduce the overall sample complexity of learning for an RL agent that is trying to master those tasks. The agent can produce faster learning mainly by reusing those abstractions across multiple tasks~\citep{sutton1999between,solway2014optimal}. Many recent approaches have 
empirically validated this by demonstrating that hand-designed temporal abstractions can often lead to improved learning performance on a variety of challenging multi-task RL domains~\citep{Imazeki2003,kulkarni2016hierarchical,nachum2018data,riedmiller2018learning}. However, if the agent has abstractions that are not useful to a downstream task, it can significantly hurt the performance by making exploration harder~\cite{jong2008utility}, emphasising the challenge involved in carefully hand-designing abstractions that are useful \textit{in general}, across many domains and tasks. Thus, the automated discovery of temporal abstractions from experience without extensive domain-specific knowledge remains an important open problem for reducing sample complexity in RL. 


The \textit{main contribution} of this work is our hierarchical agent architecture and an associated meta-gradient algorithm to discover temporally-extended actions in the form of options that can be reused across many tasks. Previously, meta-gradients have been successfully used for learning hyperparameters~\citep{xu2018meta,zahavy2020self}, intrinsic rewards~\citep{zheng2018learning,zheng2019can,rajendran2019should}, and auxiliary tasks~\citep{veeriah2019discovery}. Our work is the first to demonstrate that they can successfully learn rich parameterisations of temporal abstractions. Our starting point is the following hypothesis: If we could \emph{discover} temporal abstractions useful across \emph{many training tasks}, they would capture regularities across those task environments and have a higher likelihood of being useful and reusable in new, \emph{previously unseen tasks.}

To discover temporal abstractions useful across tasks, our system flexibly defines \textit{task-independent} subgoals for options through separate discovered {\it rewards} and {\it terminations}, different for each option. We employ meta-gradients to discover the parameters of such option-rewards and terminations based on their utility across the many training tasks, so that the set of induced options is useful to a hierarchical agent trying to master all training tasks. The meta-gradient approach operates by evaluating a \textit{change} in the options, caused via changes to the option-rewards and terminations, w.r.to the hierarchical agent's performance on samples drawn from many tasks; then computes and uses the gradients from this evaluation to \textit{discover} the option-rewards and terminations. This differs significantly from the previous multi-task option discovery approaches~\citep{bacon2017option, frans2017meta}, where all options directly optimise the same (main task) reward, which may be insufficient to discover reusable, task-independent options.

We evaluate the proposed approach empirically in two multi-task RL settings based on an illustrative gridworld and on 3-dimensional first-person task suites from DeepMind Lab~\citep{Beattie2016}.
For each of these, we perform three types of analysis:
(1) we qualitatively demonstrate that our approach indeed produces meaningful temporally-extended options; (2) we quantitatively show that our discovered options are extensively used by the agent while learning to solve training tasks; (3) we show that our discovered options support faster learning, i.e., transfer better, in test tasks, compared to options discovered by two strong hierarchical baselines (MLSH \& Option-Critic).

\section{MODAC: Meta-gradients for Option Discovery using Actor-Critic}  

\noindent{\bf Why Meta-gradients?} As described in the previous section, the motivation behind our work is to discover temporal abstractions in the form of options that are {\it generally} useful across many training tasks; and to leverage them to allow effective transfer of acquired skills to new tasks. This is based on our hypothesis that options, if useful across many tasks, capture intrinsic properties about those tasks that could 
lead to better transfer to unseen tasks. For useful options to be discovered, an option-based agent needs to evaluate whether \emph{a change} in a given option is useful. Essentially, the agent needs to compute the gradient of future performance with respect to the parameterisation of each option, while the option-based behaviour is itself adapted by the conventional RL gradients; this requires the computation of a gradient through a gradient, and thus meta-gradients is used as the mechanism for driving option discovery. We make this idea concrete in our learning agent MODAC (which stands for {\bf M}eta-gradients for {\bf O}ption {\bf D}iscovery using {\bf A}ctor-{\bf C}ritic).

\noindent{\bf Background on Options:}\label{sec:background}
Options are closed-loop behaviours over extended periods of time. An option is formally defined by specifying an \textit{initiation set} (states where the option may be invoked), an \textit{option-policy} (maps states to actions), and a \textit{termination} function (maps states to probability of terminating execution of the option). The option-policy may be defined implicitly as the policy that maximises an \textit{option-reward} function describing the subgoal (or \textit{intention}) for that option.
\begin{figure*}[t]
    \centering
    \includegraphics[width=0.75\linewidth]{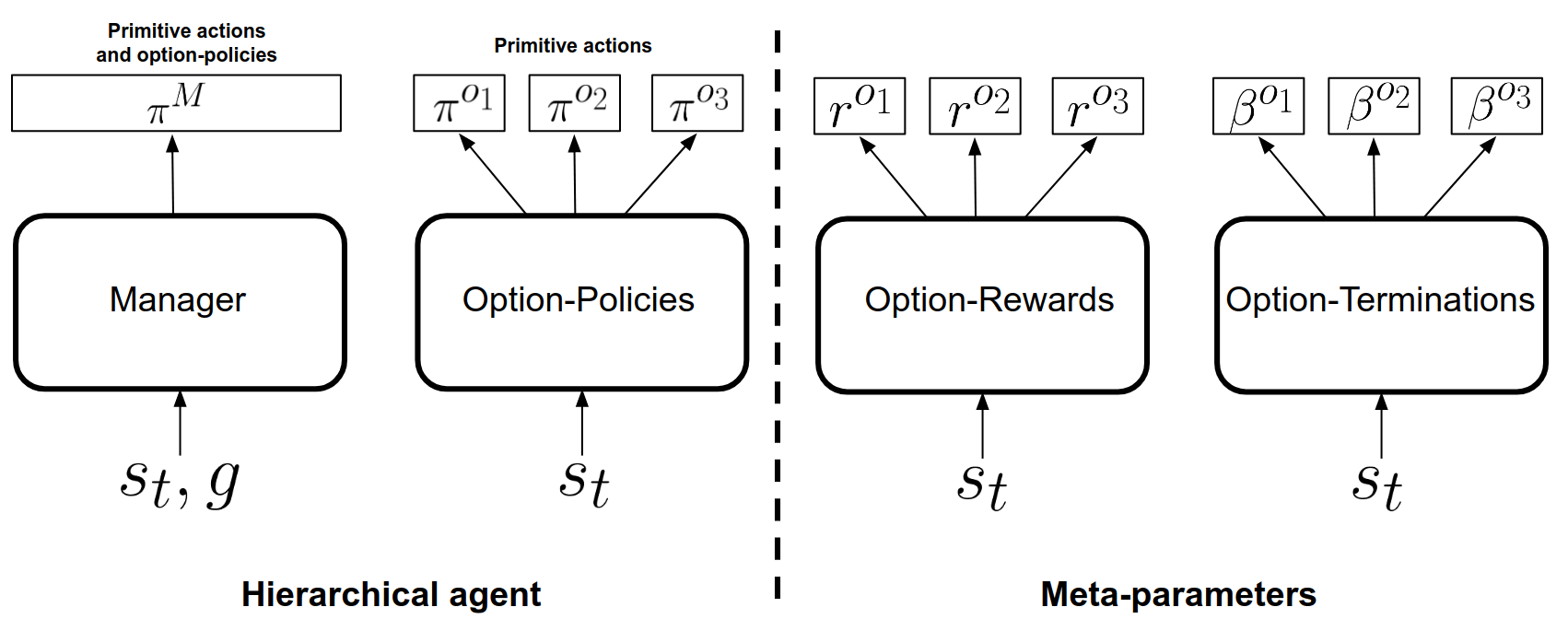}
    \caption{MODAC: features four networks, namely: the manager-policy, option-policies, option-rewards, and option-terminations. Manager and option-policies are trained via (direct) RL. Option-rewards and terminations define the semantics for the option-policies and are discovered (indirectly) using meta-gradient RL. More details in Sec.~\ref{sec:architecture}.}
    \label{fig:hrl_architecture}
\end{figure*}
\subsection{Agent Architecture}\label{sec:architecture}
We base our agent's architecture on the standard hierarchical agent from \citet{sutton1999between}, where a manager chooses among both primitive actions\footnote{Primitive actions are a special case of options that terminate after one step. See Sec. 3.1, pg. 194 of ~\citet{sutton1999between}.} and temporally-extended options; and extend it to the multi-task setup. The agent (shown in Fig.~\ref{fig:hrl_architecture}) follows a {\it call-and-return} option execution-model and consists of the following four modules:

A {\bf Manager} network parameterised by $\theta^M$ implements task-conditional policy $\pi^M$. It maps a sequence of observations (henceforth the state $s_t \in \mathcal{S}$) and a task encoding $g \in \mathcal{G}$ to the union of the set of option-policies $\mathcal{O}$ and the set of primitive actions $\mathcal{A}$, i.e., $\pi^M: \mathcal{S} \times \mathcal{G} \rightarrow \{\mathcal{O}, \mathcal{A} \} $. We denote a sample from $ \pi^M $ as $ o \sim \pi^M(s, g)$. The Manager is trained to maximise the task reward.

The {\bf Option-policy} network represents policies for each discovered option in our hierarchical agent. It takes state as input and has multiple output head that each produce a distribution over primitive actions: the $i^{th}$ head's output represents the option-policy $\pi^{o_i}: \mathcal{S} \rightarrow \mathcal{A}$).  $\theta^{o_i}$ denotes the parameters of $\pi^{o_i}$. Each option-policy learns to maximise the corresponding option-rewards produced from the option-reward network described next.
    

The {\bf Option-reward} network defines the subgoals/intentions for the option-policies. It takes state as input and outputs a scalar reward, one per option-policy, for each primitive action $a \in \mathcal{A}$ (the $i^{th}$ option-reward head's output is $r^{o_i}: \mathcal{S} \rightarrow \mathbb{R}^{|\mathcal{A}|}$). $\eta^{r^{o_i}}$ denotes the parameters of $r^{o_i}$. 
    
{\bf Option-termination} network defines the termination function for the option-policies and has an identical structure to the option-reward network: it also takes state as input and outputs a scalar value $\beta^{o_i}: \mathcal{S} \rightarrow \{0, 1\}$ for each  option-policy, interpreted as the probability of termination for the associated option. $\eta^{\beta^{o_i}}$ denotes the parameters of $\beta^{o_i}$.

Each module in the agent closely follows the options framework: the manager corresponds to the \emph{policy-over-options} from the options framework as it learns a mapping from states to options. Each option is defined by its associated option-policy and termination defined by the respective networks (initiation set for each option is softly induced by the manager's policy). Option-reward along with their termination defines an option's subgoal.

Note, among the four modules, only the manager gets the task goal as input while those corresponding to options (i.e., option-policy, reward and termination) do not. 
This architecturally enforces our objective of discovering task-independent options that are useful across multiple training tasks, and also supports their transfer to new, previously unseen tasks. We emphasise that it is possible for MODAC to solve each task optimally as the manager gets task as input and can select primitive actions if need be.  Also, this form of hiding task goals from the temporal abstractions has been shown to be effective for transfer in several multi-task hierarchical RL approaches~\citep{dayan1993feudal,heess2016learning,jaderberg2016reinforcement,nachum2018data}. Separately, note that our assumption of access to task encodings is standard in many multi-task RL~\citep{Beattie2016,plappert2018multi} and robotics domains~\citep{kolve2017ai2,deitke2020robothor}.

We motivate the choice of allowing the manager to pick between option-policies and primitive actions as opposed to option-policies alone with the following example: consider the case where options take you to the doorways in a building with rooms. Suppose the goal was to go to the middle of some room. Now there is no policy that maps states to options that can achieve that middle-of-the-room goal. But if we allow the manager to choose both options and primitive actions, then it could get to the doorway of the room by traveling from doorway to doorway and then pick primitive actions to get to the middle of the target room. This is a far more flexible use of options. Indeed, this is the way options were presented originally~\cite{sutton1999between}.

In the next section, we present a discovery algorithm based on meta-gradients that is used to train each network.

\subsection{MODAC's Discovery Algorithm}\label{sec:general_discovery_options}
\noindent{\bf Algorithm Overview:} We extend the general meta-gradient algorithm~\citep{xu2018meta} to discover options within our hierarchical architecture. 
It consists of two nested loops: an \emph{inner-loop} that updates the manager to maximise the discounted sum of task rewards and that updates the option-policies to maximise the discounted sum of its corresponding option-rewards (with discounts provided by option-terminations); and an \emph{outer-loop} that evaluates the \emph{updated} manager and option-policies on new transitions produced by the agent and then updates the option-rewards and option-terminations by computing meta-gradients by back-propagating through the inner-loop updates.

In the inner-loop, the parameters of the manager $\theta^M$ and option-policy $\{\theta^{ o_i }\}$ are updated with transitions produced by the agent on a random sample of tasks. In the outer-loop, the updated manager and option-policies are evaluated on new transitions drawn from another random sample of tasks, and then the meta-parameters of the option-reward $\{\eta^{r^{o_i}}\}$ and option-termination $\{\eta^{\beta^{o_i}}\}$ networks are updated by back-propagating through the inner-loop updates. 

Below, we instantiate the algorithm for an actor-critic architecture, leaving adaptation to other RL updates to future work. A detailed derivation is presented in the Appendix.



{\it Inner-loop:} Consider a $n$-step trajectory $ \{ s_{t}, a_{t}, r_{t+1}, r^{o}_{t+1}, \beta^{o}_{t+1}, \pi^{o}, g\}_{t=t_0}^{t_0 + n} $ generated by following an option-policy $\pi^{o}$ until its termination ($\beta^o_{t_0+n} = 1$), where $\pi^{o}$ was sampled from the manager's policy $\pi^{M}$ at $t=t_0$ while interacting with task $g$.
For such a trajectory, the inner-loop updates to the option policy and manager parameters are: 
\begin{align}
    \theta^{o} \gets &\theta^{o} + \alpha \big( G_{t}^{o} - v^{o}  (s_{t}) \big) \cdot \nonumber\\
    &\nabla_{\theta^{o}} \big[\log\pi^{o}(a_{t} | s_{t}) - \kappa^{o} v^{o}(s_t) \big] \label{eqn:option_policy_inner_update}\\
    \theta^{M} \gets &\theta^{M} + \alpha \big( G^{M}_{t_0} - v^{M}(s_{t_0}, g) \big) \cdot \nonumber\\
    &\nabla_{\theta^{M}}\big[ \log\pi^{M}(o | s_{t_0}, g) - \kappa^{M} v^{M}(s_{t_0}, g) \big]\label{eqn:manager_policy_inner_update}
\end{align}
where $\kappa^M, \kappa^{o}$ weight the value updates relative to the policy updates, and $G_{t}^{o}$ and $G_{t}^{M}$ are $n$-step returns for the option-policy and manager:
\begin{align}
    G^{o}_{t} &= \sum_{j=1}^{n}(1 - \beta^o_{t+j})^j r^{o}_{t+j} + (1 - \beta^o_{t+n})^{n+1}v^{o}(s_{t+n})\label{eqn:Go} \\
    G^{M}_{t} &= \sum_{j=1}^{n}\gamma^j r_{t+j} - \gamma^n c + \gamma^{n+1}v^{M}(s_{t+n})\label{eqn:Gm} 
\end{align}
where $c$ is a {\it switching cost} added, on option terminations, to the per-step rewards used in manager's update. \textit{The switching cost hyperparameter encourages the manager to pick options that are temporally-extended} (therefore aiding their discovery).

{\it Outer-loop:} In the outer-loop update to the option-reward and option-termination meta-parameters, we use a different trajectory $ \{ s_{t}, a_{t}, r_{t+1}, \pi^{o_t}, g \}_{t=t_0 + n + 1}^{t_0 + n + m} $ generated by interacting with the environment using the latest inner-loop parameters $\theta^{o_i},\ i = 1 \dots K$. Since this trajectory is used to evaluate the change made to the manager and option-policy parameters, we refer to it as a \emph{validation} trajectory. The task $g$ may be different from the one used in inner-loop update. On this validation trajectory, we compute the meta-update to the option-reward and option-termination, back-propagating through the inner-loop updates\footnote{The option-policy parameters are a function of option-reward and termination parameters, and thus, allow for computing the meta-gradients.}:
\begin{align}
    \forall i, \ \ \eta^{r^{o_i}} \gets \eta^{r^{o_i}} +& \alpha_\eta \big( G^{M}_{t} - v^{M}(s_{t}, g) \big).\nonumber\\
    &\nabla_{\eta^{r^{o_i}}}\log\pi^{o_{i}}(a_{t} | s_{t})\nonumber\\ 
    \forall i, \ \ \eta^{\beta^{o_i}} \gets \eta^{\beta^{o_i}} +& \alpha_\eta \big( G^{M}_{t} - v^{M}(s_{t}, g) \big).\nonumber\\
    &\nabla_{\eta^{\beta^{o_i}}}\log\pi^{o_{i}}(a_{t} | s_{t})\label{eqn:option_reward_termination_outer_update}
\end{align}

{\it Pseudocode:} The algorithm for training a MODAC agent is summarised in Alg.~\ref{alg:disc_options_algorithm}. MODAC utilises the inner-loop updates (see Eqns.\ref{eqn:option_policy_inner_update},~\ref{eqn:manager_policy_inner_update}) to train the parameters of the option-policies and manager, and in the outer-loop, discovers the option-reward and termination parameters via meta-gradients obtained using the updated option-policy parameters (see Eqn.~\ref{eqn:option_reward_termination_outer_update}). 
\textit{Furthermore, meta-gradients are efficiently computed through backward-mode autodifferentiation, thus making its computational complexity similar to that of the forward computation}~\citep{griewank2008evaluating}. 

\begin{algorithm}
  \caption{Meta-gradient algorithm for option discovery}\label{alg:disc_options_algorithm}
  \begin{algorithmic}
      \STATE Initialise parameters $ \theta^{M}$, $(\theta^{o_{i}}, \eta^{r^{o_{i}}}, \eta^{\beta^{o_{i}}})\ \forall i = 1 \dots K$  
      \STATE Sample task $g \sim \mathcal{G}$, state $s \sim S_{0}(g)$,  option-policy $o \sim \pi^{M}(s, g)$
    \REPEAT {} 
        \FOR{$ l = 1, 2, $ {\bfseries to} $L$} 
            \STATE Re-sample ($ s $, $ g $, $ o $) when starting a new episode
            \IF {$ \beta^{o}(s) == 1 $}
                \STATE $ o \sim \pi^{M}(s, g) $ 
            \ENDIF
            \STATE Obtain transition using option-policy $s \sim E(s, \pi^{o}(s))$ \hfill \# {\bf E denotes the environment}
            \STATE \# {\bf Inner-loop update}
            \STATE Update option-policy $\theta^o$ with Eqn.~\ref{eqn:option_policy_inner_update}
            \STATE Update manager $\theta^M$ on states where it samples $o$ with Eqn.~\ref{eqn:manager_policy_inner_update}
        \ENDFOR
        \STATE Obtain transitions from another task using the updated manager $\theta^M$ and option-policy $\{\theta^{o_i}\}$ parameters
        \STATE \# {\bf Outer-loop update}
        \STATE Update option-reward $\eta^{r^{o_i}}$ and termination $\eta^{\beta^{o_i}}$ with Eqn.~\ref{eqn:option_reward_termination_outer_update}   \ \  $\forall i = 1 \dots K$
    \UNTIL maximum number of timesteps
  \end{algorithmic}
\end{algorithm}

\subsection{Differences Between MODAC and Prior Option-Discovery Methods}
Previous option-discovery works that have been explored in a multi-task setup, (c.f. MLSH, Option-Critic, Coagent Networks) usually train all the options to optimise the task reward for the current task generating the data, though gradients are accumulated across tasks.  
In contrast, in our agent, each option-policy optimises a \textit{different} objective, parameterised by the corresponding task-independent option-reward and termination that are discovered to be \textit{directly} useful across many tasks. The MODAC architecture and the meta-gradient learning updates discover task-independent, general-purpose, disentangled options that can help the agent not just to achieve higher rewards during training, but also to speed up learning in new tasks.

\section{Empirical Results}
We empirically evaluated MODAC using a mix of gridworld and navigation tasks from DeepMind Lab. Our main numeric metric for the quality of the options discovered by our agent was their usefulness for transfer. In addition to this, we evaluated their quality in various ways, including whether they were temporally extended and whether they are diverse in their behaviour. Our experiments all included a {\it training} phase followed by a {\it testing} phase.

{\it Training:} For every episode, a training task was selected randomly from a set of training tasks that were distinct from the test tasks. In this phase, MODAC jointly learned the manager, option-policies, option-rewards, and option-terminations with transitions from the episode. As noted earlier, in this phase a penalty in the form of switching cost is added to the task rewards that is used in the manager's learning update (see Eqn.~\ref{eqn:manager_policy_inner_update},~\ref{eqn:Gm}). The switching cost encourages the manager to pick and thus help discover temporally-extended options by penalising the manager for picking primitive actions too often.

{\it Testing:} We froze option-policies and option-termination functions after training, and transferred them to a new manager with randomly initialised parameters; then evaluated how fast the manager could learn on different unseen test tasks. Performance on each test task was evaluated independently, re-initialising the manager every time, and the results presented are averages across test tasks. If the manager finds the transferred options to be useful for maximising rewards from test tasks, then the manager would naturally pick those options without need of the switching cost penalty. Thus the switching cost is not used in this phase.

{\it Baselines:} We compared the performance of MODAC at test time with that of \textit{Flat}, a non-hierarchical actor-critic agent, and also to the performance of hierarchical agents that used options discovered using MLSH~\citep{frans2017meta} and a multi-task extension of the Option-Critic~\citep{harb2018waiting}, as these hierarchical approaches also use multi-task performance on rewards to drive their discovery process. We use an identical train/test setup for all the hierarchical agents. Details on all the agents, their hyperparameter choices and other implementation details are in the Appendix. 

\subsection{An Illustrative Gridworld}
\label{sec:gridoworld_results}

\begin{figure}[t] 
\centering
  \subfigure[][]{%
    \includegraphics[width=0.15\textwidth]{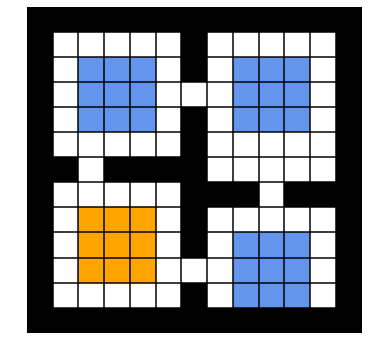}
  } 
  \quad
  \subfigure[][]{%
    \includegraphics[width=0.25\textwidth]{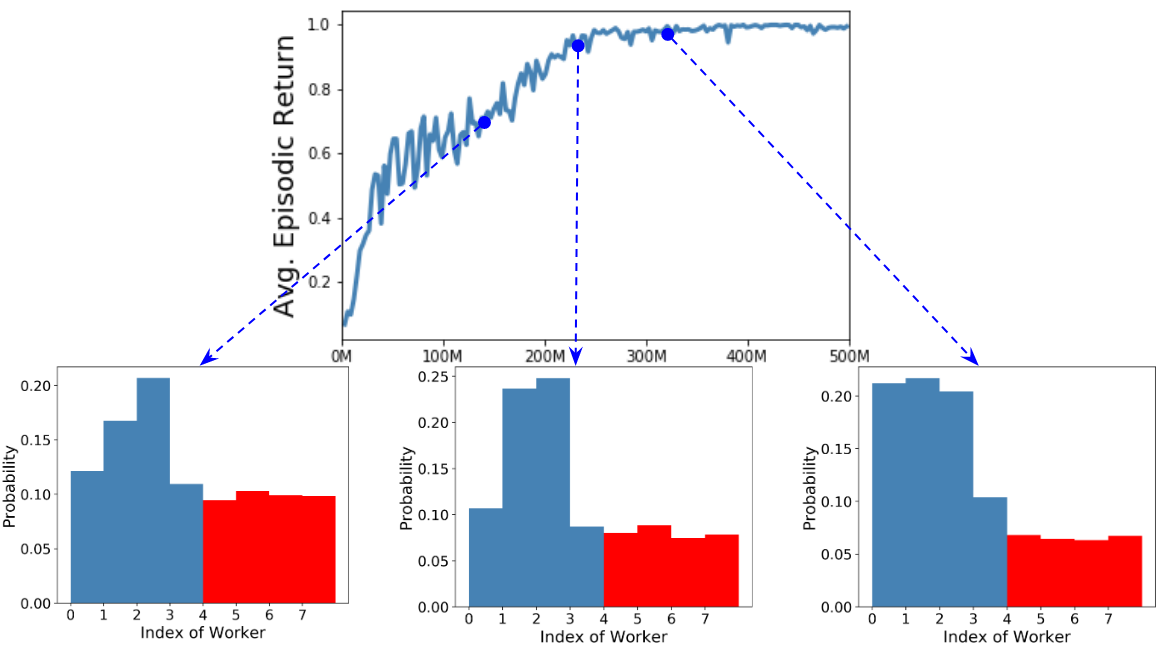}
  }
  \caption{Figure (a) shows the gridworld's layout, the subset of goal locations (in light blue) that are used as training tasks and the disjoint set of goal locations (in orange) used as test tasks. Figure (b) shows how MODAC chose among options and primitive actions at three different points during training, as those options were being discovered. In the histogram, each of blue and red bars measure the frequency of selecting an option and primitive action respectively. From this, we can see that the agent picked options more often than primitive actions throughout the training phase.\vspace{-0.3in}} \label{fig:four_room_train_test_distribution_and_usage} 
\end{figure}

\noindent \textbf{Domain Description:} Consider a simple gridworld with $4$ connected rooms, whose layout is shown in Fig.~\ref{fig:four_room_train_test_distribution_and_usage}a. The agent receives a reward of $1$ on reaching the goal and $0$ at every other time step. During training, at the start of each episode, a new training-goal location is randomly chosen from the set shown in blue in Fig.~\ref{fig:four_room_train_test_distribution_and_usage}a. During testing, a goal position is chosen randomly from a disjoint set shown in orange in Fig.~\ref{fig:four_room_train_test_distribution_and_usage}a and remains fixed for all episodes. The agent's observations included its current location and the grid's layout (in $2$ separate image channels). The agent was also fed the goal location during training (as the $3$rd channel). Each test task was evaluated separately and the overall transfer performance was computed by averaging across all test runs. 

\begin{figure*}[t] 
\centering
  \subfigure[][]{%
    \includegraphics[width=0.18\textwidth]{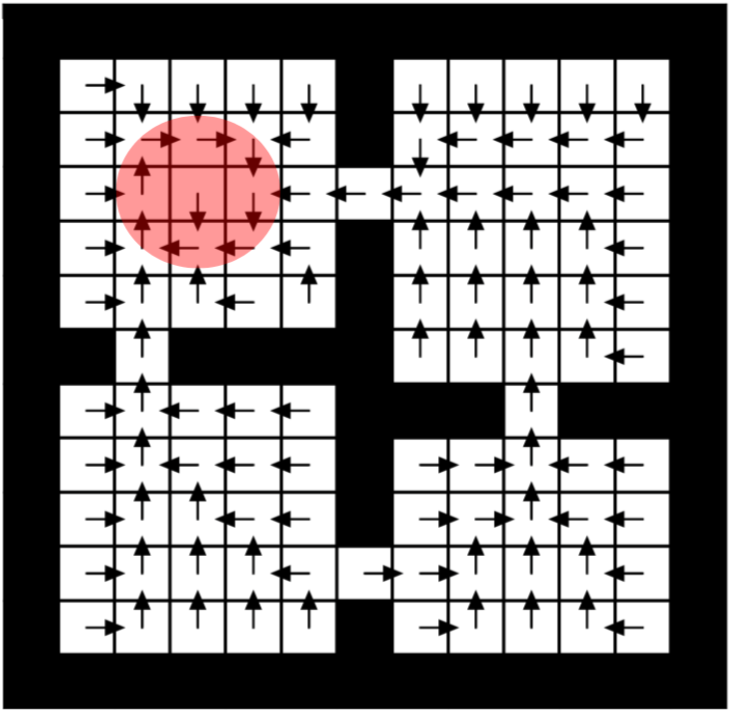}
  } 
  \quad 
  \subfigure[][]{%
    \includegraphics[width=0.18\textwidth]{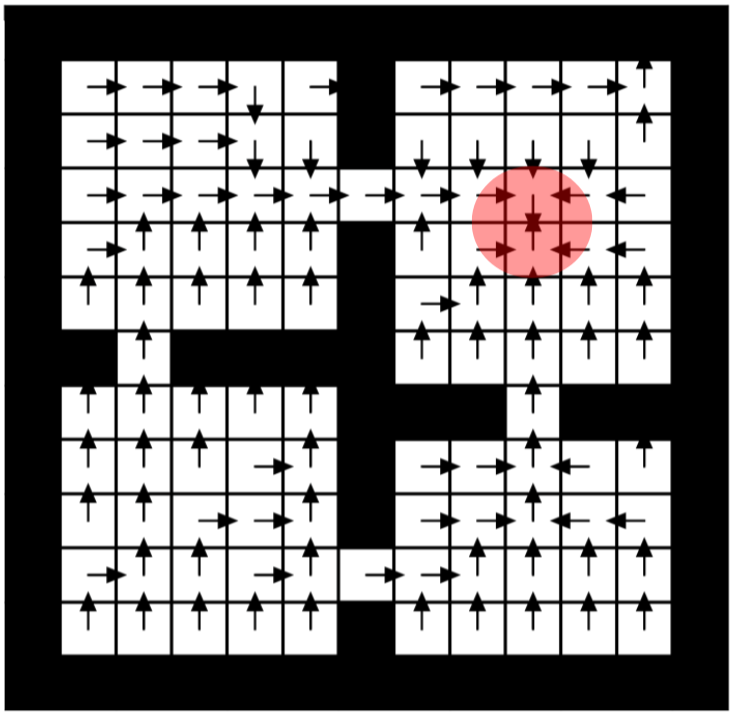}
  } 
  \quad 
  \subfigure[][]{%
    \includegraphics[width=0.18\textwidth]{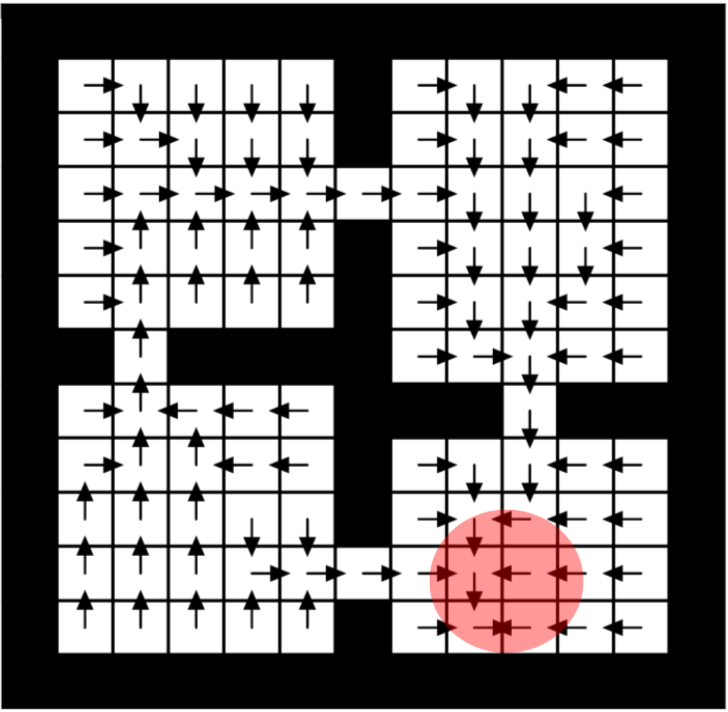}
  } 
  \quad
  \subfigure[][]{%
    \includegraphics[width=0.18\textwidth]{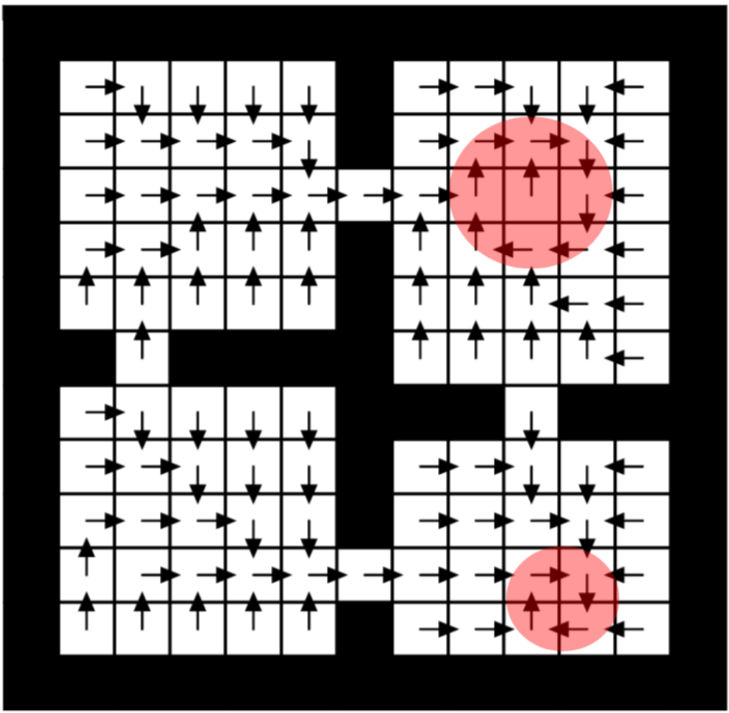}
  } 
  \caption{The red circle approximately marks the destination/subgoal states for each discovered option-policy (arrows indicate the direction of movement). Options (a$-$c) each led the agent directly to one of the three areas where the training tasks were concentrated. Option (d) appeared to be redundant and led to either one of two such areas depending on the start state.} \label{fig:four_room_options}
\end{figure*}

\noindent \textbf{Visualisation:}\label{sec:visualise} Fig.~\ref{fig:four_room_options} shows the option-policies discovered in the gridworld (arrows indicate the direction of movement), at the end of training. 
We found these discovered options to make intuitive sense, given the distribution of training tasks. The first three options each led into one of the rooms where the training goal states (in blue) were concentrated: specifically, from any state the options shown in Fig.~\ref{fig:four_room_options}a$-$c steered the agent to the upper-left, upper-right and lower-right rooms, respectively. The option in Fig.~\ref{fig:four_room_options}d seems redundant given the options in Fig.~\ref{fig:four_room_options}b and c. We hypothesise that such redundancy was due to the training task distribution only featuring goals in three rooms, therefore requiring only three options to capture most of the structure in the space of behaviours required by the training tasks. 
\begin{figure}[htp] 
\centering
    \subfigure[][]{%
    \includegraphics[width=0.2\textwidth]{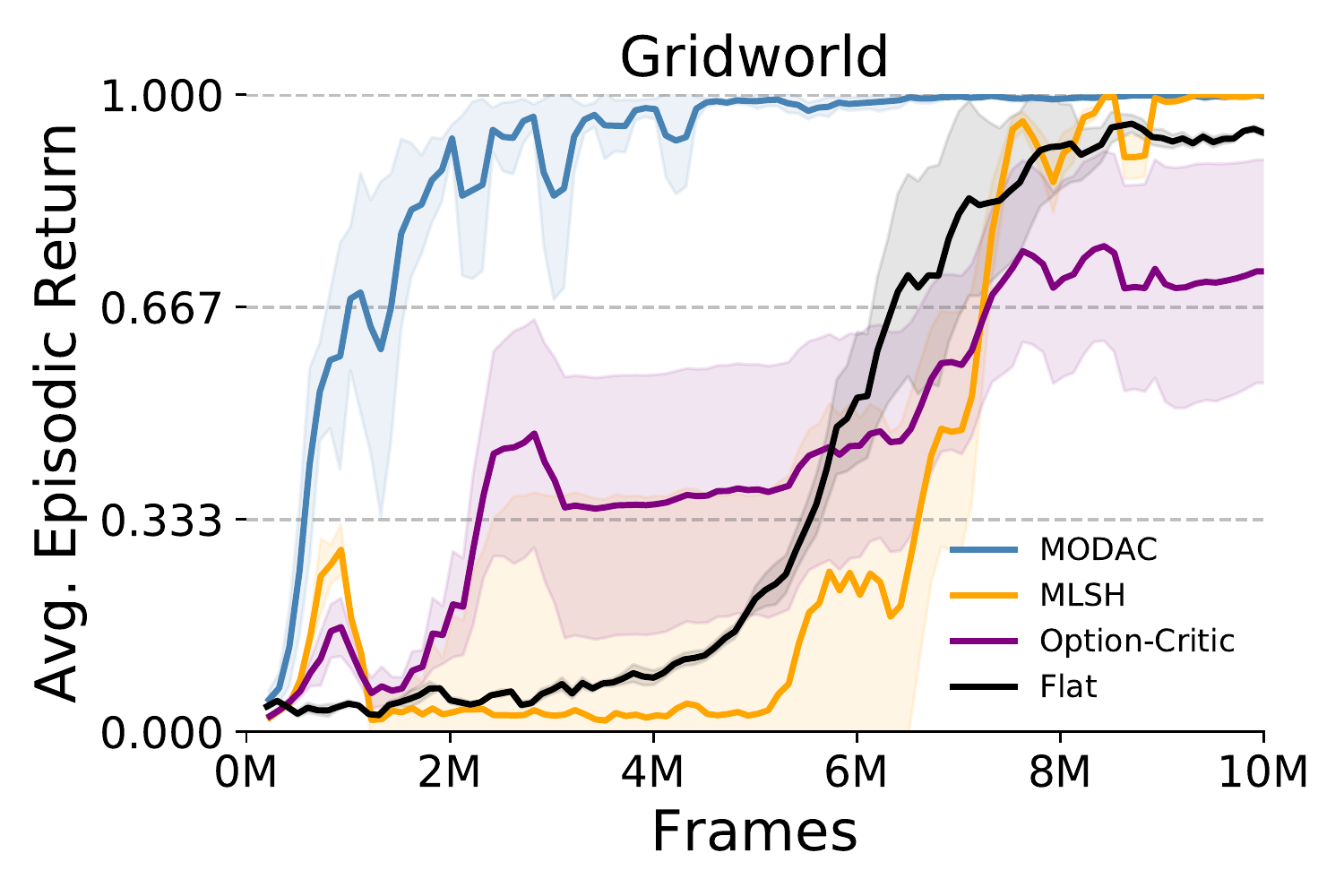}
    }
    \quad
    \subfigure[][]{%
    \includegraphics[width=0.2\textwidth]{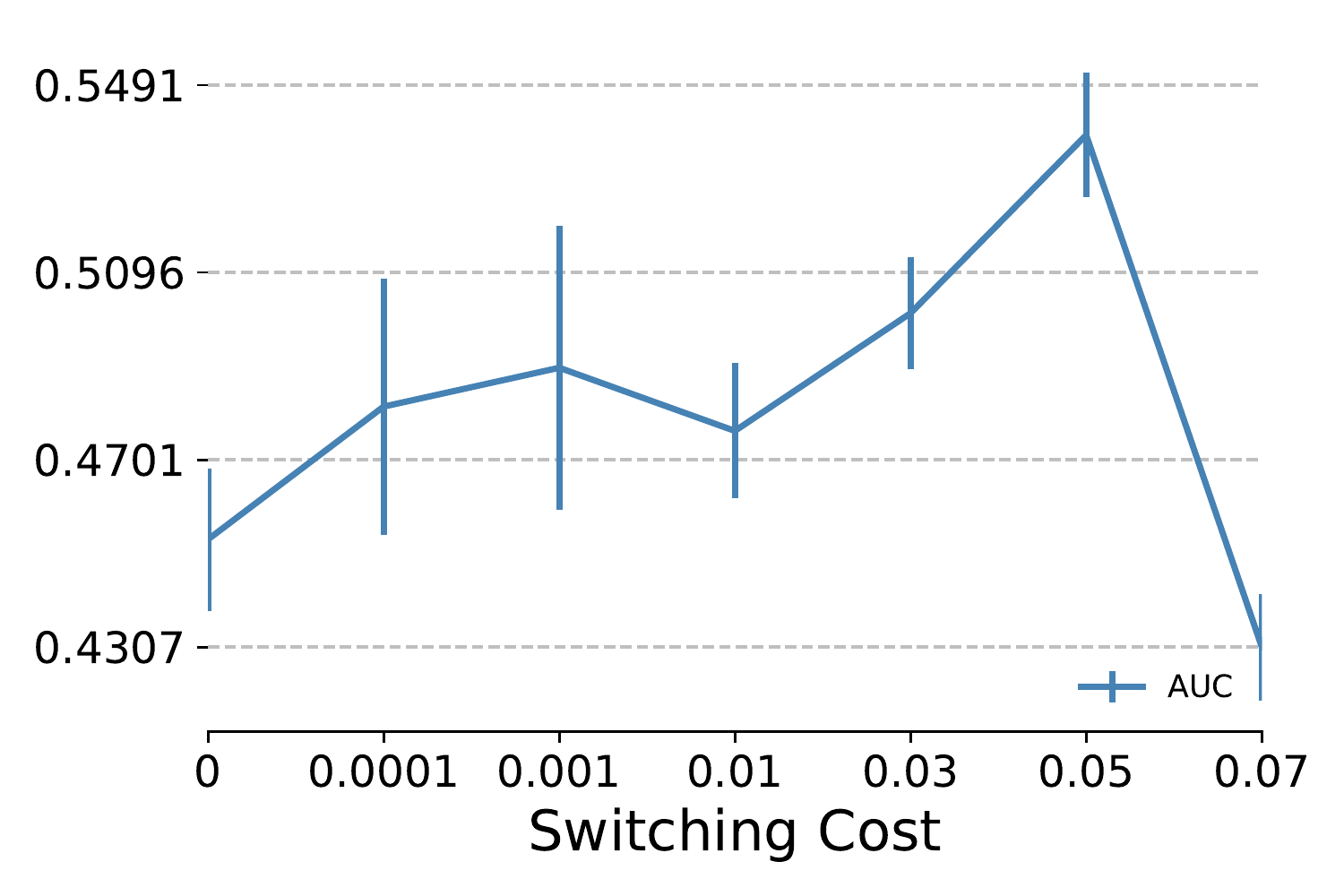}
  } 
  \caption{Figure (a) shows the (transfer) performance on held-out tasks of options learned by MODAC, MLSH, Option-Critic, and a Flat agent. The agent with options discovered by MODAC learned significantly faster than other baselines, demonstrating the usefulness of those options during transfer. Figure (b) shows MODAC's average (transfer) performance, as a function of switching cost $c$ used during training.}\label{fig:transfer_four_room} 
\end{figure}

\noindent\textbf{Quantitative Analysis:} 
(1) We measured how the manager selected among options and primitive actions at three points during \textit{training}. Fig.~\ref{fig:four_room_train_test_distribution_and_usage}b plots the manager's choices as histograms. The $4$ blue bars denote how often each of the $4$ options was picked, red bars depict how often primitive actions were chosen. The manager continued to pick both options and primitive actions throughout training. Consistent with the option-redundancy identified in Sec.~\ref{sec:visualise}, only $3$ of the $4$ options were selected frequently towards the end of training.~\\
(2) Option-policies lasted $5.46$ steps on average, and so were temporally extended and given how often options are picked were overwhelmingly responsible for behaviour.~\\
(3) Fig.~\ref{fig:transfer_four_room}a shows the performance on \textit{test tasks}, when a randomly initialised manager was provided with the (fixed) options discovered by MODAC at the end of the training phase from $6$ independent runs ($10$M frames were used for training). The speed of learning on the test tasks was substantially faster compared to a Flat agent that only used primitive actions; it also outperformed the baselines with access to the options discovered by MLSH or the Option-Critic.~\\
(4) During transfer to test tasks, the manager selected options $56.11\%$ of the time. But recall that options last more than 5 steps and so in effect options controlled behavior more than $85\%$ of the time.~\\
(5) Fig.~\ref{fig:transfer_four_room}b shows transfer performance (averaged throughout the test phase) as a function of the switching cost used in training. We found a sweet spot at a cost of $0.05$, for which the discovered options were maximally useful.



\subsection{Scaling Up to DeepMind Lab}\label{sec:dmlab_expt}
\begin{figure*}[h!] 
\centering
  \subfigure[][]{%
    \includegraphics[width=0.2\linewidth]{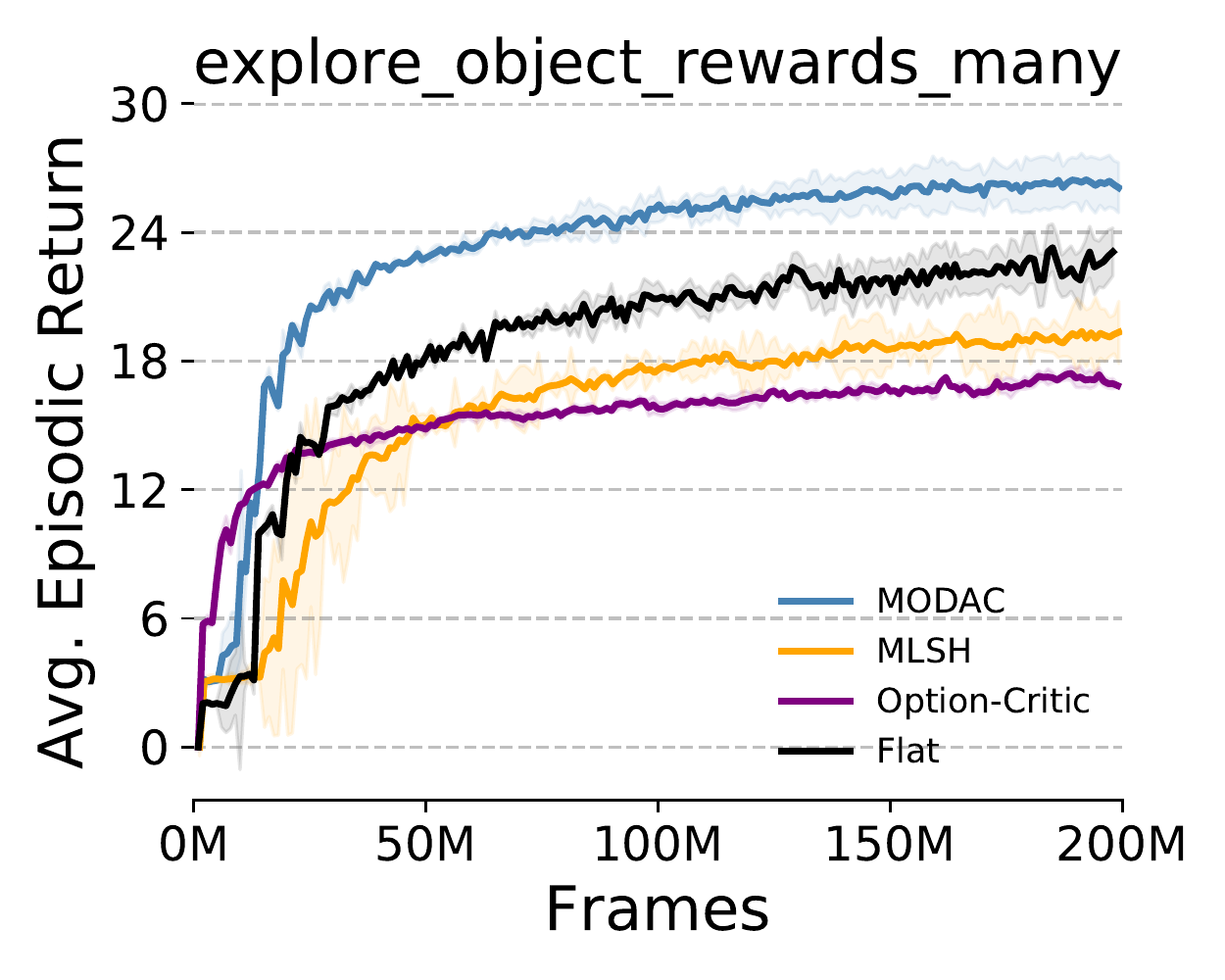}
  } 
  \quad 
  \subfigure[][]{%
    \includegraphics[width=0.2\linewidth]{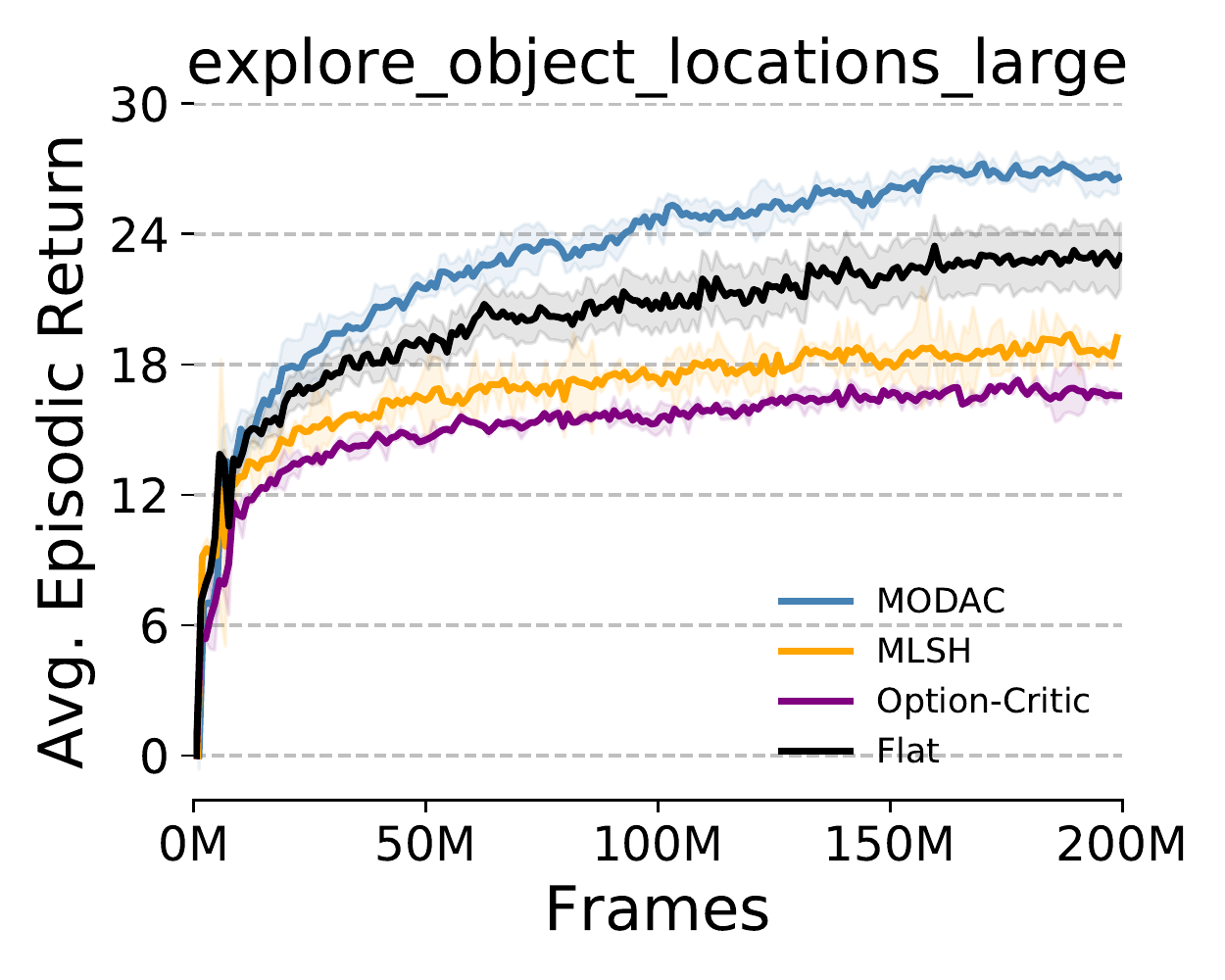}
  } 
  \quad 
  \subfigure[][]{%
    \includegraphics[width=0.2\linewidth]{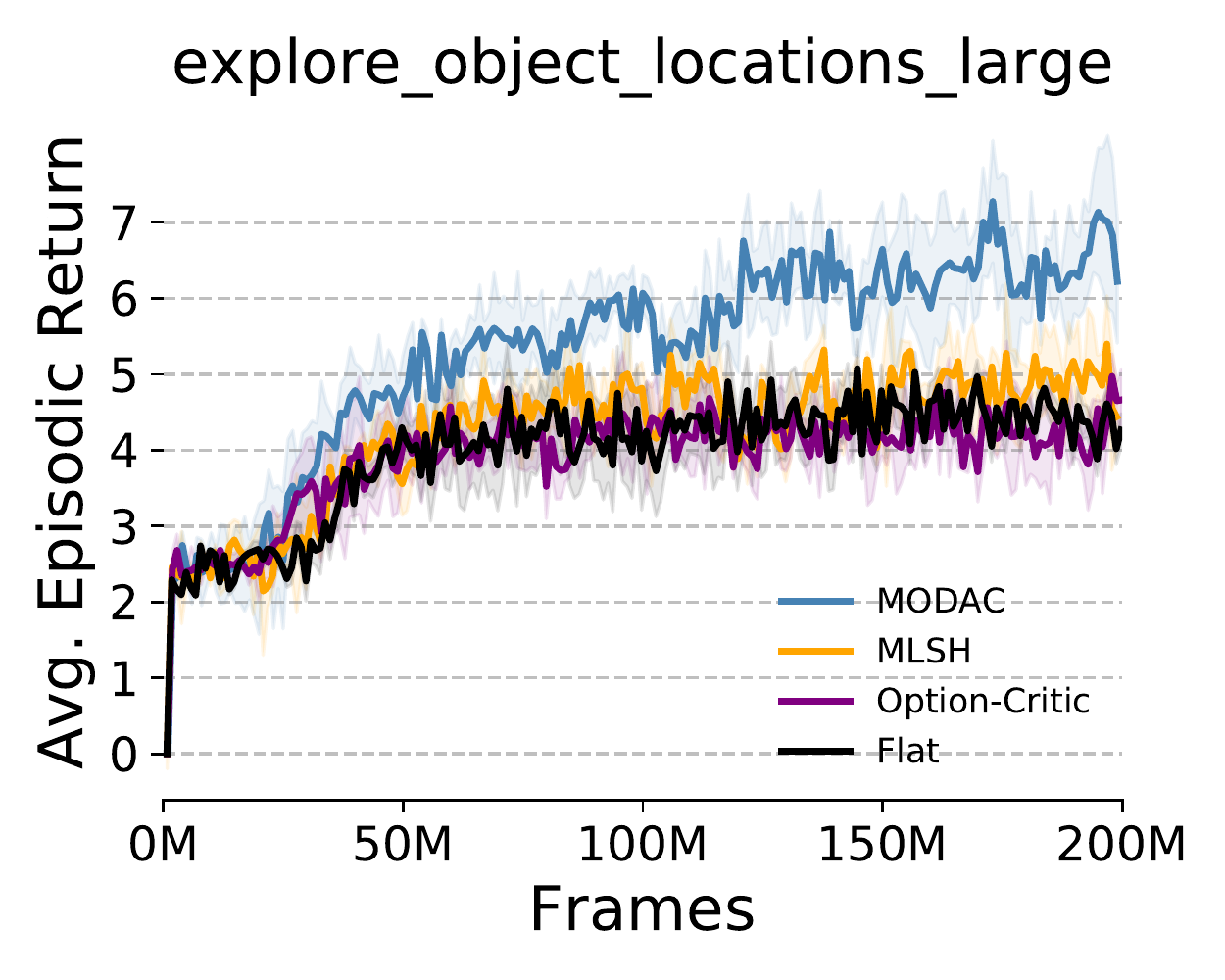}
  }
  \quad
  \subfigure[][]{%
    \includegraphics[width=0.2\linewidth]{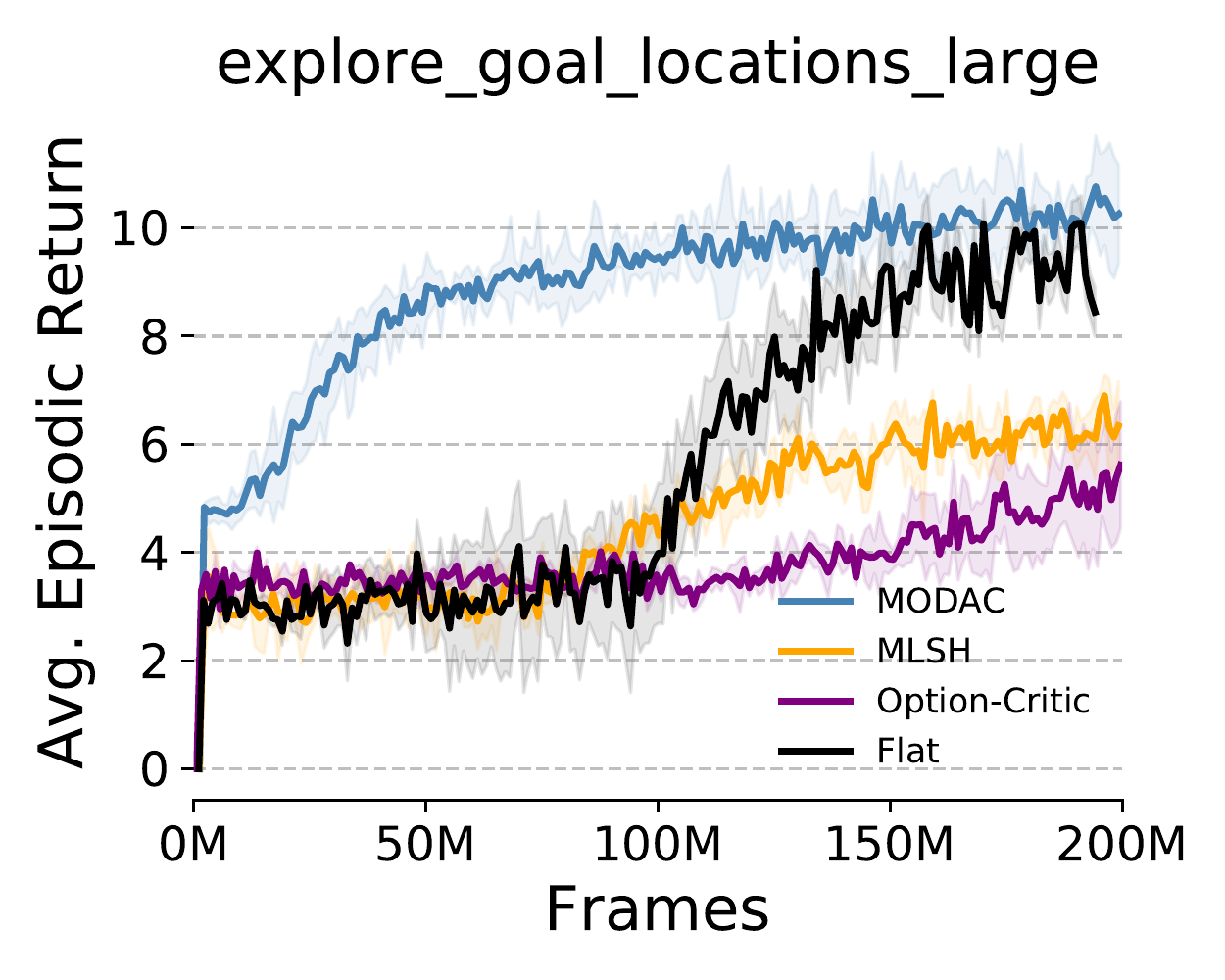}
  }
  \caption{Transfer experiments on DeepMind Lab. Options, discovered from training tasks, were transferred to the corresponding test tasks for hierarchical agents. Figures show the (transfer) learning performance of different agents while they learned to maximise rewards in the test task. MODAC with discovered options learned better and thus was able to achieve better asymptotic performance on $3$ of those $4$ new, unseen tasks, while learning substantially faster on the $4$th task.} \label{fig:dmlab_transfer_learning} 
\end{figure*}

We applied MODAC to DeepMind Lab~\citep{Beattie2016}, a challenging suite of RL tasks with consistent physics and action spaces, and a first-person view as observations to the agent (hence there is partial observability). We evaluated our approach on $4$ different task \textbf{sets} (see titles of Fig.~\ref{fig:dmlab_transfer_learning} for their names), where each set corresponds to a different type of navigation problem. For instance, the task set \texttt{explore\_goal\_locations} requires the agent to explore a maze to visually identify and then reach a goal location identifiable by a special 3D marker.
Each set includes `simple' and `hard' variants; in simple variants the layout includes at most a handful of rooms, while harder variants require navigation in large mazes. In both cases, procedural generation is used to create a different layout for every episode. The challenge is to discover options on simple tasks that can be useful to learn faster on hard tasks. More details in Appendix.
\begin{figure*}[t] 
\centering
  \subfigure[][]{%
    \includegraphics[width=0.17\textwidth]{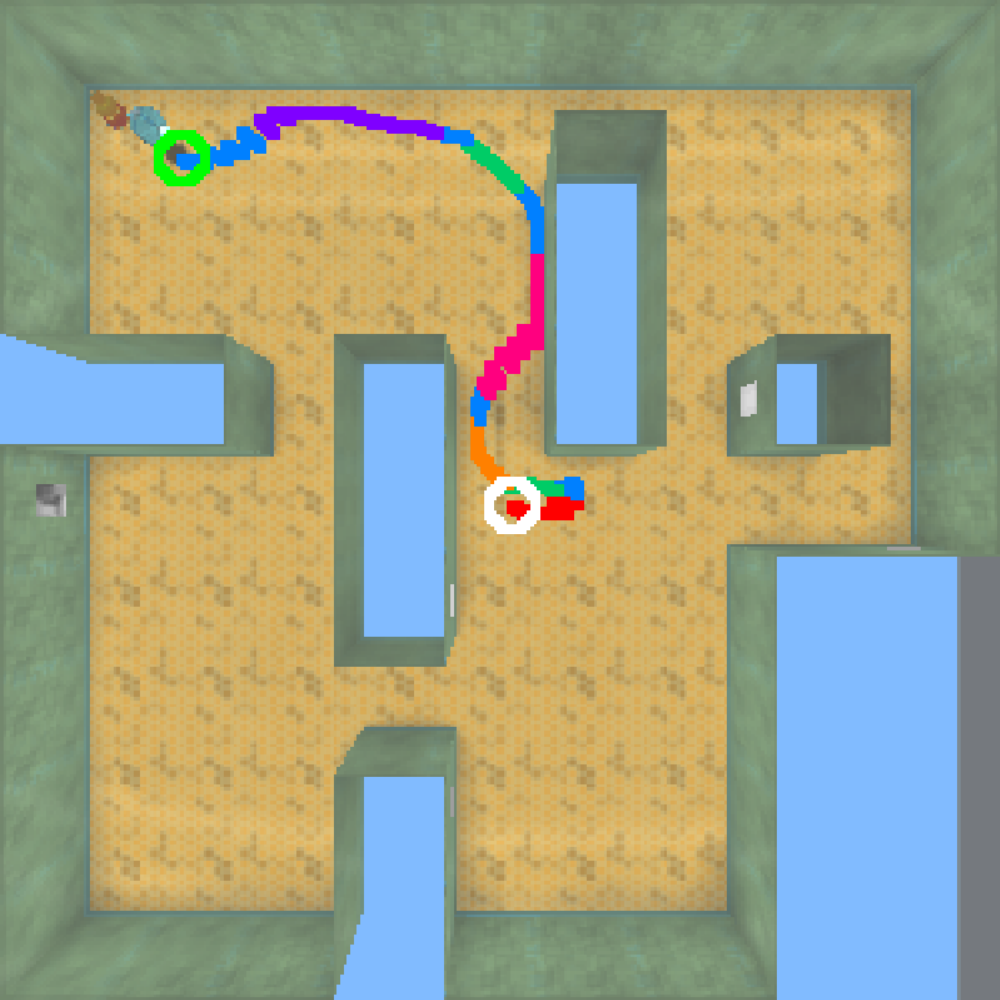}
  }
  \quad
  \subfigure[][]{%
    \includegraphics[width=0.17\textwidth]{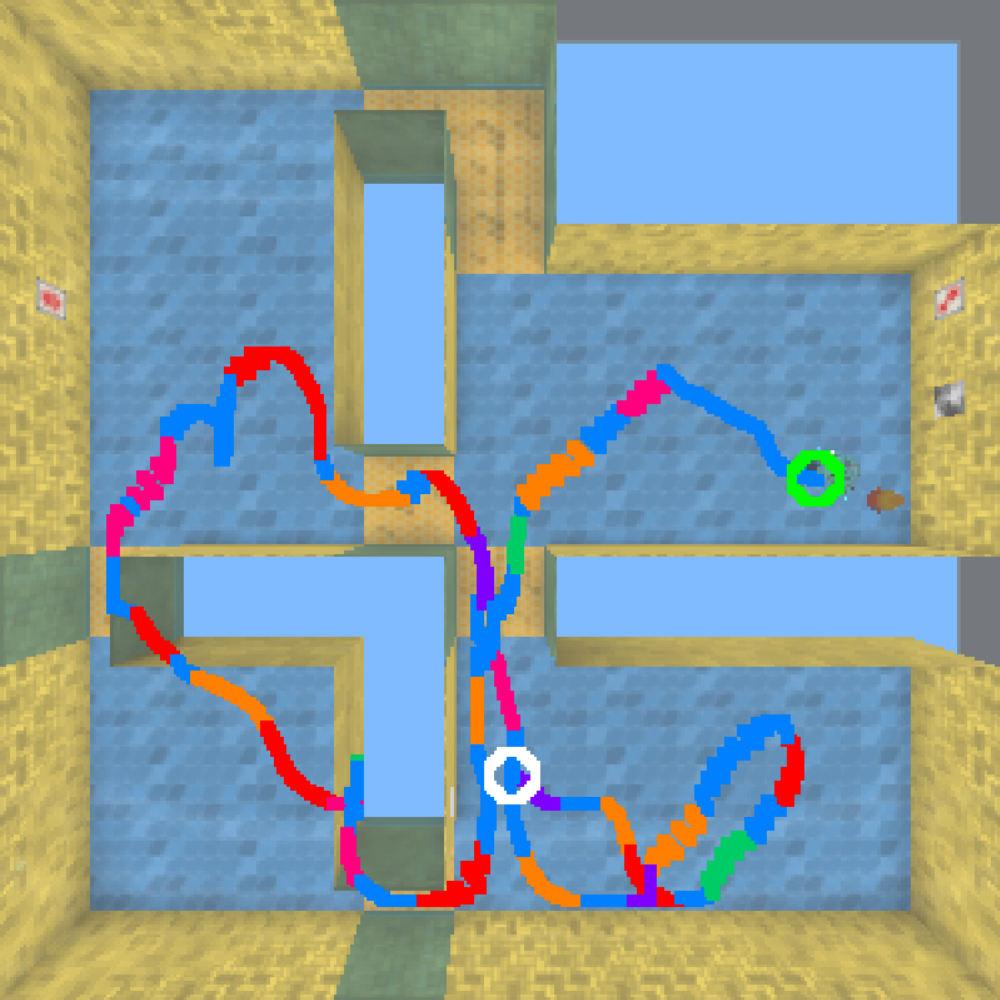}
  } 
  \quad
    \subfigure[][]{%
    \includegraphics[width=0.17\textwidth]{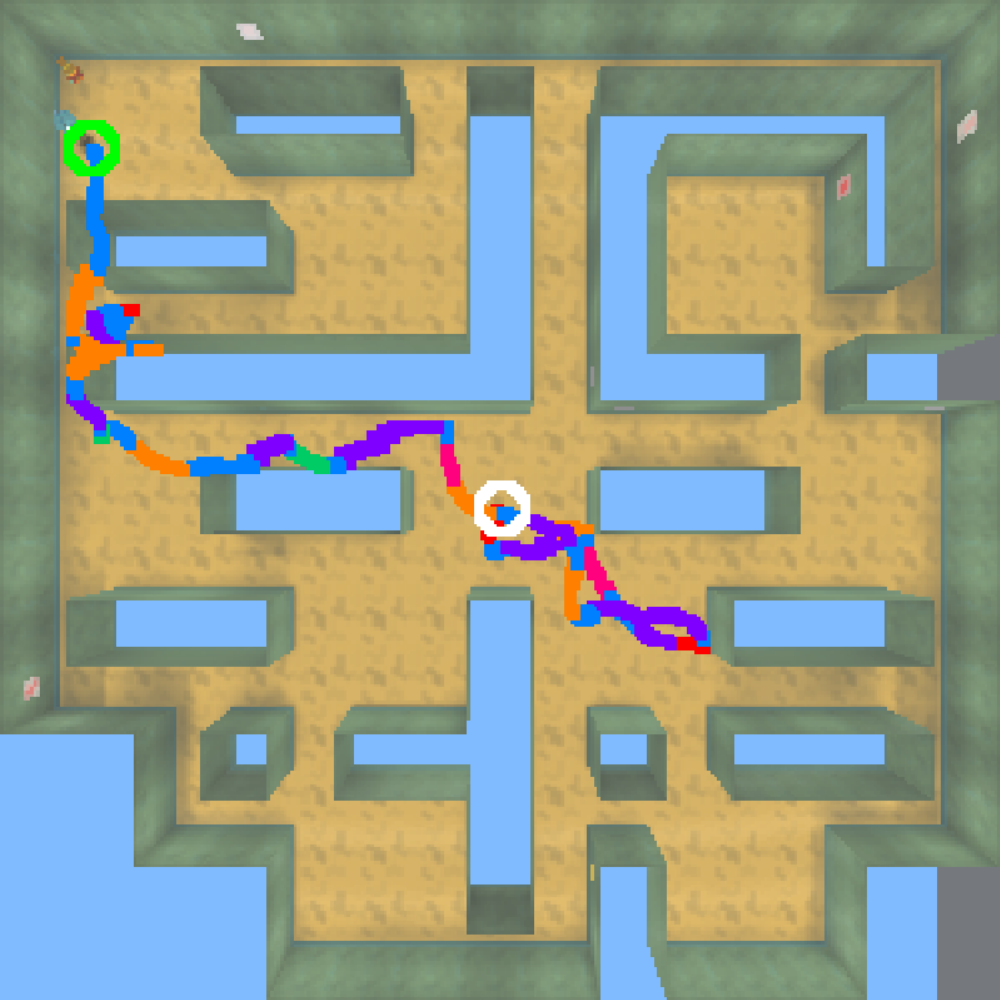}
  }
  \quad
  \subfigure[][]{%
    \includegraphics[width=0.17\textwidth]{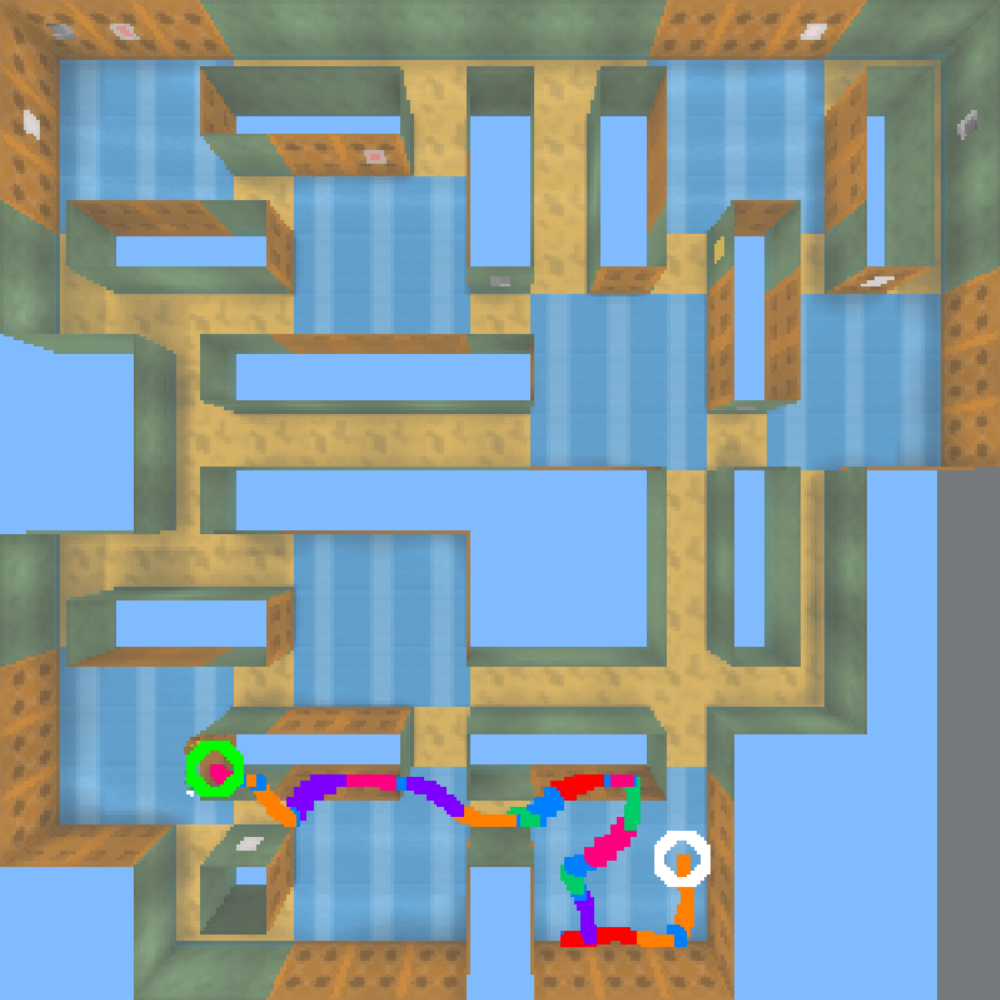}
  } 
  \caption{Option Execution by a Manager. The figures show $4$ distinct samples of trajectories generated by a manager with access to both discovered options (marked in non-blue colours) and primitive actions (in blue). (a) and (b) correspond to mazes in the training set. (c) and (d) are sampled from the testing set. The agent's starting and final positions are highlighted by a white and green circles, respectively. In all cases, the agent successfully achieves the task's objective of reaching the rewarding goal location by using a mixture of primitive actions and discovered options.
  } \label{fig:dmlab_viz_trajectory}
\end{figure*}

\begin{figure*}[t] 
\centering
  \subfigure[][]{%
    \includegraphics[width=0.17\textwidth]{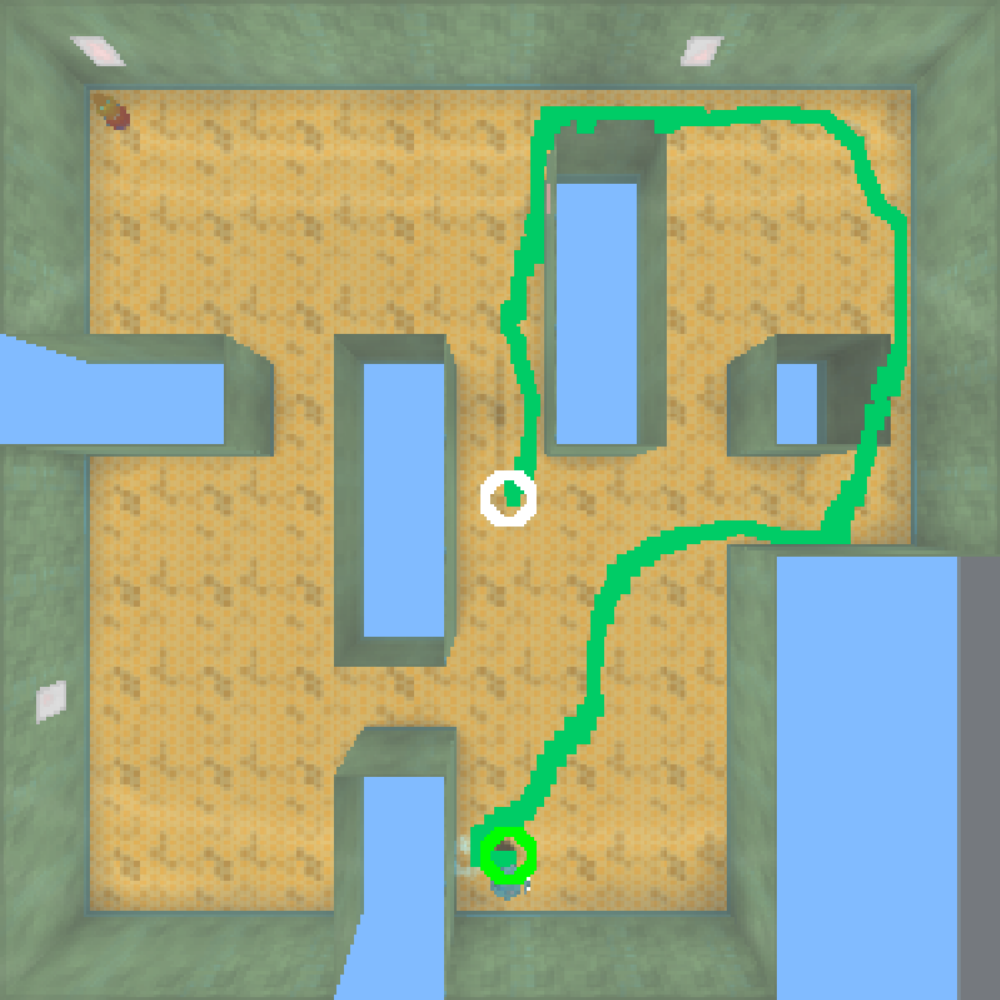}
  } 
  \quad
    \subfigure[][]{%
    \includegraphics[width=0.17\textwidth]{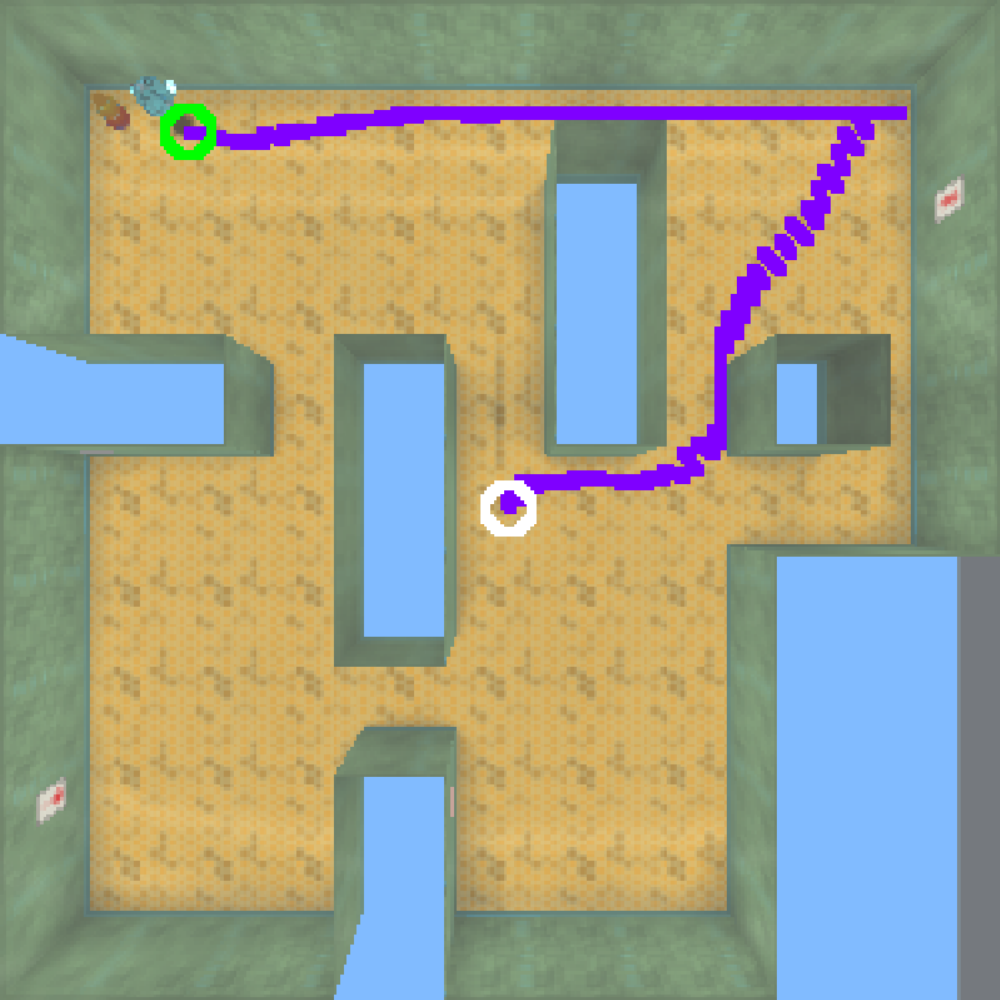}
  }
  \quad
  \subfigure[][]{%
    \includegraphics[width=0.17\textwidth]{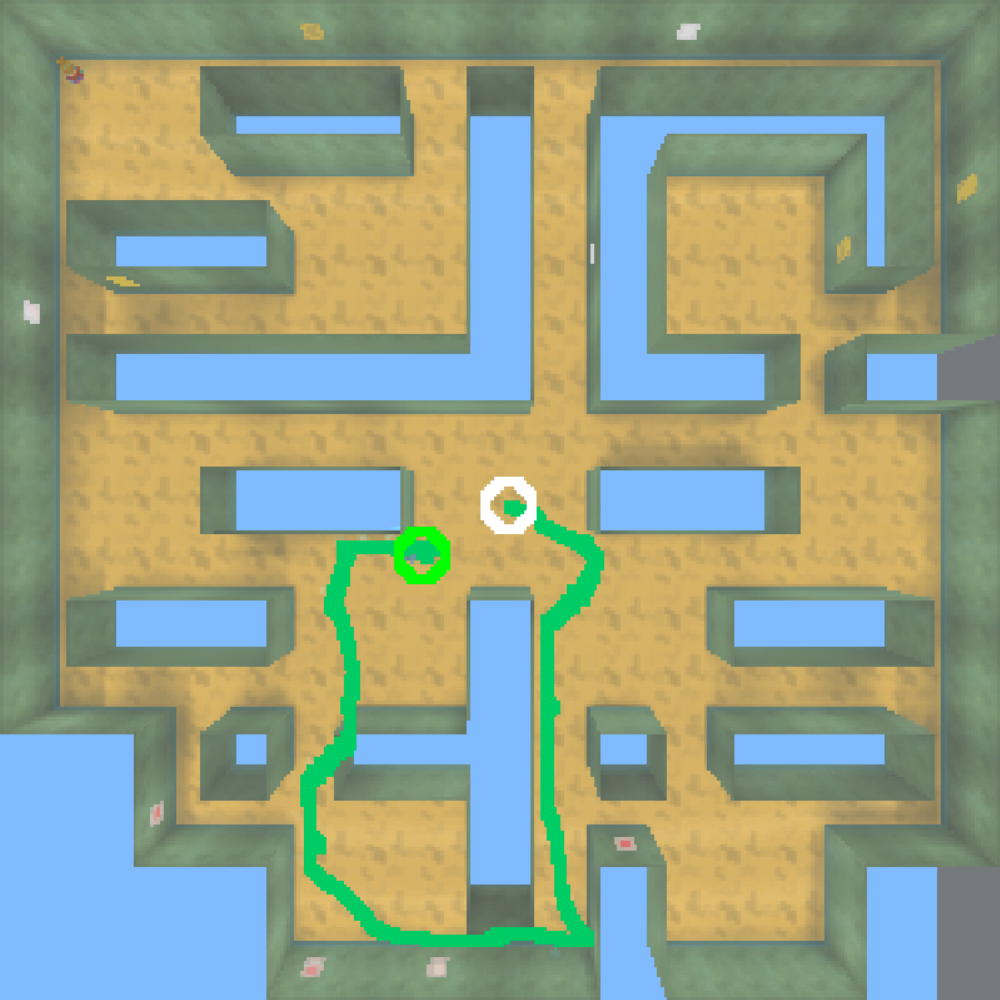}
  } 
  \quad
    \subfigure[][]{%
    \includegraphics[width=0.17\textwidth]{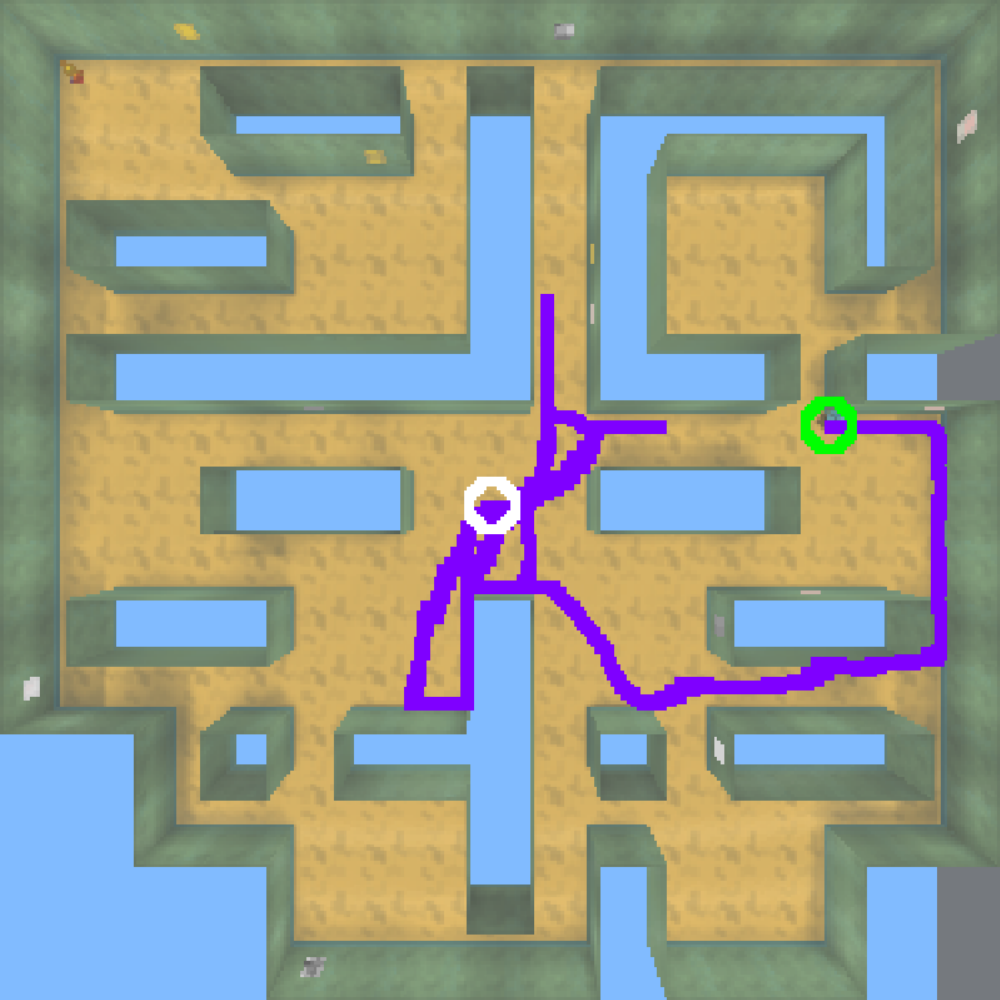}
  }
  \caption{Sampled Option-Policies. The task is to reach the rewarding goal location that was always in the top left corner but not in the line of sight for the agent at the start of the episode. The agent's starting and option-termination positions are highlighted by a white and green circles. In (a) and (b) we show the (very different) trajectories followed by two different option-policies when initialised in the same state within a maze from the training set. In (c) and (d) we show two trajectories in a maze from the test set, for the same pair of option-policies (again initialised in the same state). 
  }\label{fig:dmlab_viz_options_train}
\end{figure*}

\noindent \textbf{Visualisation:} 
In \texttt{explore\_goal\_locations}, the 
state space is obviously too large to visualise entire option policies in one plot as we did for the gridworld. Instead, we generated sample trajectories of experience, and used the DEBUG information provided by DeepMind Lab environments (which is not fed to the agent) to visualise trajectories by drawing the path taken by the agent onto a top-down view of the maze. The segments corresponding to primitive actions were coloured in blue, while those corresponding to the discovered options were assigned an arbitrary different colour. The agent's start and end locations were highlighted by white and green circles, respectively.

We generated sample trajectories using a trained manager with access to both discovered options and primitive actions. When an option is selected, it is executed until the option-termination function is triggered, at which point the manager chooses again. The manager made extensive use of the learned options in both the training tasks, for which a sample is shown in Fig.~\ref{fig:dmlab_viz_trajectory}a and b, and in the testing tasks, shown in Fig.~\ref{fig:dmlab_viz_trajectory}c and d. In the simple training tasks, the manager often only required a handful of option executions to reach the goal. For instance, in Fig.~\ref{fig:dmlab_viz_trajectory}a, after the goal (top-left corner) enters in the line of sight of the agent, it only takes two options to take the agent to the goal. In the hard test task, the manager still relied extensively on the learned options, but it needed to chain a higher number of them during the course of an episode. In Fig.~\ref{fig:dmlab_viz_trajectory}c, the agent was spawned in the centre of the maze at the beginning of an episode. The agent used a mix of options and primitive actions in order to identify and reach the goal location (top-left). In both the training and test tasks, we often found that the manager used options extensively to explore the maze, but it relied on primitive actions for the \textit{last mile navigation} (i.e. in proximity of the goal).

Fig.~\ref{fig:dmlab_viz_options_train} visualised trajectories generated by following $2$ discovered option-policies on the training task set. After seeding the episode in the same way, we observed that the option-policies produced quite diverse behaviours. For instance, we found that the options in Fig.~\ref{fig:dmlab_viz_options_train}a and b explored the maze in very different ways. We note that when the latter entered in line of site of the goal -- in the top left corner -- it marched straight into it. In Fig.~\ref{fig:dmlab_viz_options_train}c and d, we visualised the execution of the same pair of options when triggered on the larger maze from the testing task set; while neither happened to encounter the goal marker in those two episodes, both demonstrated a meaningful temporally extended behaviour that resulted in 
good exploration of a vast portion of the maze.


\noindent\textbf{Quantitative Analysis:} In each of the $4$ training task-sets we discovered $5$ options (using a switching cost $c = 0.03$), and observed that the average length of the options was $12$ steps. Fig.~\ref{fig:dmlab_transfer_learning} shows the performance on testing task sets, averaged across $6$ independent runs, for randomly initialised managers, given access to the pre-trained options discovered by MODAC ($200$M frames were used for training). Those agents learned to maximise rewards faster than the Flat agent that learned with primitive actions alone, and reached higher asymptotic performance in $3$ of them. The transfer performance with MODAC-discovered options was also better than that of the MLSH and Option-Critic baselines in all $4$ domains. We again measured the distribution of the manager's choices at transfer time: options were selected $63.76\%$ of the time, which given that options last about 12 steps implies that our discovered options were responsible for behaviour more than $95\%$ of the time. 

\begin{wrapfigure}{r}{0.25\textwidth}
    \centering
    \includegraphics[width=0.25\textwidth]{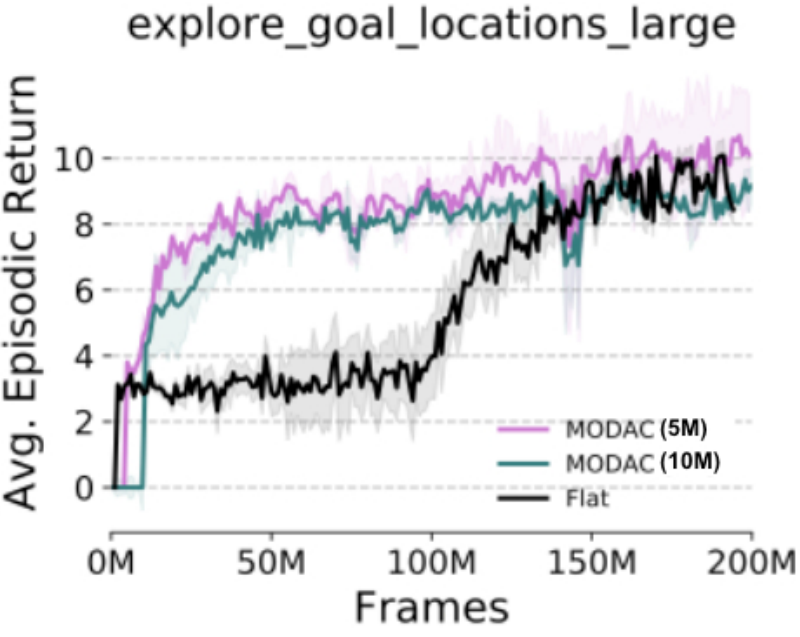}
    \caption{Shows (transfer) performance on a DeepMind Lab task set when fewer samples ($5$M and $10$M) are used for discovering options during training, compared to $200$M used in our main experiments. Learning curves for MODAC are right-shifted to account for the number of samples used in training.
    }\label{fig:dmlab_sample_efficiency}
\end{wrapfigure}

\noindent\textbf{Additional Results:} We studied the effect of MODAC's transfer performance when fewer samples are used in the training phase. Fig.~\ref{fig:dmlab_sample_efficiency} shows performance on one DeepMind Lab task set when only $5$M and $10$M samples are used; \emph{in this case the x-axis includes the training samples for the MODAC curves}. The MODAC agent learned faster than the Flat agent, showing that useful options can be discovered with small enough training samples to outperform the Flat agent in a comparison that takes all samples into account. In Appendix, we present results of additional studies carried out to answer the following empirical questions: (1) How are the options qualitatively different when the number of options (a hyperparameter) is varied? (2) Could MODAC discover options in Atari games from unsupervised training tasks that could be useful to maximise the game score at test time? In addition we provide more qualitative visualisations from the MODAC agent and its option-policies discovered from DeepMind Lab task sets.

\section{Related Work}
\noindent {\bf Hierarchical Reinforcement Learning (HRL):}
Prior HRL work has shown that temporal abstractions lead to faster learning in both single-task~\citep{sutton1999between,dayan1993feudal,dietterich2000hierarchical,Ghavamzadeh2003HierarchicalPG,kulkarni2016hierarchical} and multi-task/transfer setup~\citep{Konidaris2007BuildingPO,Brunskill2014,mankowitz2016adaptive,tessler2017deep}. The aforementioned works require hand-designed temporal abstractions; which can be challenging in scenarios where the agent interacts with a distribution of tasks. In contrast, we aim to discover temporal abstractions purely from experience \textit{without} human supervision or domain knowledge. 

\noindent {\bf Discovery of Temporal Abstractions:}
Majority of the prior HRL work for discovering abstractions are either \textit{unsupervised} approaches\footnote{By unsupervised, we refer to approaches that do not require extrinsic reward signal to drive the discovery process.} or operate in a single-task setup. Older unsupervised approaches to discovery have exploited graph-theoretic properties of the environments such as bottlenecks~\citep{mcgovern2001automatic}, centrality~\citep{csimcsek2005identifying}, and through clustering of states~\citep{mannor2004dynamic}. More recent unsupervised approaches discover options by learning proto-value functions~\citep{machado2017laplacian}, successor feature representations~\citep{c.2018eigenoption}, entropy minimisation~\citep{harutyunyan2019termination}, maximising empowerment~\citep{gregor2016variational}, maximising diversity~\citep{eysenbach2018diversity} or through probabilistic inference~\citep{ranchod2015nonparametric,daniel2016probabilistic}. Recent approaches for discovery in single-task setup learn agents with architectures that are inspired from options framework~\citep{thomas2011policy,mnih2016strategic,bacon2017option,kostas2019asynchronous} and feudal RL framework~\citep{jaderberg2016reinforcement}. In contrast to the aforementioned approaches, our approach is motivated by the multi-task RL setup where the objective is to discover task-independent options that can be reused by the RL agent to master all of its tasks.

Among prior work that investigates discovery in \textit{multi-task} settings, Meta-Learning Shared Hierarchies (MLSH)~\citep{frans2017meta} learns a set of option-policies by directly optimising task rewards via a joint actor-critic update. In contrast, our approach discovers options through the subgoals flexibly defined by option-rewards and terminations; also, our approach learns the temporal scale of each option, while most prior approaches, including MLSH, fix the temporal scale.

\section{Conclusions}
We introduced MODAC, a novel hierarchical agent and a meta-gradient algorithm, for option discovery in a multi-task RL setup. Through visualisations from gridworld and DeepMind Lab, we showed that the options discovered capture diverse and important properties of the behaviours required by the training task distribution. Through DeepMind Lab, we show that MODAC is scalable, and thus, viable for option discovery in challenging RL domains. A promising direction for future work is to investigate whether meta-gradients can also be applied to the discovery of options suitable for planning in \emph{model-based} RL.

\section{Acknowledgements}
Part of this work was conducted at the University of Michigan by Vivek Veeriah where he is supported by DARPA's L2M program. Any opinions, findings, conclusions, or recommendations expressed here are those of the authors and do not necessarily reflect the views of the sponsors.

\bibliography{example_paper}

\begin{thebibliography}{52}
\providecommand{\natexlab}[1]{#1}
\providecommand{\url}[1]{\texttt{#1}}
\expandafter\ifx\csname urlstyle\endcsname\relax
  \providecommand{\doi}[1]{doi: #1}\else
  \providecommand{\doi}{doi: \begingroup \urlstyle{rm}\Url}\fi

\bibitem[Bacon et~al.(2017)Bacon, Harb, and Precup]{bacon2017option}
Bacon, P.-L., Harb, J., and Precup, D.
\newblock The option-critic architecture.
\newblock In \emph{Thirty-First AAAI Conference on Artificial Intelligence},
  2017.

\bibitem[Beattie et~al.(2016)Beattie, Leibo, Teplyashin, Ward, Wainwright,
  K{\"{u}}ttler, Lefrancq, Green, Vald{\'{e}}s, Sadik, Schrittwieser, Anderson,
  York, Cant, Cain, Bolton, Gaffney, King, Hassabis, Legg, and
  Petersen]{Beattie2016}
Beattie, C., Leibo, J.~Z., Teplyashin, D., Ward, T., Wainwright, M.,
  K{\"{u}}ttler, H., Lefrancq, A., Green, S., Vald{\'{e}}s, V., Sadik, A.,
  Schrittwieser, J., Anderson, K., York, S., Cant, M., Cain, A., Bolton, A.,
  Gaffney, S., King, H., Hassabis, D., Legg, S., and Petersen, S.
\newblock Deepmind lab.
\newblock \emph{CoRR}, abs/1612.03801, 2016.

\bibitem[Bradbury et~al.(2018)Bradbury, Frostig, Hawkins, Johnson, Leary,
  Maclaurin, and Wanderman-Milne]{jax2018github}
Bradbury, J., Frostig, R., Hawkins, P., Johnson, M.~J., Leary, C., Maclaurin,
  D., and Wanderman-Milne, S.
\newblock Jax: composable transformations of python+numpy programs, 2018.

\bibitem[Brunskill \& Li(2014)Brunskill and Li]{Brunskill2014}
Brunskill, E. and Li, L.
\newblock Pac-inspired option discovery in lifelong reinforcement learning.
\newblock In \emph{Proceedings of the 31st International Conference on
  International Conference on Machine Learning - Volume 32}, ICML’14, pp.\
  II–316–II–324. JMLR.org, 2014.

\bibitem[Daniel et~al.(2016)Daniel, Van~Hoof, Peters, and
  Neumann]{daniel2016probabilistic}
Daniel, C., Van~Hoof, H., Peters, J., and Neumann, G.
\newblock Probabilistic inference for determining options in reinforcement
  learning.
\newblock \emph{Machine Learning}, 104\penalty0 (2-3):\penalty0 337--357, 2016.

\bibitem[Dayan \& Hinton(1993)Dayan and Hinton]{dayan1993feudal}
Dayan, P. and Hinton, G.~E.
\newblock Feudal reinforcement learning.
\newblock In \emph{Advances in neural information processing systems}, pp.\
  271--271. Morgan Kaufmann Publishers, 1993.

\bibitem[Deitke et~al.(2020)Deitke, Han, Herrasti, Kembhavi, Kolve, Mottaghi,
  Salvador, Schwenk, VanderBilt, Wallingford, et~al.]{deitke2020robothor}
Deitke, M., Han, W., Herrasti, A., Kembhavi, A., Kolve, E., Mottaghi, R.,
  Salvador, J., Schwenk, D., VanderBilt, E., Wallingford, M., et~al.
\newblock Robothor: An open simulation-to-real embodied ai platform.
\newblock In \emph{Proceedings of the IEEE/CVF Conference on Computer Vision
  and Pattern Recognition}, pp.\  3164--3174, 2020.

\bibitem[Dietterich(2000)]{dietterich2000hierarchical}
Dietterich, T.~G.
\newblock Hierarchical reinforcement learning with the maxq value function
  decomposition.
\newblock \emph{Journal of Artificial Intelligence Research}, 13:\penalty0
  227--303, 2000.

\bibitem[Espeholt et~al.(2018)Espeholt, Soyer, Munos, Simonyan, Mnih, Ward,
  Doron, Firoiu, Harley, Dunning, et~al.]{espeholt2018impala}
Espeholt, L., Soyer, H., Munos, R., Simonyan, K., Mnih, V., Ward, T., Doron,
  Y., Firoiu, V., Harley, T., Dunning, I., et~al.
\newblock Impala: Scalable distributed deep-rl with importance weighted
  actor-learner architectures.
\newblock \emph{arXiv preprint arXiv:1802.01561}, 2018.

\bibitem[Eysenbach et~al.(2018)Eysenbach, Gupta, Ibarz, and
  Levine]{eysenbach2018diversity}
Eysenbach, B., Gupta, A., Ibarz, J., and Levine, S.
\newblock Diversity is all you need: Learning skills without a reward function.
\newblock \emph{arXiv preprint arXiv:1802.06070}, 2018.

\bibitem[Frans et~al.(2017)Frans, Ho, Chen, Abbeel, and
  Schulman]{frans2017meta}
Frans, K., Ho, J., Chen, X., Abbeel, P., and Schulman, J.
\newblock Meta learning shared hierarchies.
\newblock \emph{arXiv preprint arXiv:1710.09767}, 2017.

\bibitem[Ghavamzadeh \& Mahadevan(2003)Ghavamzadeh and
  Mahadevan]{Ghavamzadeh2003HierarchicalPG}
Ghavamzadeh, M. and Mahadevan, S.
\newblock Hierarchical policy gradient algorithms.
\newblock In \emph{ICML}, 2003.

\bibitem[Gregor et~al.(2016)Gregor, Rezende, and
  Wierstra]{gregor2016variational}
Gregor, K., Rezende, D.~J., and Wierstra, D.
\newblock Variational intrinsic control.
\newblock \emph{arXiv preprint arXiv:1611.07507}, 2016.

\bibitem[Griewank \& Walther(2008)Griewank and Walther]{griewank2008evaluating}
Griewank, A. and Walther, A.
\newblock \emph{Evaluating derivatives: principles and techniques of
  algorithmic differentiation}.
\newblock SIAM, 2008.

\bibitem[Harb et~al.(2018)Harb, Bacon, Klissarov, and Precup]{harb2018waiting}
Harb, J., Bacon, P.-L., Klissarov, M., and Precup, D.
\newblock When waiting is not an option: Learning options with a deliberation
  cost.
\newblock In \emph{Thirty-Second AAAI Conference on Artificial Intelligence},
  2018.

\bibitem[Harutyunyan et~al.(2019)Harutyunyan, Dabney, Borsa, Heess, Munos, and
  Precup]{harutyunyan2019termination}
Harutyunyan, A., Dabney, W., Borsa, D., Heess, N., Munos, R., and Precup, D.
\newblock The termination critic.
\newblock In \emph{The 22nd International Conference on Artificial Intelligence
  and Statistics}, pp.\  2231--2240, 2019.

\bibitem[Heess et~al.(2016)Heess, Wayne, Tassa, Lillicrap, Riedmiller, and
  Silver]{heess2016learning}
Heess, N., Wayne, G., Tassa, Y., Lillicrap, T., Riedmiller, M., and Silver, D.
\newblock Learning and transfer of modulated locomotor controllers.
\newblock \emph{arXiv preprint arXiv:1610.05182}, 2016.

\bibitem[Imazeki \& Maeno(2003)Imazeki and Maeno]{Imazeki2003}
Imazeki, K. and Maeno, T.
\newblock Hierarchical control method for manipulating/grasping tasks using
  multi-fingered robot hand.
\newblock volume~4, pp.\  3686 -- 3691 vol.3, 11 2003.
\newblock ISBN 0-7803-7860-1.
\newblock \doi{10.1109/IROS.2003.1249728}.

\bibitem[Jaderberg et~al.(2016)Jaderberg, Mnih, Czarnecki, Schaul, Leibo,
  Silver, and Kavukcuoglu]{jaderberg2016reinforcement}
Jaderberg, M., Mnih, V., Czarnecki, W.~M., Schaul, T., Leibo, J.~Z., Silver,
  D., and Kavukcuoglu, K.
\newblock Reinforcement learning with unsupervised auxiliary tasks.
\newblock \emph{arXiv preprint arXiv:1611.05397}, 2016.

\bibitem[Jong et~al.(2008)Jong, Hester, and Stone]{jong2008utility}
Jong, N.~K., Hester, T., and Stone, P.
\newblock The utility of temporal abstraction in reinforcement learning.
\newblock In \emph{AAMAS (1)}, pp.\  299--306. Citeseer, 2008.

\bibitem[Jouppi et~al.(2017)Jouppi, Young, Patil, Patterson, Agrawal, Bajwa,
  Bates, Bhatia, Boden, Borchers, Boyle, Cantin, Chao, Clark, Coriell, Daley,
  Dau, Dean, Gelb, Ghaemmaghami, Gottipati, Gulland, Hagmann, Ho, Hogberg, Hu,
  Hundt, Hurt, Ibarz, Jaffey, Jaworski, Kaplan, Khaitan, Koch, Kumar, Lacy,
  Laudon, Law, Le, Leary, Liu, Lucke, Lundin, MacKean, Maggiore, Mahony,
  Miller, Nagarajan, Narayanaswami, Ni, Nix, Norrie, Omernick, Penukonda,
  Phelps, Ross, Salek, Samadiani, Severn, Sizikov, Snelham, Souter, Steinberg,
  Swing, Tan, Thorson, Tian, Toma, Tuttle, Vasudevan, Walter, Wang, Wilcox, and
  Yoon]{TPUs}
Jouppi, N.~P., Young, C., Patil, N., Patterson, D.~A., Agrawal, G., Bajwa, R.,
  Bates, S., Bhatia, S., Boden, N., Borchers, A., Boyle, R., Cantin, P., Chao,
  C., Clark, C., Coriell, J., Daley, M., Dau, M., Dean, J., Gelb, B.,
  Ghaemmaghami, T.~V., Gottipati, R., Gulland, W., Hagmann, R., Ho, R.~C.,
  Hogberg, D., Hu, J., Hundt, R., Hurt, D., Ibarz, J., Jaffey, A., Jaworski,
  A., Kaplan, A., Khaitan, H., Koch, A., Kumar, N., Lacy, S., Laudon, J., Law,
  J., Le, D., Leary, C., Liu, Z., Lucke, K., Lundin, A., MacKean, G., Maggiore,
  A., Mahony, M., Miller, K., Nagarajan, R., Narayanaswami, R., Ni, R., Nix,
  K., Norrie, T., Omernick, M., Penukonda, N., Phelps, A., Ross, J., Salek, A.,
  Samadiani, E., Severn, C., Sizikov, G., Snelham, M., Souter, J., Steinberg,
  D., Swing, A., Tan, M., Thorson, G., Tian, B., Toma, H., Tuttle, E.,
  Vasudevan, V., Walter, R., Wang, W., Wilcox, E., and Yoon, D.~H.
\newblock In-datacenter performance analysis of a tensor processing unit.
\newblock \emph{CoRR}, abs/1704.04760, 2017.

\bibitem[Kolve et~al.(2017)Kolve, Mottaghi, Han, VanderBilt, Weihs, Herrasti,
  Gordon, Zhu, Gupta, and Farhadi]{kolve2017ai2}
Kolve, E., Mottaghi, R., Han, W., VanderBilt, E., Weihs, L., Herrasti, A.,
  Gordon, D., Zhu, Y., Gupta, A., and Farhadi, A.
\newblock Ai2-thor: An interactive 3d environment for visual ai.
\newblock \emph{arXiv preprint arXiv:1712.05474}, 2017.

\bibitem[Konidaris \& Barto(2007)Konidaris and Barto]{Konidaris2007BuildingPO}
Konidaris, G. and Barto, A.~G.
\newblock Building portable options: Skill transfer in reinforcement learning.
\newblock In \emph{IJCAI}, 2007.

\bibitem[Kostas et~al.(2019)Kostas, Nota, and Thomas]{kostas2019asynchronous}
Kostas, J., Nota, C., and Thomas, P.~S.
\newblock Asynchronous coagent networks: Stochastic networks for reinforcement
  learning without backpropagation or a clock.
\newblock \emph{arXiv preprint arXiv:1902.05650}, 2019.

\bibitem[Kulkarni et~al.(2016)Kulkarni, Narasimhan, Saeedi, and
  Tenenbaum]{kulkarni2016hierarchical}
Kulkarni, T.~D., Narasimhan, K.~R., Saeedi, A., and Tenenbaum, J.~B.
\newblock Hierarchical deep reinforcement learning: Integrating temporal
  abstraction and intrinsic motivation.
\newblock \emph{arXiv preprint arXiv:1604.06057}, 2016.

\bibitem[Machado \& Bowling(2016)Machado and Bowling]{machado2016learning}
Machado, M.~C. and Bowling, M.
\newblock Learning purposeful behaviour in the absence of rewards.
\newblock \emph{arXiv preprint arXiv:1605.07700}, 2016.

\bibitem[Machado et~al.(2017)Machado, Bellemare, and
  Bowling]{machado2017laplacian}
Machado, M.~C., Bellemare, M.~G., and Bowling, M.
\newblock A laplacian framework for option discovery in reinforcement learning.
\newblock In \emph{Proceedings of the 34th International Conference on Machine
  Learning-Volume 70}, pp.\  2295--2304. JMLR. org, 2017.

\bibitem[Machado et~al.(2018)Machado, Rosenbaum, Guo, Liu, Tesauro, and
  Campbell]{c.2018eigenoption}
Machado, M.~C., Rosenbaum, C., Guo, X., Liu, M., Tesauro, G., and Campbell, M.
\newblock Eigenoption discovery through the deep successor representation.
\newblock In \emph{International Conference on Learning Representations}, 2018.

\bibitem[Mankowitz et~al.(2016)Mankowitz, Mann, and
  Mannor]{mankowitz2016adaptive}
Mankowitz, D.~J., Mann, T.~A., and Mannor, S.
\newblock Adaptive skills adaptive partitions (asap).
\newblock In \emph{Advances in Neural Information Processing Systems}, pp.\
  1588--1596, 2016.

\bibitem[Mann \& Mannor(2014)Mann and Mannor]{Mann14}
Mann, T. and Mannor, S.
\newblock Scaling up approximate value iteration with options: Better policies
  with fewer iterations.
\newblock In Xing, E.~P. and Jebara, T. (eds.), \emph{Proceedings of the 31st
  International Conference on Machine Learning}, volume~32 of \emph{Proceedings
  of Machine Learning Research}, pp.\  127--135, Bejing, China, 22--24 Jun
  2014. PMLR.

\bibitem[Mannor et~al.(2004)Mannor, Menache, Hoze, and
  Klein]{mannor2004dynamic}
Mannor, S., Menache, I., Hoze, A., and Klein, U.
\newblock Dynamic abstraction in reinforcement learning via clustering.
\newblock In \emph{Proceedings of the twenty-first international conference on
  Machine learning}, pp.\ ~71, 2004.

\bibitem[McGovern \& Barto(2001)McGovern and Barto]{mcgovern2001automatic}
McGovern, A. and Barto, A.~G.
\newblock Automatic discovery of subgoals in reinforcement learning using
  diverse density.
\newblock 2001.

\bibitem[Nachum et~al.(2018)Nachum, Gu, Lee, and Levine]{nachum2018data}
Nachum, O., Gu, S.~S., Lee, H., and Levine, S.
\newblock Data-efficient hierarchical reinforcement learning.
\newblock In \emph{Advances in Neural Information Processing Systems}, pp.\
  3303--3313, 2018.

\bibitem[Nachum et~al.(2019)Nachum, Tang, Lu, Gu, Lee, and Levine]{Nachum2019}
Nachum, O., Tang, H., Lu, X., Gu, S., Lee, H., and Levine, S.
\newblock Why does hierarchy (sometimes) work so well in reinforcement
  learning?
\newblock \emph{arXiv preprint arXiv:1909.10618}, 2019.

\bibitem[Osband et~al.(2019)Osband, Roy, Russo, and Wen]{Osband2019}
Osband, I., Roy, B.~V., Russo, D.~J., and Wen, Z.
\newblock Deep exploration via randomized value functions.
\newblock \emph{Journal of Machine Learning Research}, 20\penalty0
  (124):\penalty0 1--62, 2019.

\bibitem[Plappert et~al.(2018)Plappert, Andrychowicz, Ray, McGrew, Baker,
  Powell, Schneider, Tobin, Chociej, Welinder, et~al.]{plappert2018multi}
Plappert, M., Andrychowicz, M., Ray, A., McGrew, B., Baker, B., Powell, G.,
  Schneider, J., Tobin, J., Chociej, M., Welinder, P., et~al.
\newblock Multi-goal reinforcement learning: Challenging robotics environments
  and request for research.
\newblock \emph{arXiv preprint arXiv:1802.09464}, 2018.

\bibitem[Rajendran et~al.(2019)Rajendran, Lewis, Veeriah, Lee, and
  Singh]{rajendran2019should}
Rajendran, J., Lewis, R., Veeriah, V., Lee, H., and Singh, S.
\newblock How should an agent practice?
\newblock \emph{arXiv preprint arXiv:1912.07045}, 2019.

\bibitem[Ranchod et~al.(2015)Ranchod, Rosman, and
  Konidaris]{ranchod2015nonparametric}
Ranchod, P., Rosman, B., and Konidaris, G.
\newblock Nonparametric bayesian reward segmentation for skill discovery using
  inverse reinforcement learning.
\newblock In \emph{2015 IEEE/RSJ International Conference on Intelligent Robots
  and Systems (IROS)}, pp.\  471--477. IEEE, 2015.

\bibitem[Riedmiller et~al.(2018)Riedmiller, Hafner, Lampe, Neunert, Degrave,
  Van~de Wiele, Mnih, Heess, and Springenberg]{riedmiller2018learning}
Riedmiller, M., Hafner, R., Lampe, T., Neunert, M., Degrave, J., Van~de Wiele,
  T., Mnih, V., Heess, N., and Springenberg, J.~T.
\newblock Learning by playing-solving sparse reward tasks from scratch.
\newblock \emph{arXiv preprint arXiv:1802.10567}, 2018.

\bibitem[Silver \& Ciosek(2012)Silver and Ciosek]{Silver2012CompositionalPU}
Silver, D. and Ciosek, K.
\newblock Compositional planning using optimal option models.
\newblock In \emph{Proceedings of the 29th International Coference on
  International Conference on Machine Learning}, ICML’12, pp.\  1267–1274,
  Madison, WI, USA, 2012. Omnipress.
\newblock ISBN 9781450312851.

\bibitem[{\c{S}}im{\c{s}}ek et~al.(2005){\c{S}}im{\c{s}}ek, Wolfe, and
  Barto]{csimcsek2005identifying}
{\c{S}}im{\c{s}}ek, {\"O}., Wolfe, A.~P., and Barto, A.~G.
\newblock Identifying useful subgoals in reinforcement learning by local graph
  partitioning.
\newblock In \emph{Proceedings of the 22nd international conference on Machine
  learning}, pp.\  816--823, 2005.

\bibitem[Solway et~al.(2014)Solway, Diuk, C{\'o}rdova, Yee, Barto, Niv, and
  Botvinick]{solway2014optimal}
Solway, A., Diuk, C., C{\'o}rdova, N., Yee, D., Barto, A.~G., Niv, Y., and
  Botvinick, M.~M.
\newblock Optimal behavioral hierarchy.
\newblock \emph{PLoS computational biology}, 10\penalty0 (8), 2014.

\bibitem[Sutton et~al.(1999)Sutton, Precup, and Singh]{sutton1999between}
Sutton, R.~S., Precup, D., and Singh, S.
\newblock Between mdps and semi-mdps: A framework for temporal abstraction in
  reinforcement learning.
\newblock \emph{Artificial intelligence}, 112\penalty0 (1):\penalty0 181--211,
  1999.

\bibitem[Sutton et~al.(2000)Sutton, McAllester, Singh, and
  Mansour]{sutton2000policy}
Sutton, R.~S., McAllester, D.~A., Singh, S.~P., and Mansour, Y.
\newblock Policy gradient methods for reinforcement learning with function
  approximation.
\newblock In \emph{Advances in neural information processing systems}, pp.\
  1057--1063, 2000.

\bibitem[Tessler et~al.(2017)Tessler, Givony, Zahavy, Mankowitz, and
  Mannor]{tessler2017deep}
Tessler, C., Givony, S., Zahavy, T., Mankowitz, D.~J., and Mannor, S.
\newblock A deep hierarchical approach to lifelong learning in minecraft.
\newblock In \emph{Thirty-First AAAI Conference on Artificial Intelligence},
  2017.

\bibitem[Thomas(2011)]{thomas2011policy}
Thomas, P.~S.
\newblock Policy gradient coagent networks.
\newblock In \emph{Advances in Neural Information Processing Systems}, pp.\
  1944--1952, 2011.

\bibitem[Veeriah et~al.(2019)Veeriah, Hessel, Xu, Rajendran, Lewis, Oh, van
  Hasselt, Silver, and Singh]{veeriah2019discovery}
Veeriah, V., Hessel, M., Xu, Z., Rajendran, J., Lewis, R.~L., Oh, J., van
  Hasselt, H.~P., Silver, D., and Singh, S.
\newblock Discovery of useful questions as auxiliary tasks.
\newblock In \emph{Advances in Neural Information Processing Systems}, pp.\
  9306--9317, 2019.

\bibitem[Vezhnevets et~al.(2016)Vezhnevets, Mnih, Osindero, Graves, Vinyals,
  Agapiou, and kavukcuoglu]{mnih2016strategic}
Vezhnevets, A., Mnih, V., Osindero, S., Graves, A., Vinyals, O., Agapiou, J.,
  and kavukcuoglu, k.
\newblock Strategic attentive writer for learning macro-actions.
\newblock In Lee, D.~D., Sugiyama, M., Luxburg, U.~V., Guyon, I., and Garnett,
  R. (eds.), \emph{Advances in Neural Information Processing Systems 29}, pp.\
  3486--3494. Curran Associates, Inc., 2016.

\bibitem[Xu et~al.(2018)Xu, van Hasselt, and Silver]{xu2018meta}
Xu, Z., van Hasselt, H.~P., and Silver, D.
\newblock Meta-gradient reinforcement learning.
\newblock In \emph{Advances in neural information processing systems}, pp.\
  2396--2407, 2018.

\bibitem[Zahavy et~al.(2020)Zahavy, Xu, Veeriah, Hessel, Oh, van Hasselt,
  Silver, and Singh]{zahavy2020self}
Zahavy, T., Xu, Z., Veeriah, V., Hessel, M., Oh, J., van Hasselt, H., Silver,
  D., and Singh, S.
\newblock Self-tuning deep reinforcement learning.
\newblock \emph{arXiv preprint arXiv:2002.12928}, 2020.

\bibitem[Zheng et~al.(2018)Zheng, Oh, and Singh]{zheng2018learning}
Zheng, Z., Oh, J., and Singh, S.
\newblock On learning intrinsic rewards for policy gradient methods.
\newblock In \emph{Advances in Neural Information Processing Systems}, pp.\
  4644--4654, 2018.

\bibitem[Zheng et~al.(2019)Zheng, Oh, Hessel, Xu, Kroiss, van Hasselt, Silver,
  and Singh]{zheng2019can}
Zheng, Z., Oh, J., Hessel, M., Xu, Z., Kroiss, M., van Hasselt, H., Silver, D.,
  and Singh, S.
\newblock What can intrinsic rewards capture?
\newblock \emph{arXiv preprint arXiv:1912.05500}, 2019.

\end{thebibliography}
\bibliographystyle{icml2021}

\clearpage
\onecolumn
\appendix
\section{Appendix}
\subsection{Additional Qualitative Experiments on Gridworld}
\begin{figure*}[h!] 
\centering
  \subfigure[][]{%
    \includegraphics[width=0.19\textwidth]{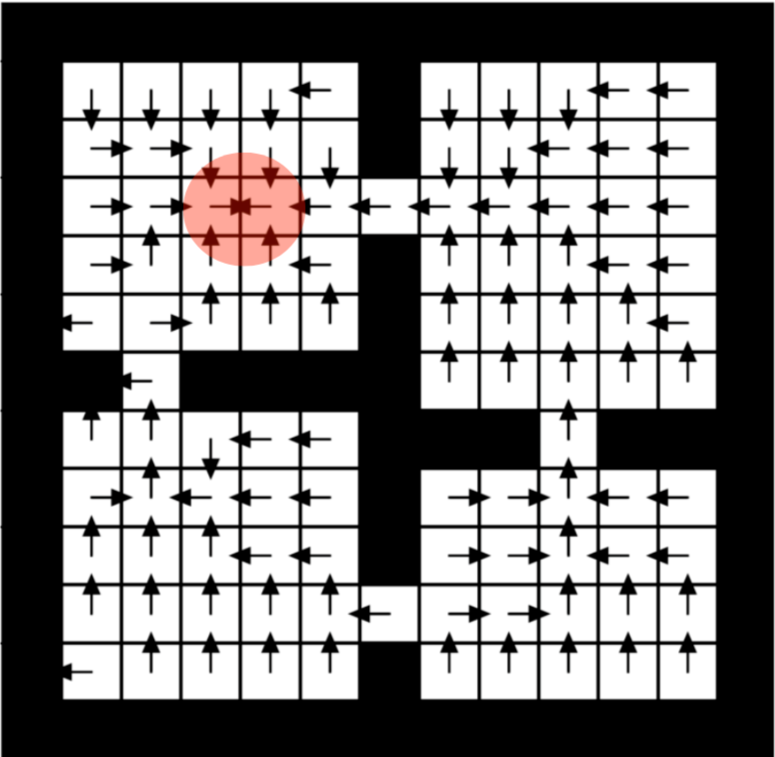}
  } 
  \quad 
  \subfigure[][]{%
    \includegraphics[width=0.19\textwidth]{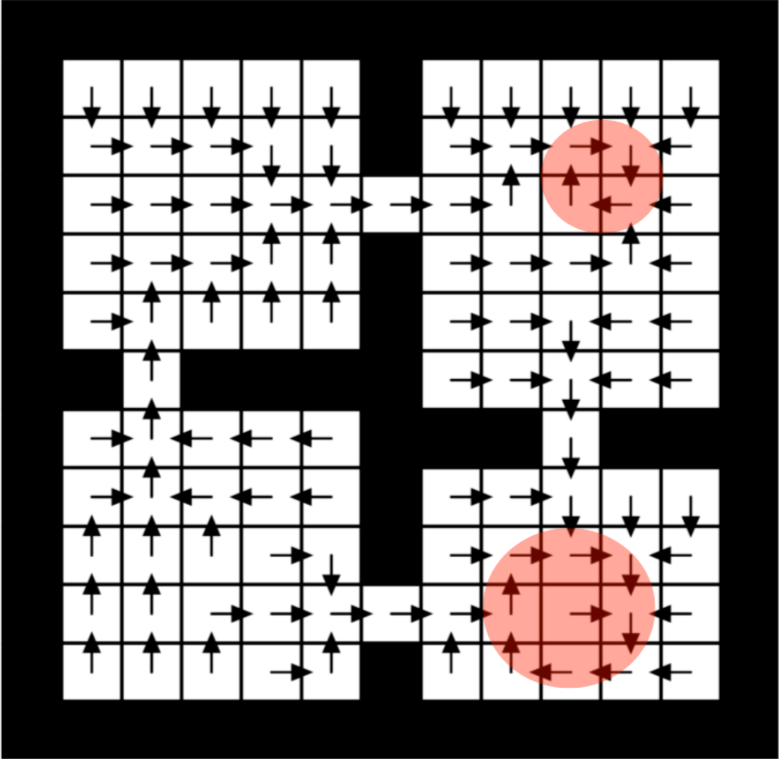}
  } 
  \caption{Option visualisations on the four-room gridworld when $2$ options were discovered. The red circle approximately marks the destination/subgoal states for each discovered option-policy. Option (a) led the agent to the upper-left room, whereas option (b) led to either the upper-right or lower-right rooms depending on the start state.} \label{fig:four_room_options_2_options}
\end{figure*}

\begin{figure*}[h!] 
\centering
  \subfigure[][]{%
    \includegraphics[width=0.19\textwidth]{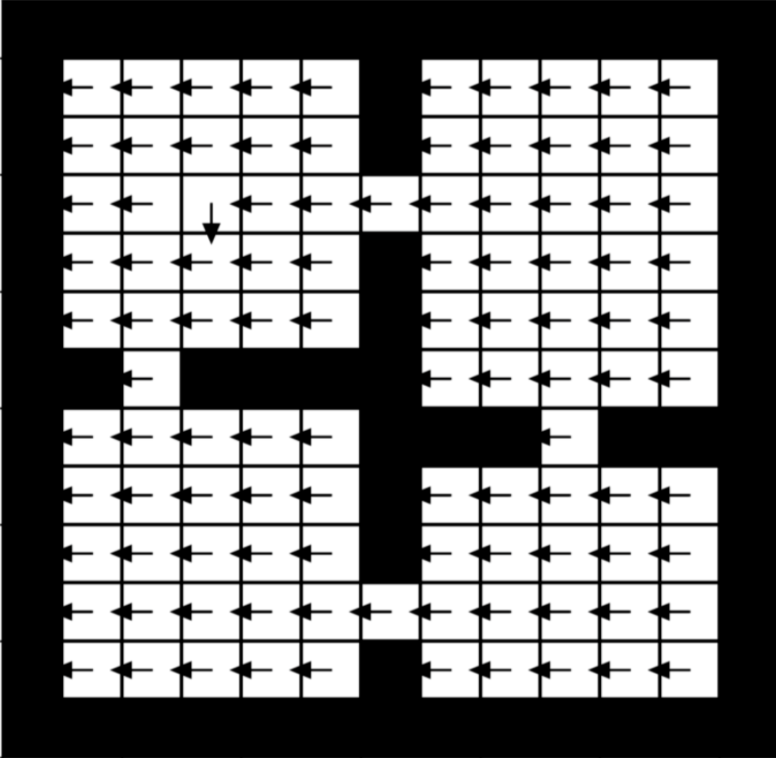}
  } 
  \quad 
  \subfigure[][]{%
    \includegraphics[width=0.19\textwidth]{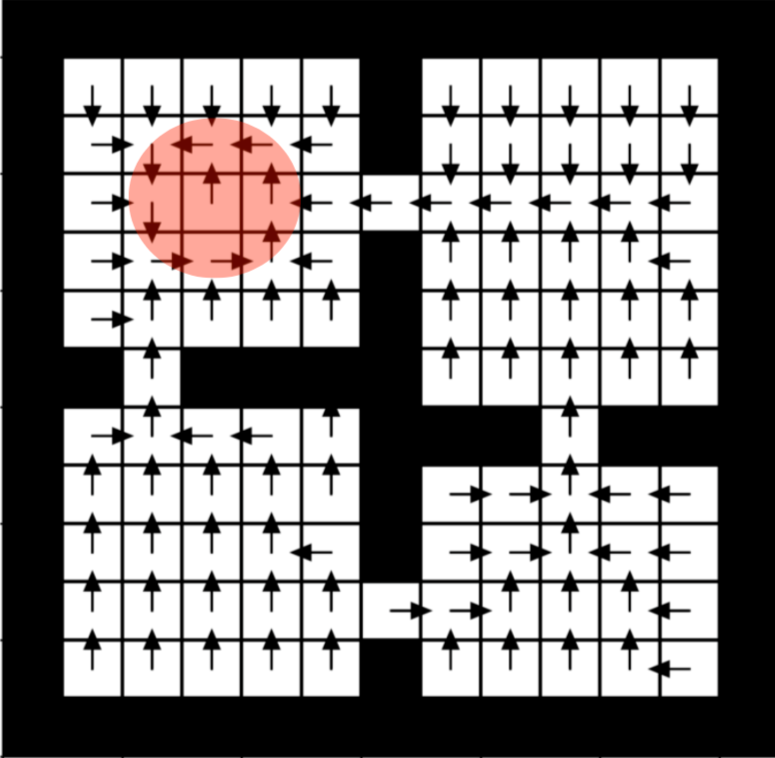}
  }
  \quad 
  \subfigure[][]{%
    \includegraphics[width=0.19\textwidth]{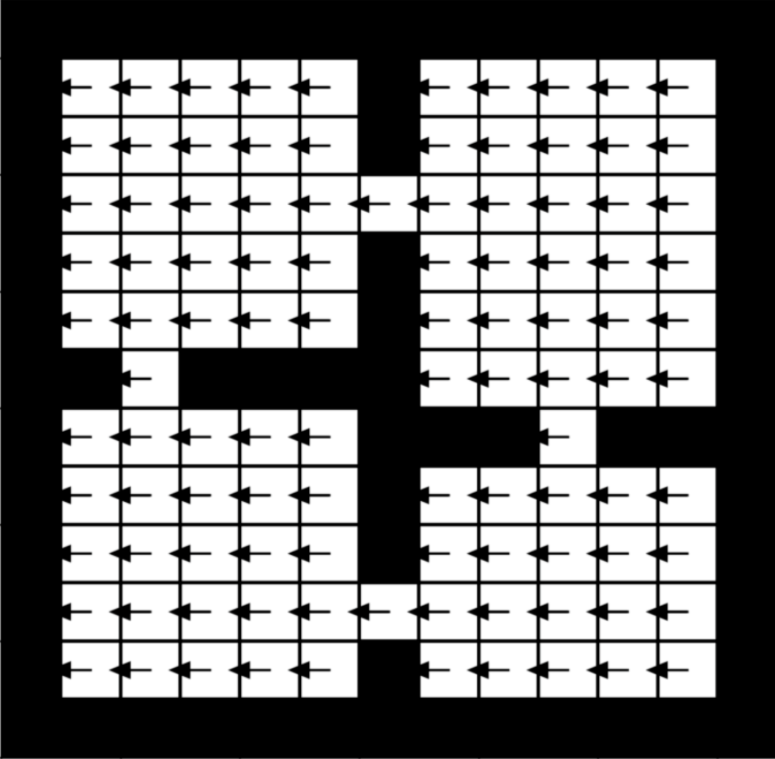}
  }
  \quad 
  \subfigure[][]{%
    \includegraphics[width=0.19\textwidth]{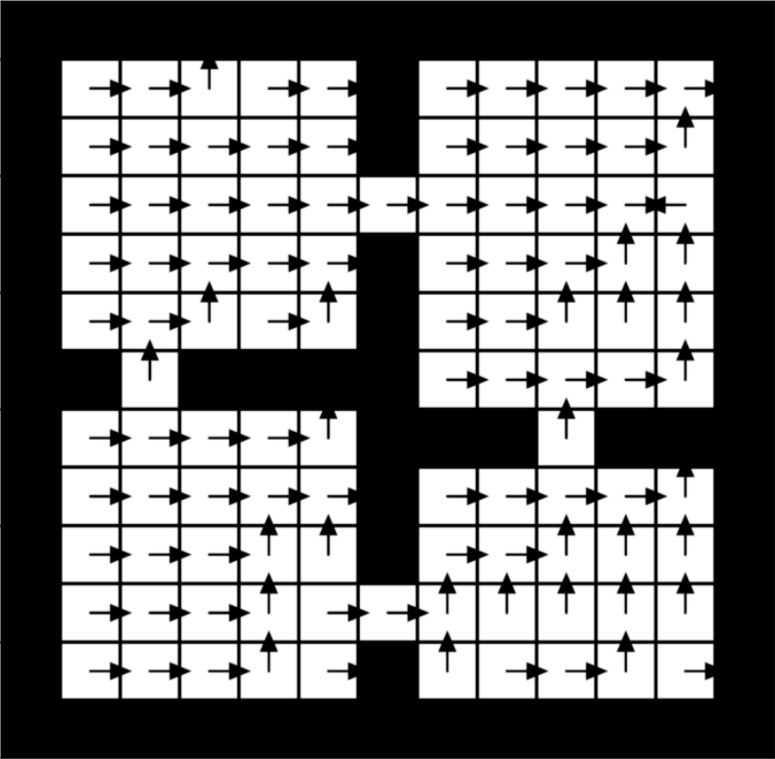}
  }
  \quad 
  \subfigure[][]{%
    \includegraphics[width=0.19\textwidth]{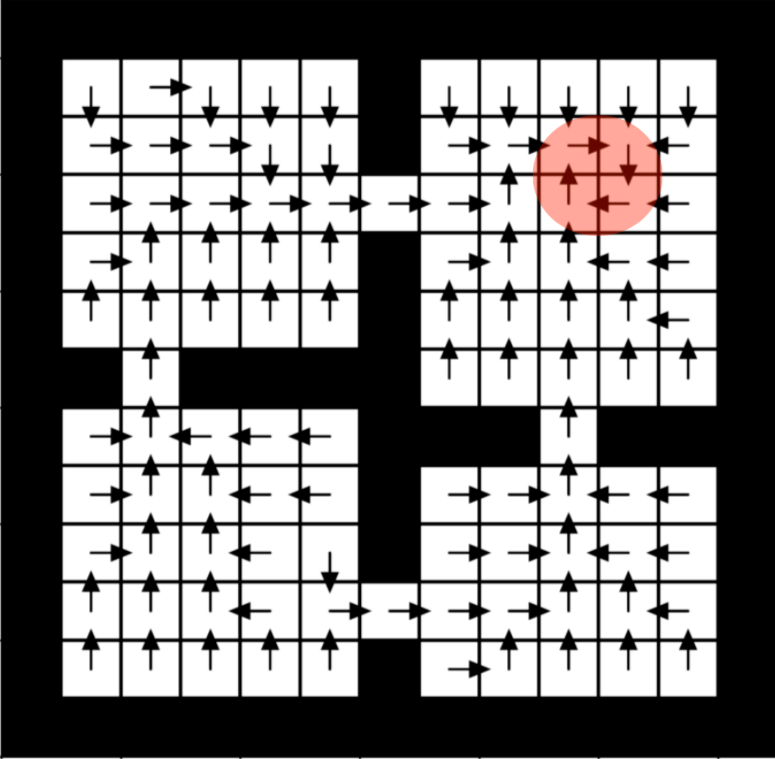}
  }
  \quad 
  \subfigure[][]{%
    \includegraphics[width=0.19\textwidth]{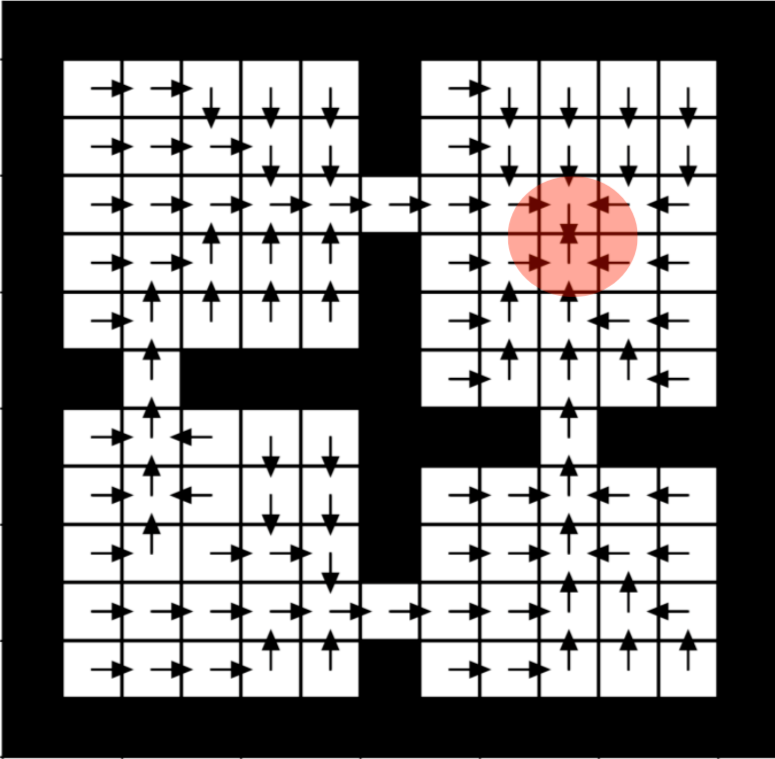}
  }
  \quad 
  \subfigure[][]{%
    \includegraphics[width=0.19\textwidth]{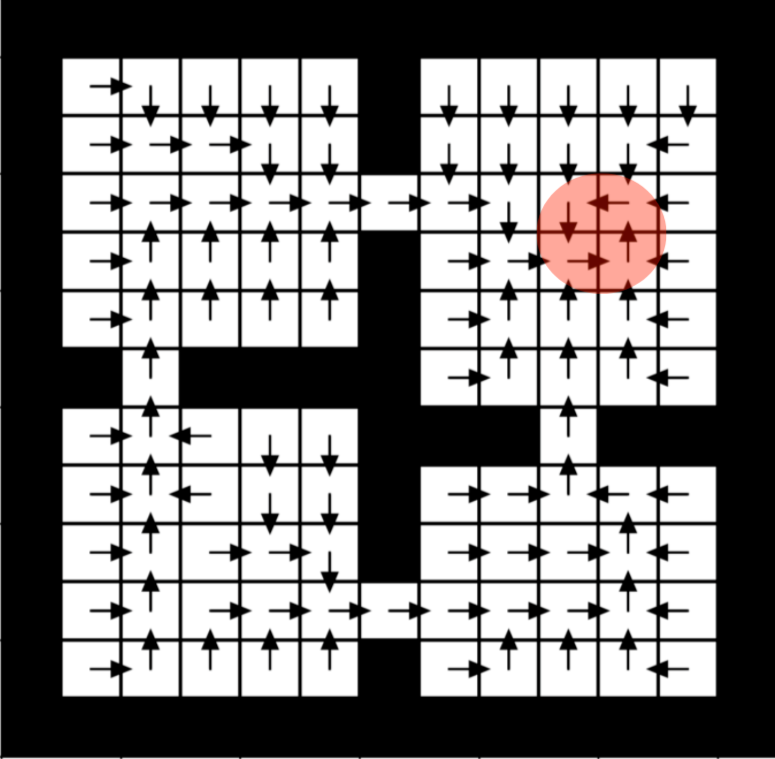}
  }
  \quad 
  \subfigure[][]{%
    \includegraphics[width=0.19\textwidth]{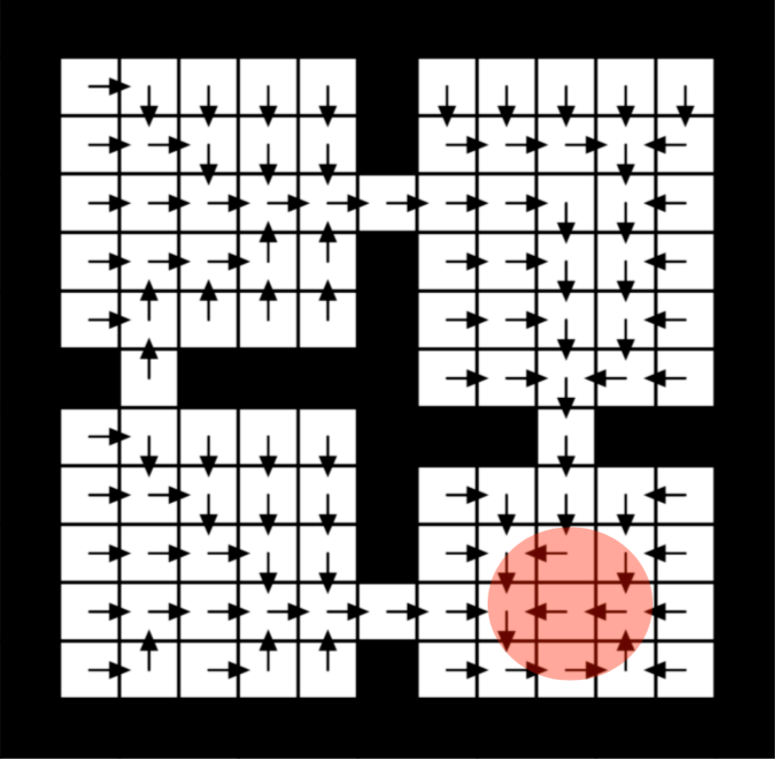}
  }
  \caption{Option visualisations on the four-room gridworld when $8$ options were discovered. The red circle approximately marks the destination/subgoal states for each discovered option-policy. Option (b, e, f, g, h) each led the agent to one of the rooms where the training tasks are concentrated. The options (a, c) were similar and seem to move the agent in the left cardinal direction. Option (d) doesn't seem to have a well-defined subgoal. We hypothesise that the options (b, e, f, g, h) were picked often compared to the options (a, c, d), during training time, which led them to have well-defined subgoals.} \label{fig:four_room_options_8_options}
\end{figure*}

For the main gridworld results in text, we discovered $4$ options ($K=4$, where $K$ is the number of options; a hyperparameter) and visualised them. This additionally brings a question which is what do these options look like when this hyperparameter is set to a different value. In this subsection, we provide visualisations for the options discovered when MODAC is trained with $K=2$ (see Fig.~\ref{fig:four_room_options_2_options}) and $K=8$ (see Fig.~\ref{fig:four_room_options_8_options}).

\subsection{Additional Qualitative Visualisations from DeepMind Lab}
\begin{figure*}[h!] 
\centering
  \subfigure[][]{%
    \includegraphics[width=0.15\textwidth]{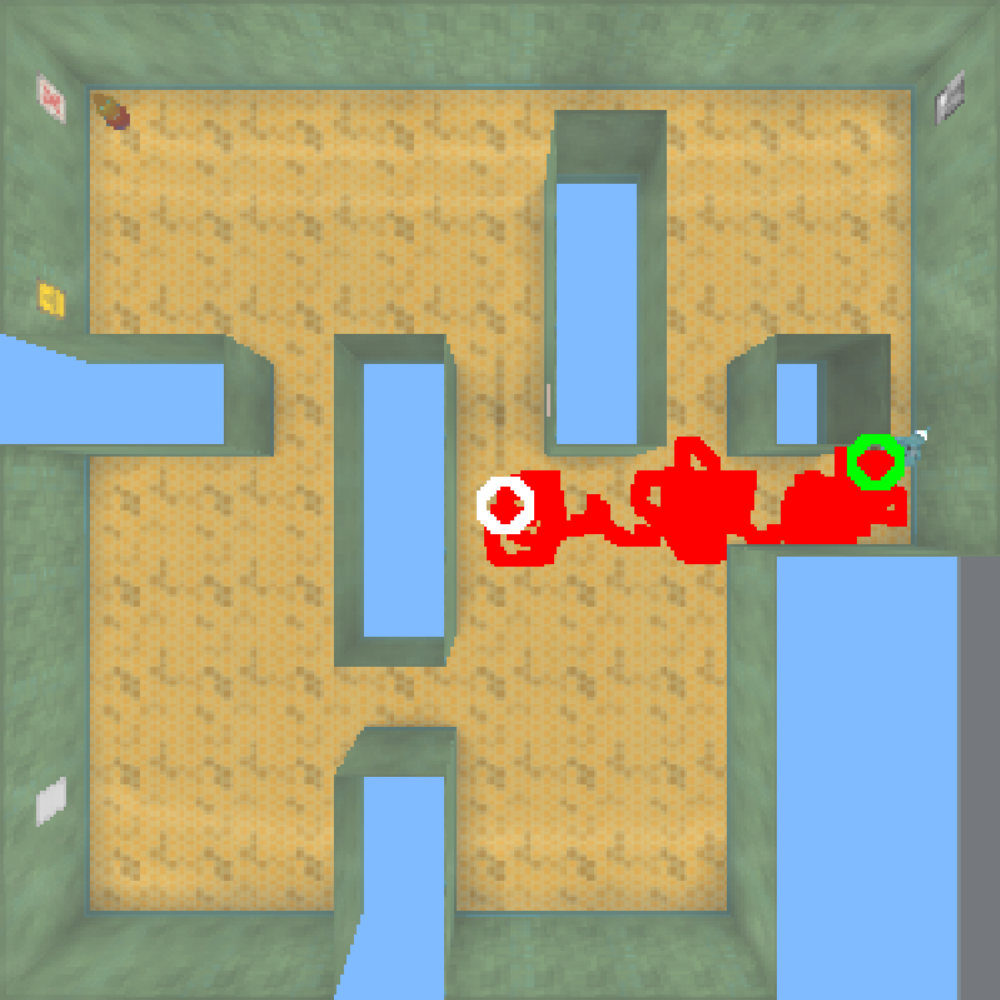}
  } 
  \quad 
  \subfigure[][]{%
    \includegraphics[width=0.15\textwidth]{figures/explore_goal_locations_small_vis/option_1.pdf}
  }
  \quad 
  \subfigure[][]{%
    \includegraphics[width=0.15\textwidth]{figures/explore_goal_locations_small_vis/option_2.pdf}
  }
  \quad 
  \subfigure[][]{%
    \includegraphics[width=0.15\textwidth]{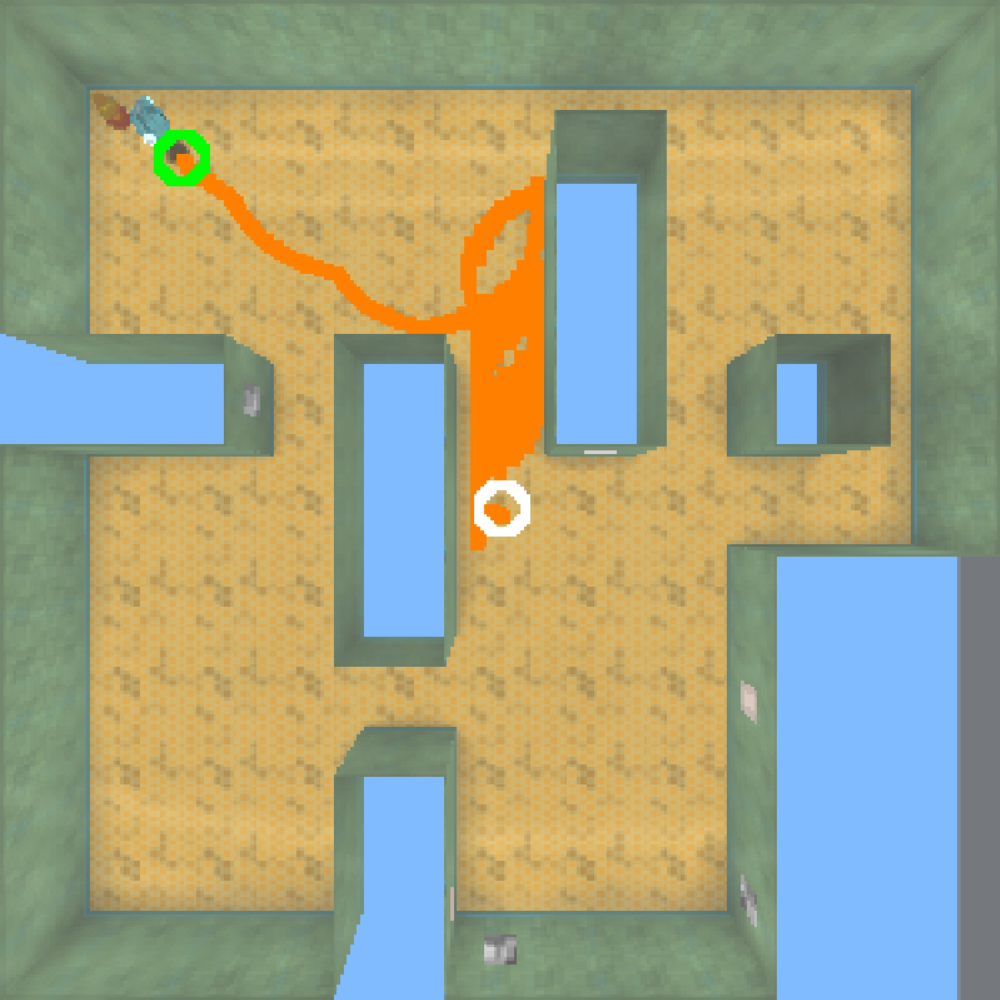}
  }
  \quad 
  \subfigure[][]{%
    \includegraphics[width=0.15\textwidth]{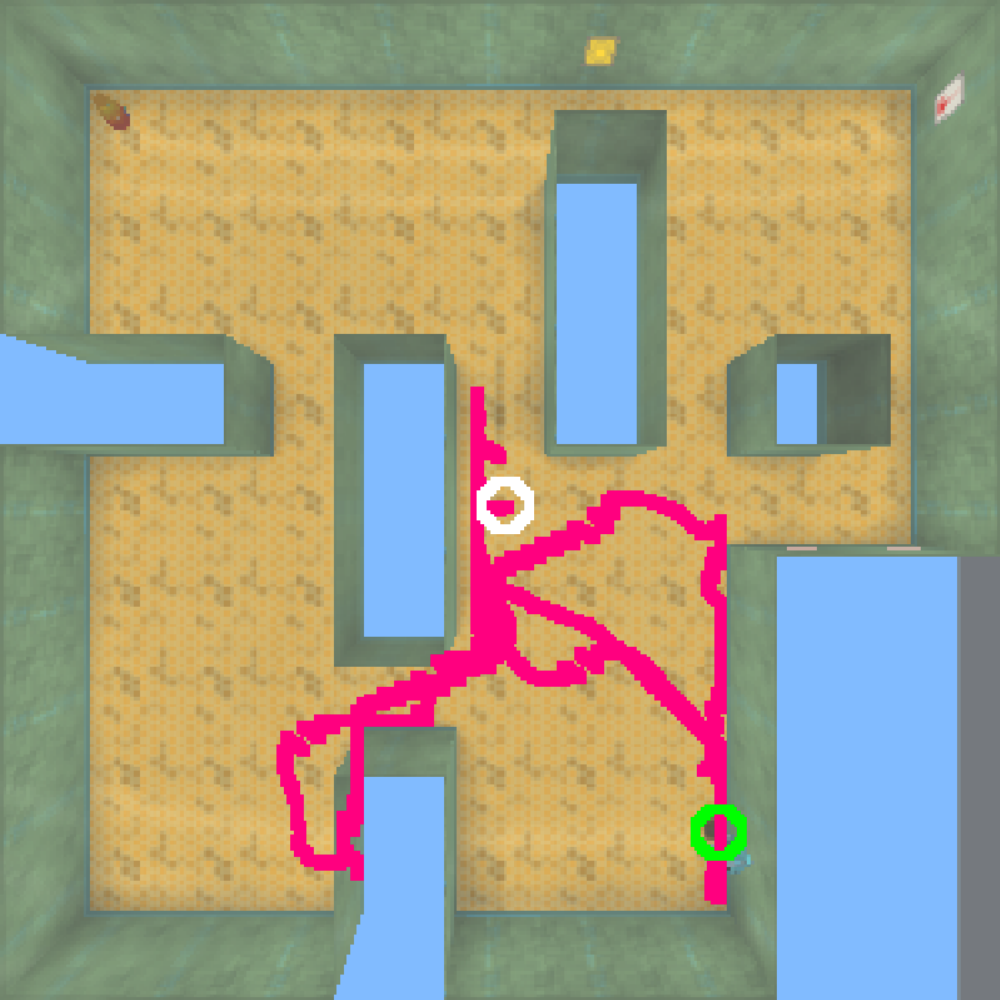}
  }
  \caption{Sampled Option-Policies on a Training Task. The goal was always in the top-left corner but not in the line of sight for the agent at the start of the episode. The agent's starting and end positions are highlighted by white and green circles. Each figure shows a trajectory by following each of the $5$ discovered options.} \label{fig:dmlab_visualise_options_training}
\end{figure*}

\begin{figure*}[h!] 
\centering
  \subfigure[][]{%
    \includegraphics[width=0.15\textwidth]{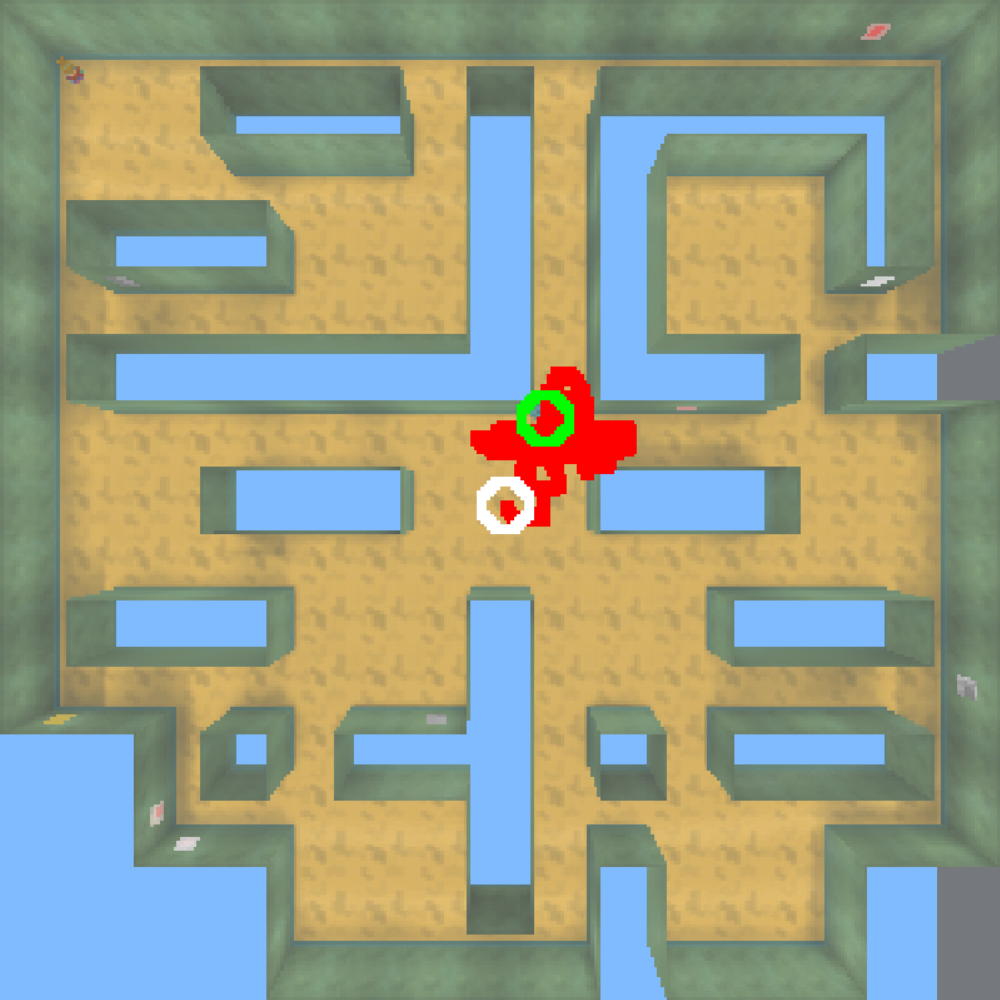}
  } 
  \quad 
  \subfigure[][]{%
    \includegraphics[width=0.15\textwidth]{figures/explore_goal_locations_large_vis/option_1.pdf}
  }
  \quad 
  \subfigure[][]{%
    \includegraphics[width=0.15\textwidth]{figures/explore_goal_locations_large_vis/option_2_updated.pdf}
  }
  \quad 
  \subfigure[][]{%
    \includegraphics[width=0.15\textwidth]{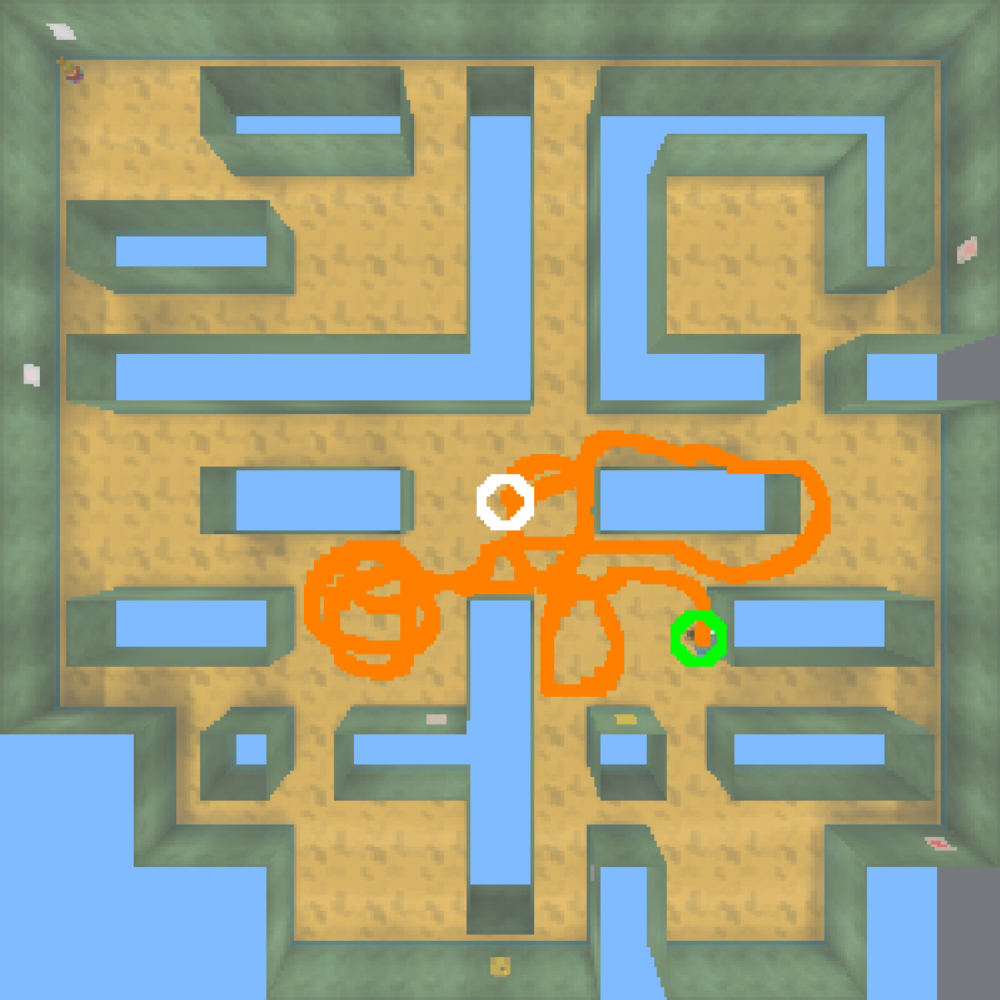}
  }
  \quad 
  \subfigure[][]{%
    \includegraphics[width=0.15\textwidth]{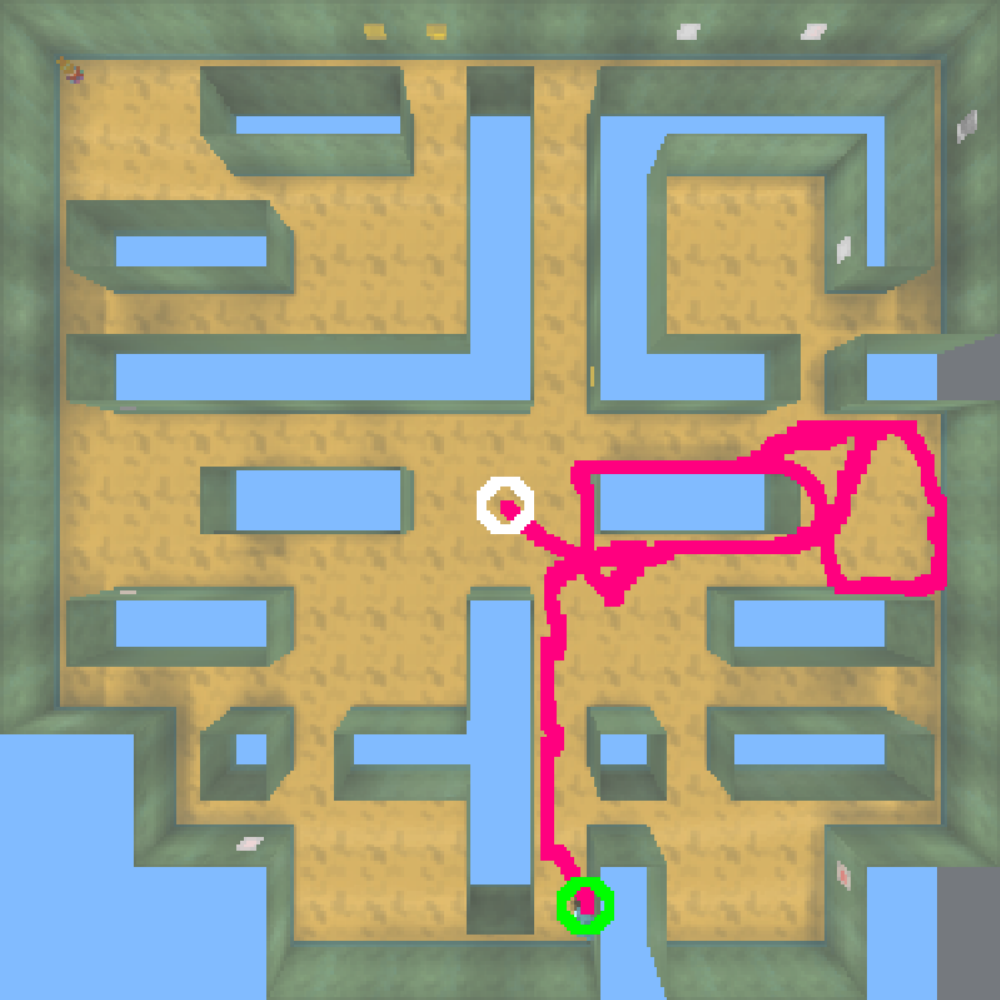}
  }
  \caption{Sampled Option-Policies on a Test Task. The goal was always in the top-left corner but not in the line of sight for the agent at the start of the episode. The agent's starting and end positions are highlighted by white and green circles. Each figure shows a trajectory by following each of the $5$ discovered options (which are obtained from the training phase).} \label{fig:dmlab_visualise_options_test}
\end{figure*}

In addition to the visualisations in the main text, we include here additional visualisations of trajectories obtained by executing all the discovered option-policies on a training task (Fig.~\ref{fig:dmlab_visualise_options_training}) and on its corresponding test task (Fig.~\ref{fig:dmlab_visualise_options_test}). From these visualisations, it can be observed that each of the option policy do produce diverse and structured exploratory behaviours in both training and test tasks. 

\begin{figure*}[h!] 
\centering
  \subfigure[][]{%
    \includegraphics[width=0.15\textwidth]{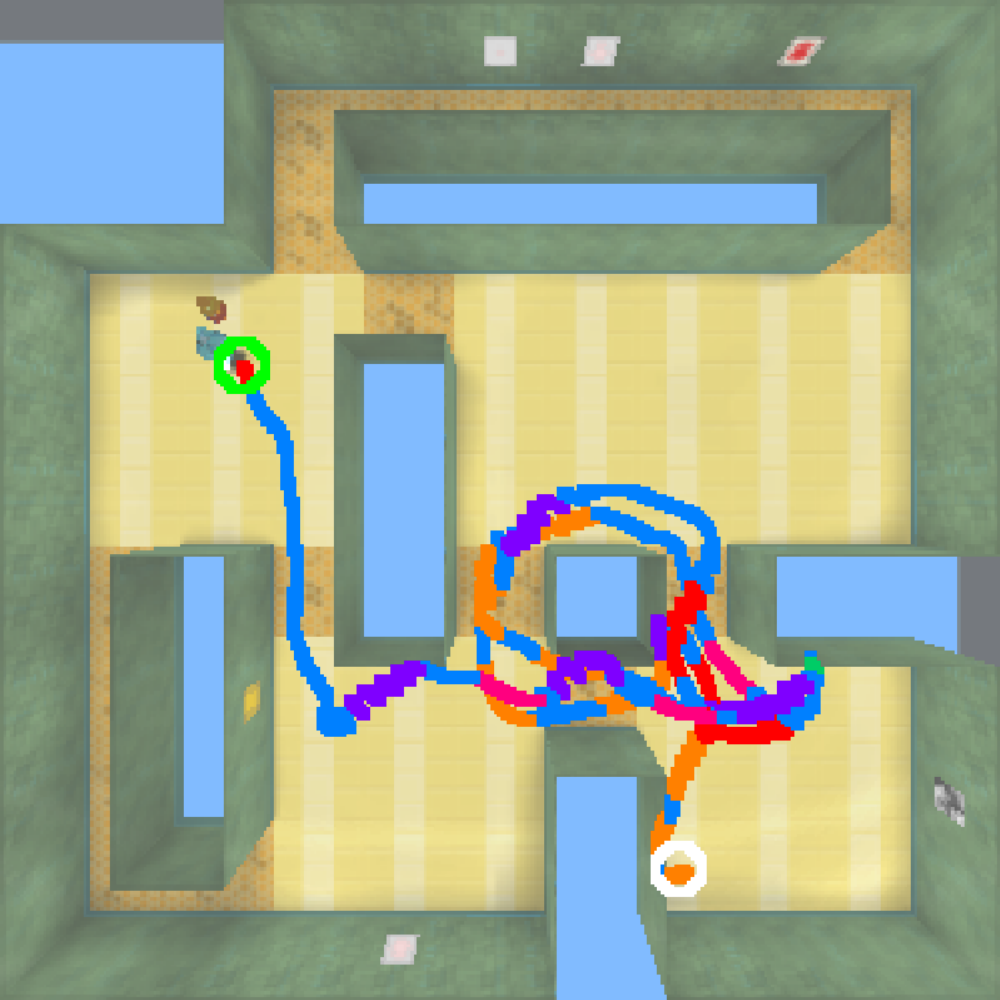}
  } 
  \quad 
  \subfigure[][]{%
    \includegraphics[width=0.15\textwidth]{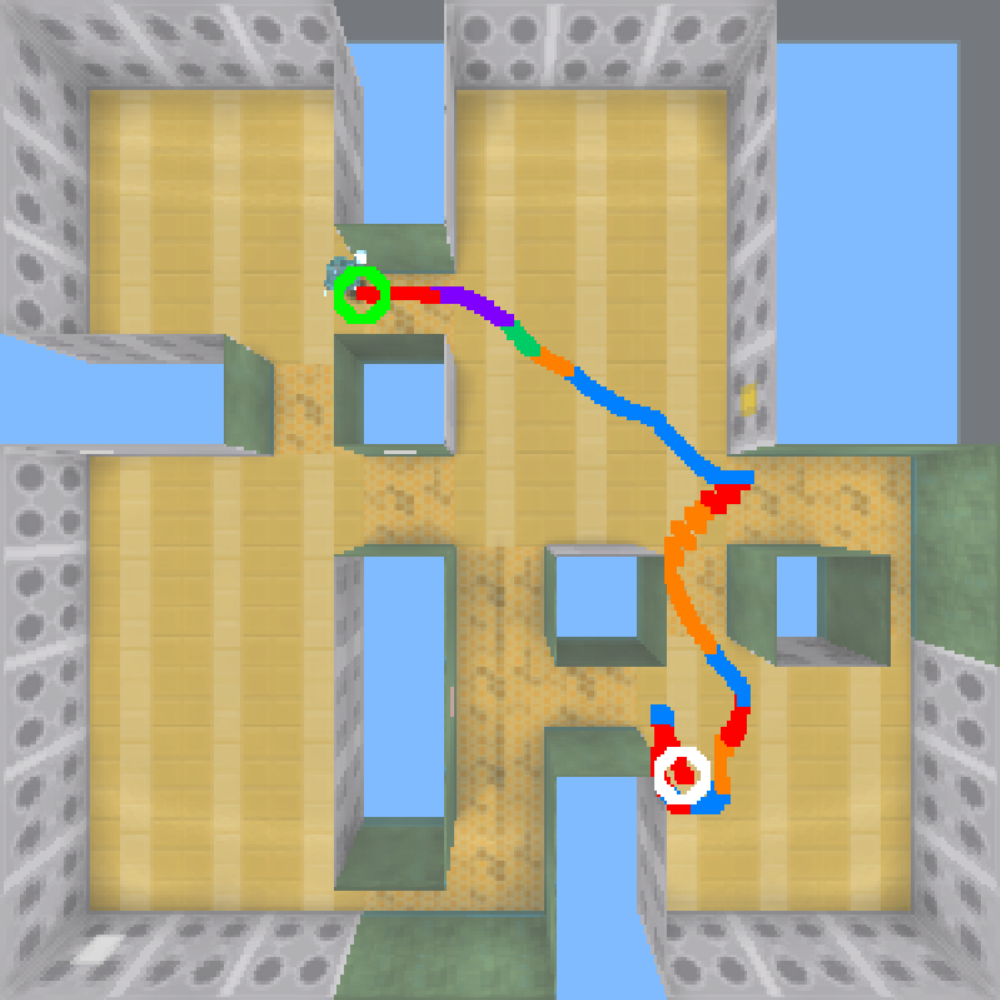}
  }
  \quad 
  \subfigure[][]{%
    \includegraphics[width=0.15\textwidth]{figures/explore_goal_locations_small_vis/examples_manager_2.pdf}
  }
  \quad 
  \subfigure[][]{%
    \includegraphics[width=0.15\textwidth]{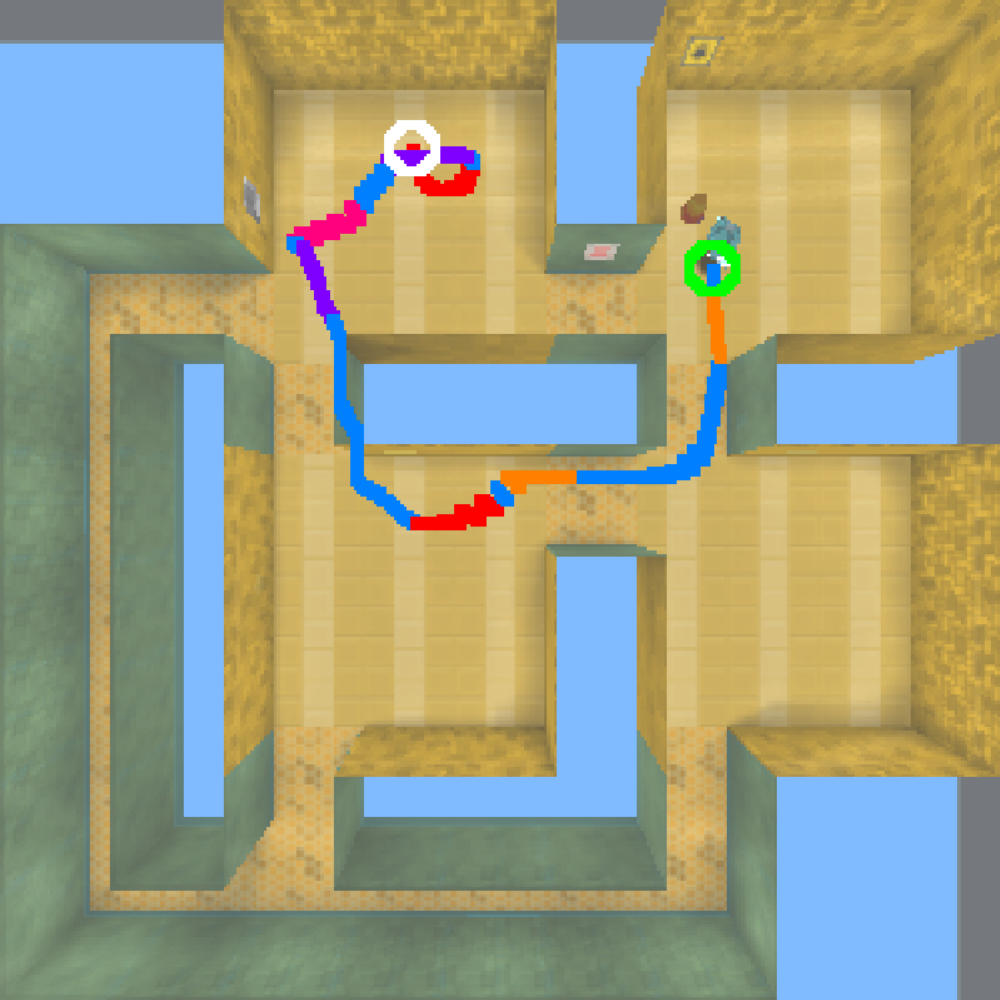}
  }
  \quad 
  \subfigure[][]{%
    \includegraphics[width=0.15\textwidth]{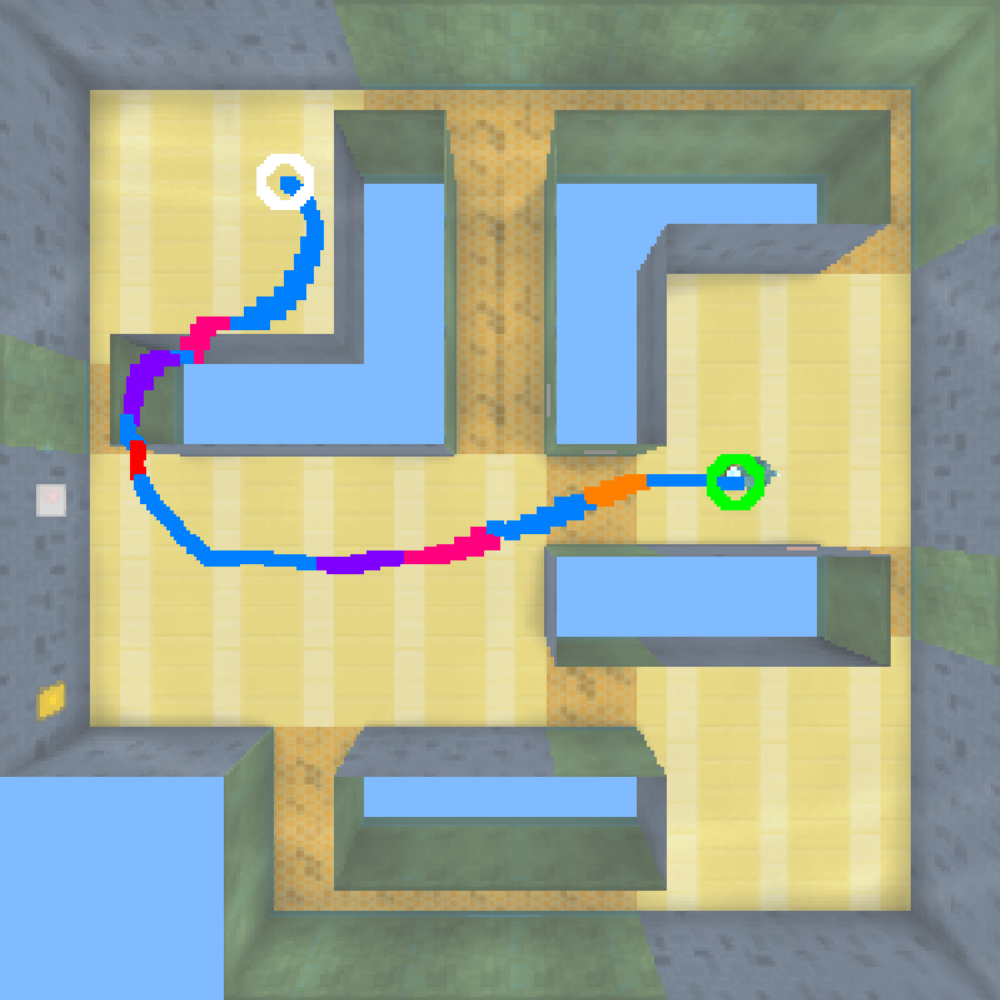}
  }
  \caption{Option Execution by a Manager on a Training Task. The figures show $5$ distinct samples of trajectories generated by a manager with access to both discovered options (marked in arbitrary colours) and primitive actions (in blue). The agent’s starting and final positions are highlighted by a white and green circles, respectively. In all cases, the agent successfully reaches the goal by using a mixture of primitive actions and discovered options.} \label{fig:dmlab_visualise_trajectory_training}
\end{figure*}

\begin{figure*}[h!] 
\centering
  \subfigure[][]{%
    \includegraphics[width=0.15\textwidth]{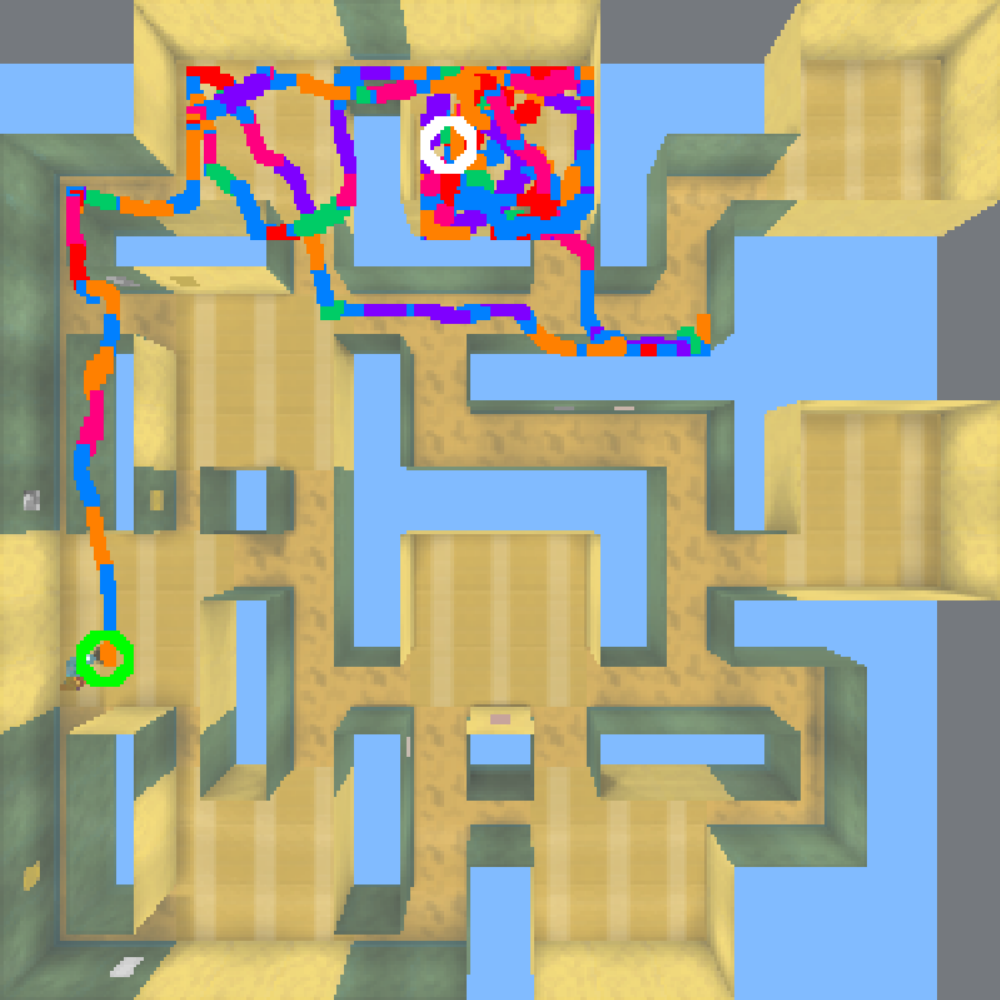}
  } 
  \quad 
  \subfigure[][]{%
    \includegraphics[width=0.15\textwidth]{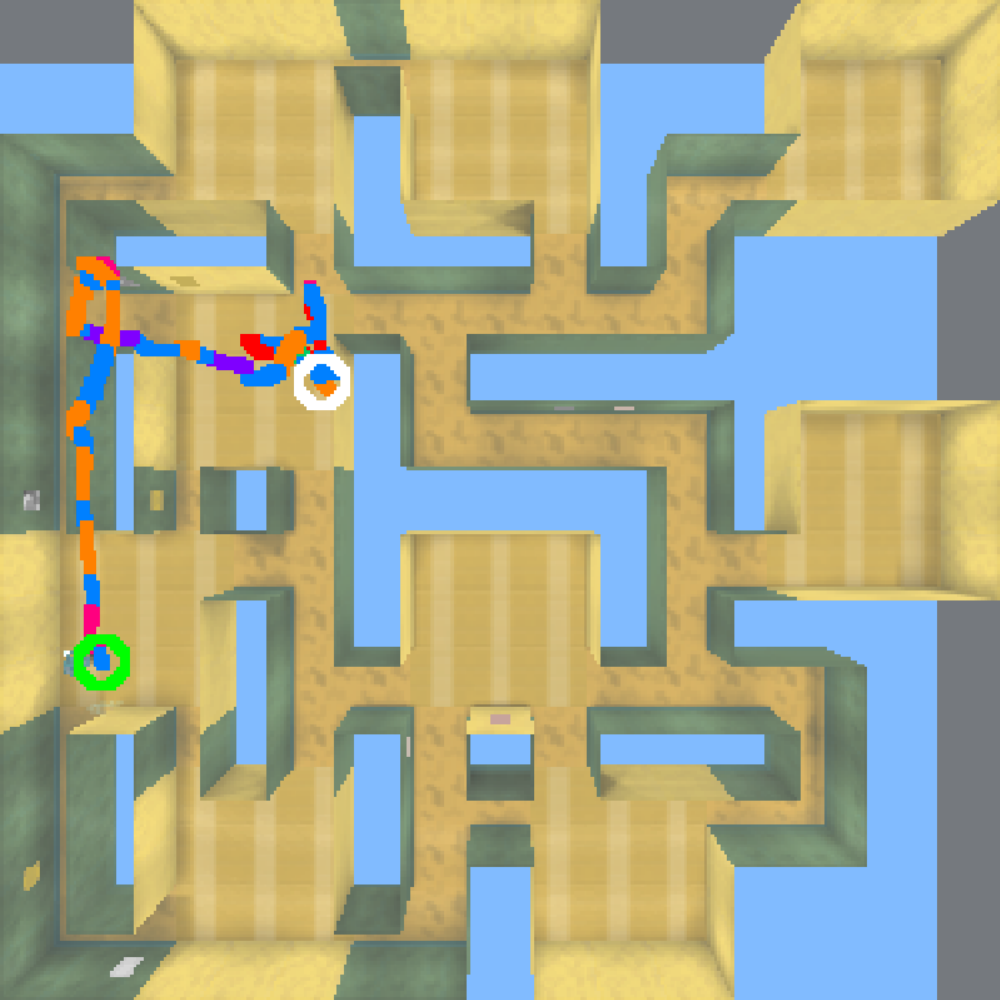}
  }
  \quad 
  \subfigure[][]{%
    \includegraphics[width=0.15\textwidth]{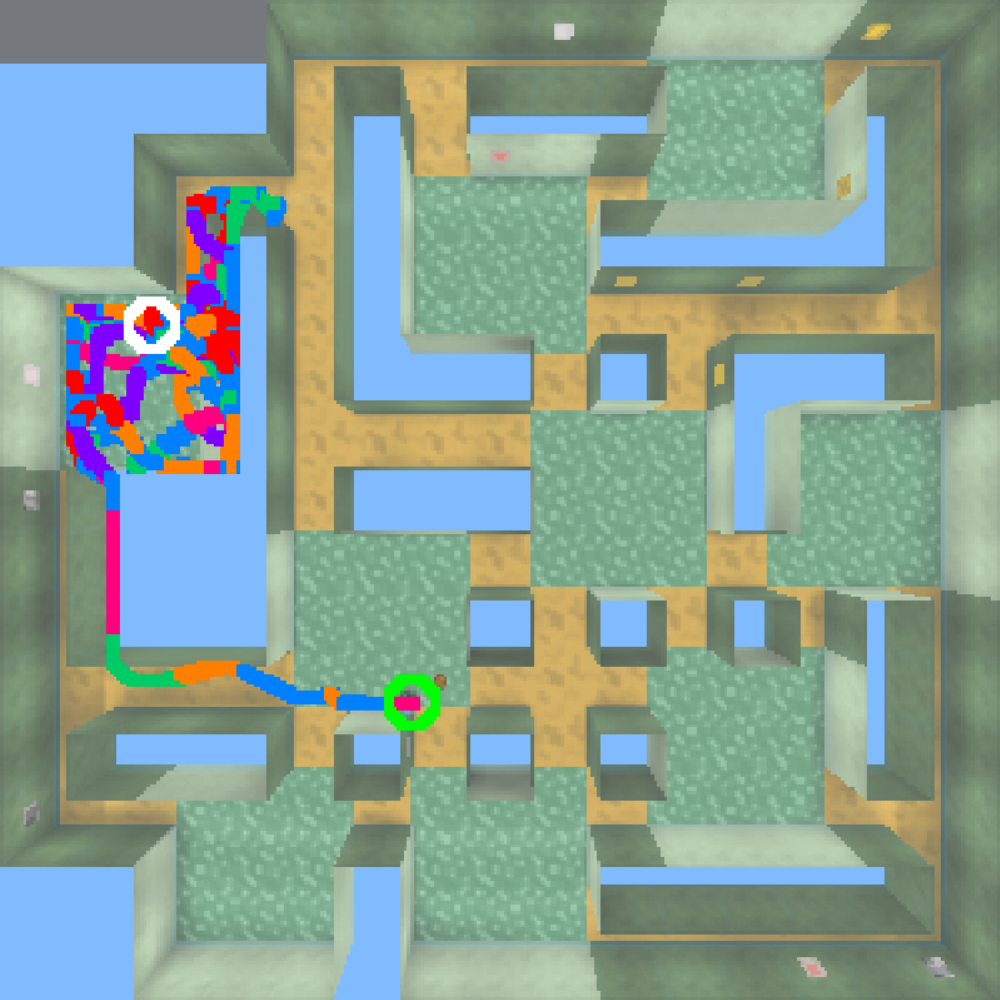}
  }
  \quad 
  \subfigure[][]{%
    \includegraphics[width=0.15\textwidth]{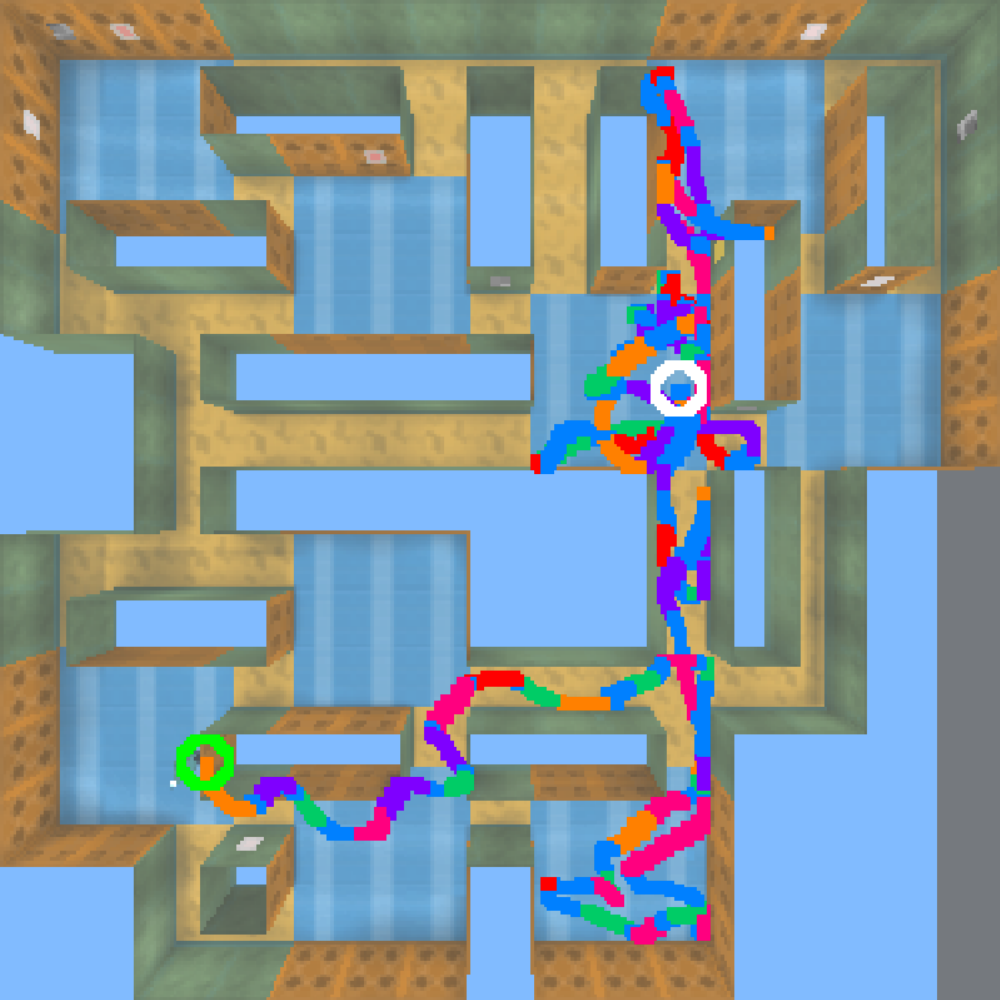}
  }
  \quad 
  \subfigure[][]{%
    \includegraphics[width=0.15\textwidth]{figures/explore_goal_locations_large_vis/examples_manager_4.pdf}
  }
  \caption{Option Execution by a Manager on a Test Task. The figures show $5$ distinct samples of trajectories generated by a manager with access to both discovered options (obtained from the training phase; marked in arbitrary colours) and primitive actions (in blue). The agent’s starting and final positions are highlighted by a white and green circles, respectively. In all cases, the agent successfully reaches the goal by using a mixture of primitive actions and discovered options.} \label{fig:dmlab_visualise_trajectory_test}
\end{figure*}

We also include visualisations of a trained MODAC agent picking options to produce behaviour in order to complete an episode on a training task (Fig.~\ref{fig:dmlab_visualise_trajectory_training}) and in its corresponding test task (Fig.~\ref{fig:dmlab_visualise_trajectory_test}). In these Figures, primitive actions are coloured in blue and options are coloured in an arbitrary, different colour. The agent's start state and end state are highlighted with white and green circles respectively. 

These figures show that the MODAC agent relies on picking options for producing behaviour in both training and test task, thus validating that our approach does indeed learn reusable and transferrable options, which is the primary reason behind their improved transfer performance.

\subsection{Additional Experiments on Atari}
Here, we study the question of whether MODAC can discover options in Atari games from unsupervised learning tasks that could become useful to maximise the game score at test time. 

Atari games, unlike DeepMind Lab tasks, have mutually inconsistent game dynamics and thus the problem of discovering options useful across distinct games would require significant new work on a separate problem, that of learning cross-game abstractions that can then support shared options. Therefore, we considered each Atari game as a separate test domain and, instead, procedurally generated multiple training tasks within each game. Specifically, we used pixel-control tasks, defined by~\citet{jaderberg2016reinforcement}, as our set of generated unsupervised training tasks. Those tasks were quite different from the test task, which was the usual task of maximising the Atari game score. Importantly, in defining the training tasks, we ignored episode terminations in the pixel-control task definition to avoid any information leaks from the test task. The challenge for MODAC was to use the generated pixel-control tasks to discover options that could speed up learning if provided to a randomly initialised manager solving the corresponding Atari game. Note that this is quite different from the typical use of pixel-control tasks, where they are used to aid learning of good state representations. In our case, the manager policy did not share any weights with the option-policy and termination networks, therefore any improvements in learning efficiency can only be attributed to the options themselves, and not to representation learning.

\begin{figure*}[h] 
\vspace{-2mm}
\centering
  \subfigure[][]{%
    \includegraphics[width=0.23\textwidth]{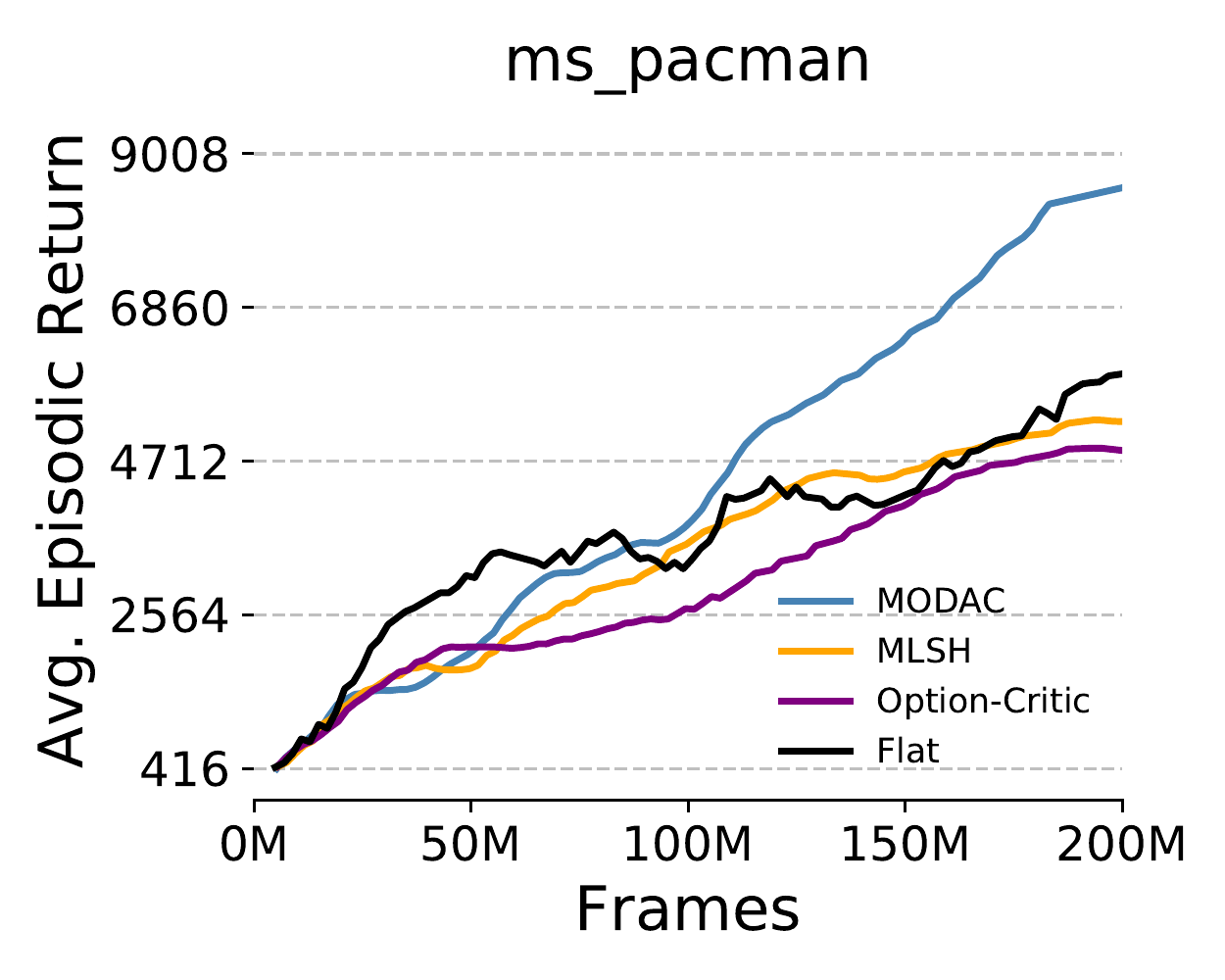}
  } 
  \
  \subfigure[][]{%
    \includegraphics[width=0.23\textwidth]{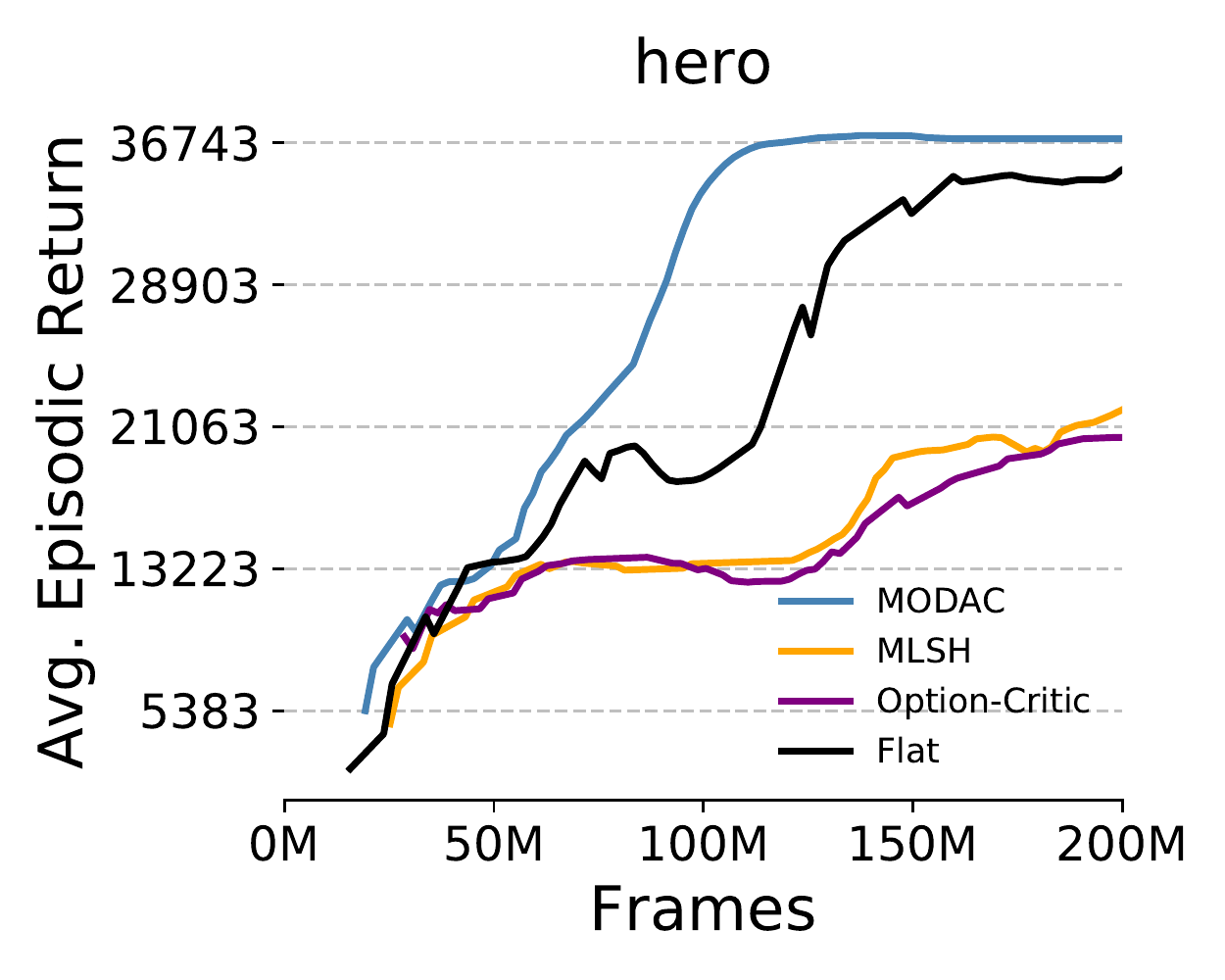}
  } 
  \ 
  \subfigure[][]{%
    \includegraphics[width=0.23\textwidth]{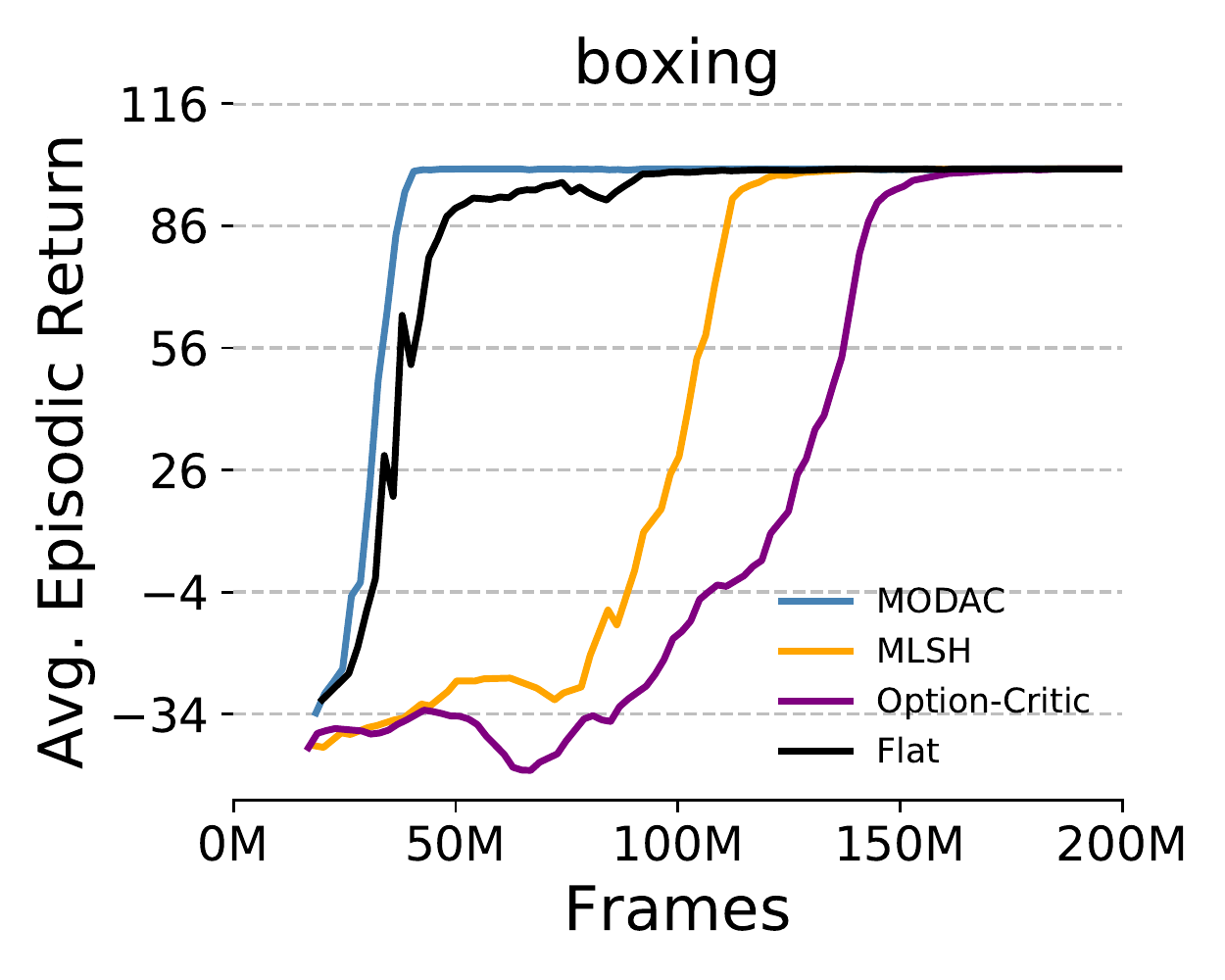}
  } 
  \ 
  \subfigure[][]{%
    \includegraphics[width=0.23\textwidth]{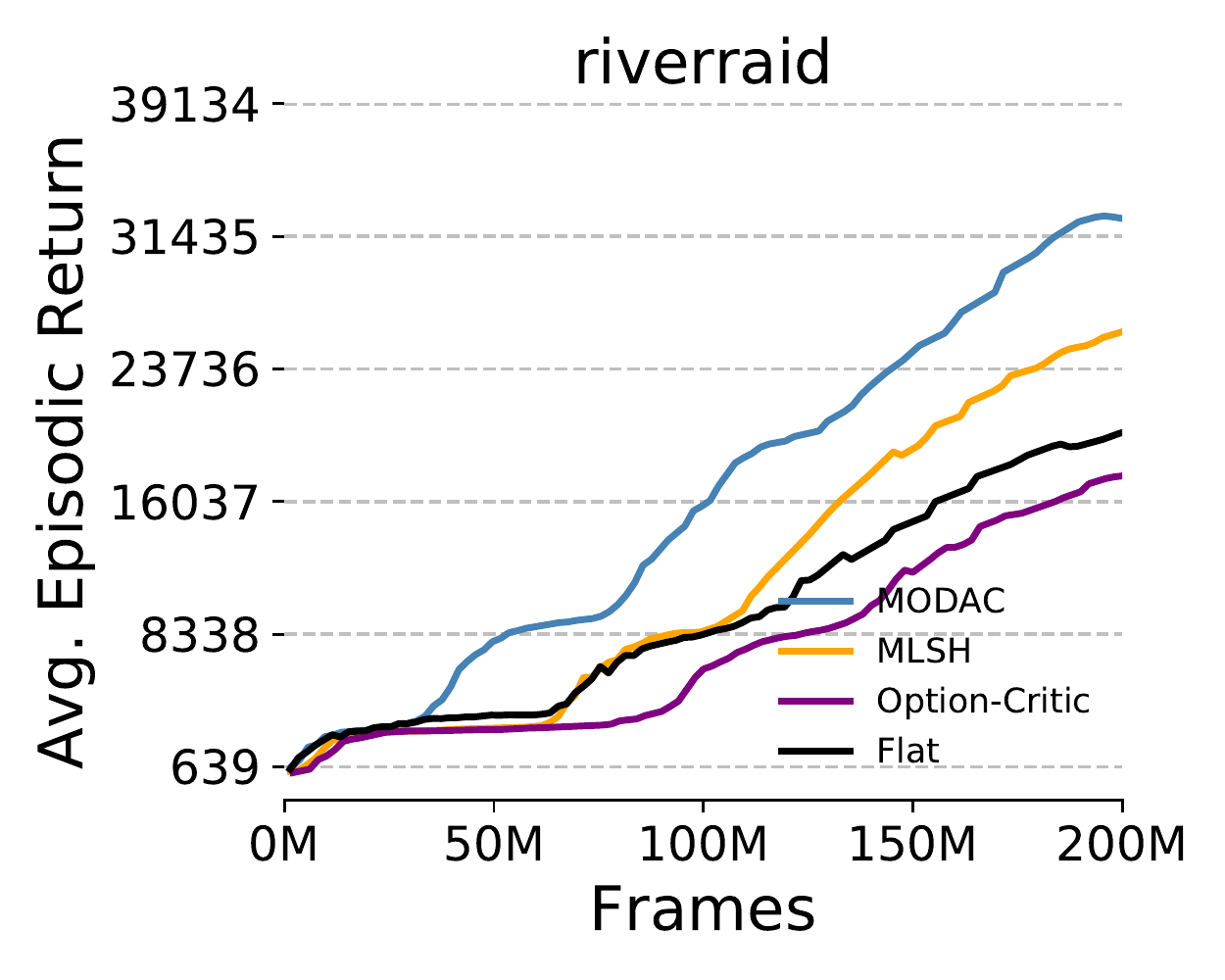}
  }

  \caption{Transfer experiments on Atari games. Figures show the performance of different agents learning to maximise rewards from the main task while having access to options discovered from pixel-control tasks defined on that same game. MODAC with discovered options learned faster in all $4$ Atari games and thereby was able to achieve better asymptotic performance on $3$ of these $4$ Atari games.} \label{fig:atari_transfer_learning} 
\end{figure*}

\noindent\textbf{Quantitative Analysis:} 
We choose $4$ Atari games (\textit{Boxing, Hero, MsPacman, Riverraid}), where pixel-control was separately found useful for representation learning. We discovered $5$ options (with a switching cost $c = 0.1$), and their average length was $7$ steps. Fig.~\ref{fig:atari_transfer_learning} shows the transfer performance of a randomly initialised manager, when given access to the pre-trained options discovered by MODAC on the pixel-control tasks defined on the corresponding Atari game. In all $4$ games, the agent that had access to the options discovered by MODAC learned to maximise game-score rewards much faster than a Flat agent learned using primitive actions alone. We measured the distribution of the manager's choices at transfer time and observed that options were selected $60.79\%$ of the time, which implies that our discovered options were responsible for agent's performance. In MsPacman, the Flat agent learned faster but saturated at a lower level, perhaps showing that the use of options during transfer can help explore better. This is consistent with the findings by~\citet{tessler2017deep} that options help initially for exploration. In all $4$ games, the transfer-learning performance of MODAC is much better than the transfer-learning performance of MLSH and Option-Critic. 

Note that during the training phase, options  were discovered from the pixel-control training tasks by ignoring the episode terminations (i.e., unsupervised). This was done deliberately in order to avoid leaking of any task-relevant information from the test task (which is to maximise the Atari game score). Here, we look at the effect of discovering options when episode terminations are not ignored. 
\begin{figure*}[h!] 
\centering
  \subfigure[][]{%
    \includegraphics[width=0.23\textwidth]{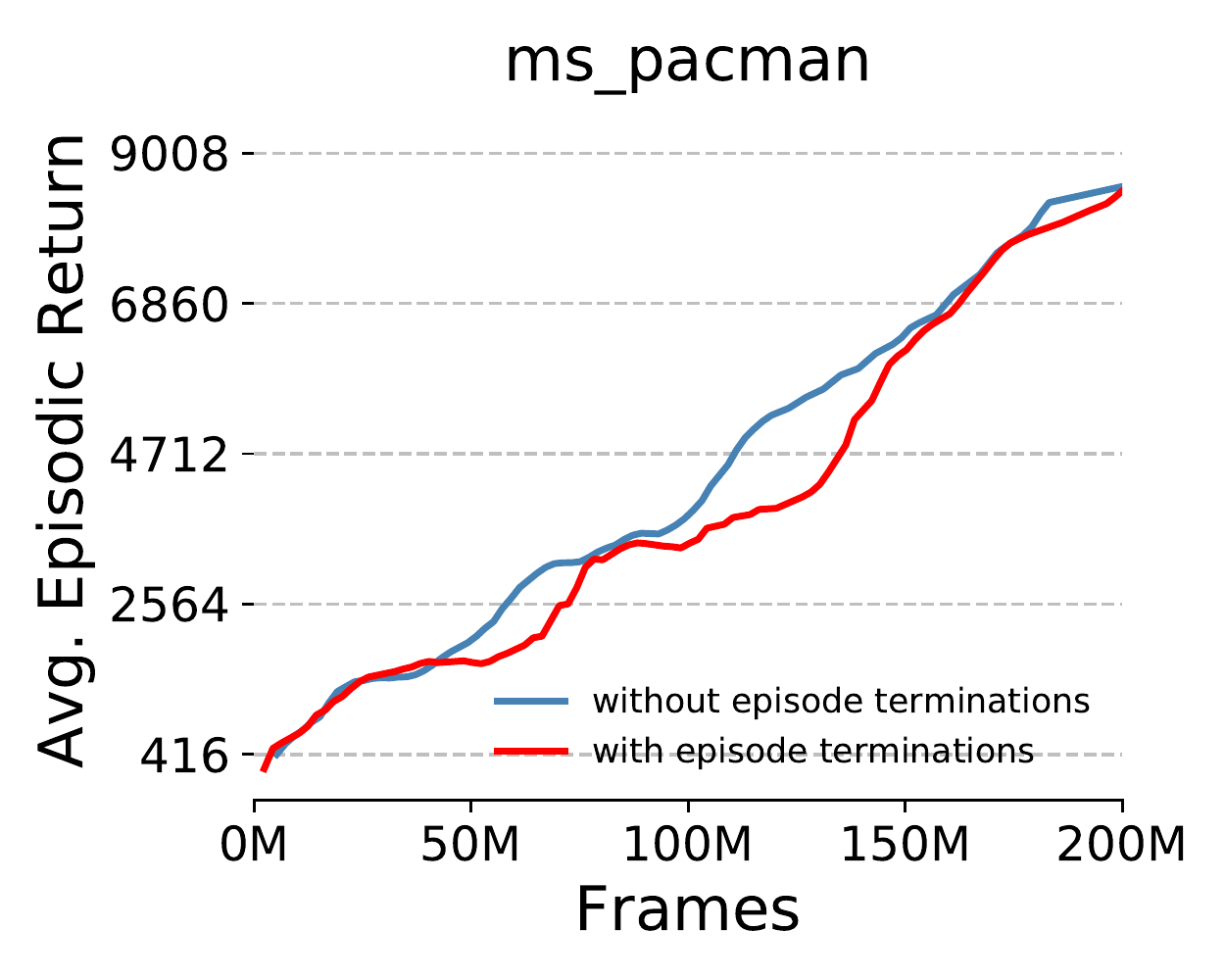}
  } 
  \
  \subfigure[][]{%
    \includegraphics[width=0.23\textwidth]{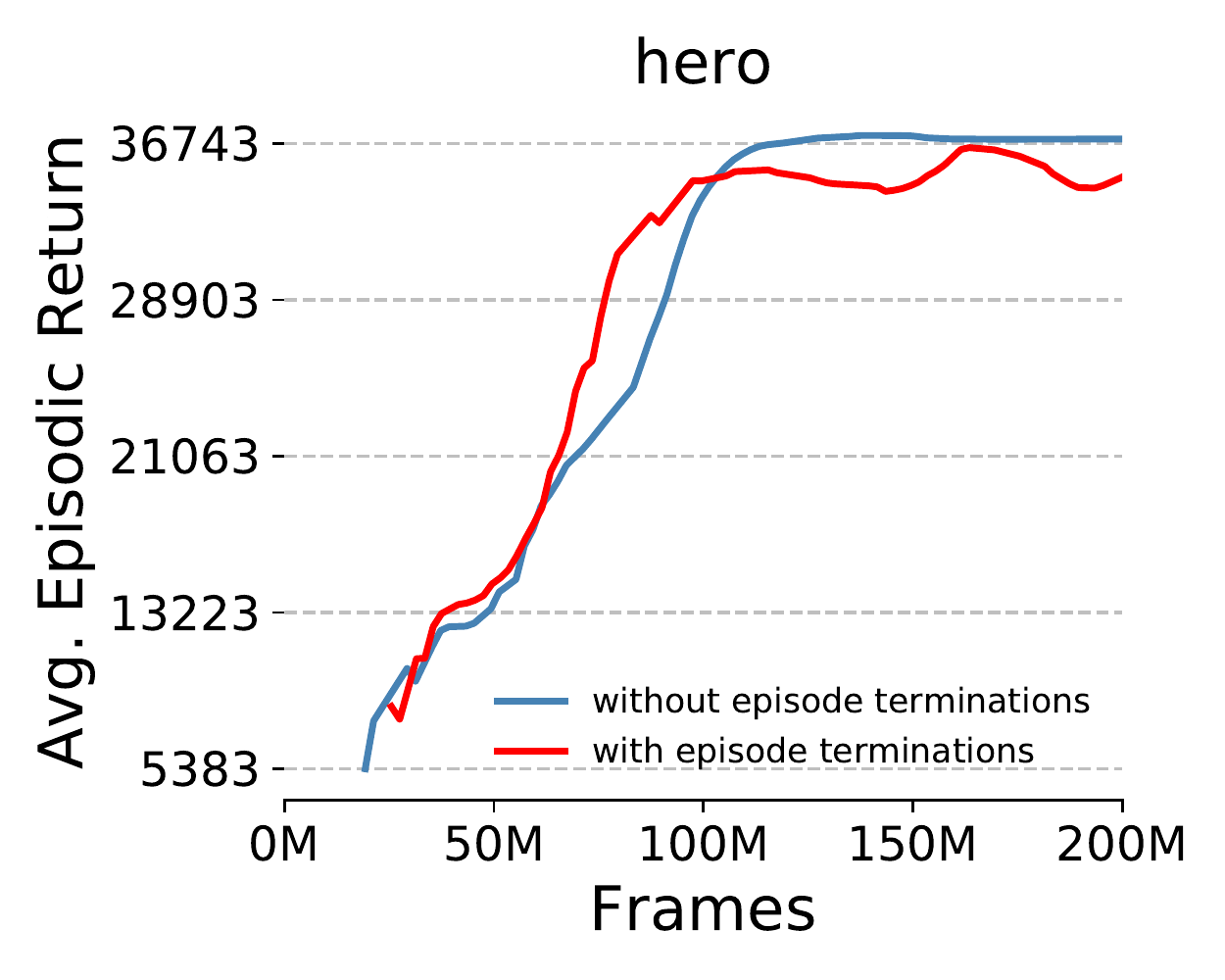}
  } 
  \ 
  \subfigure[][]{%
    \includegraphics[width=0.23\textwidth]{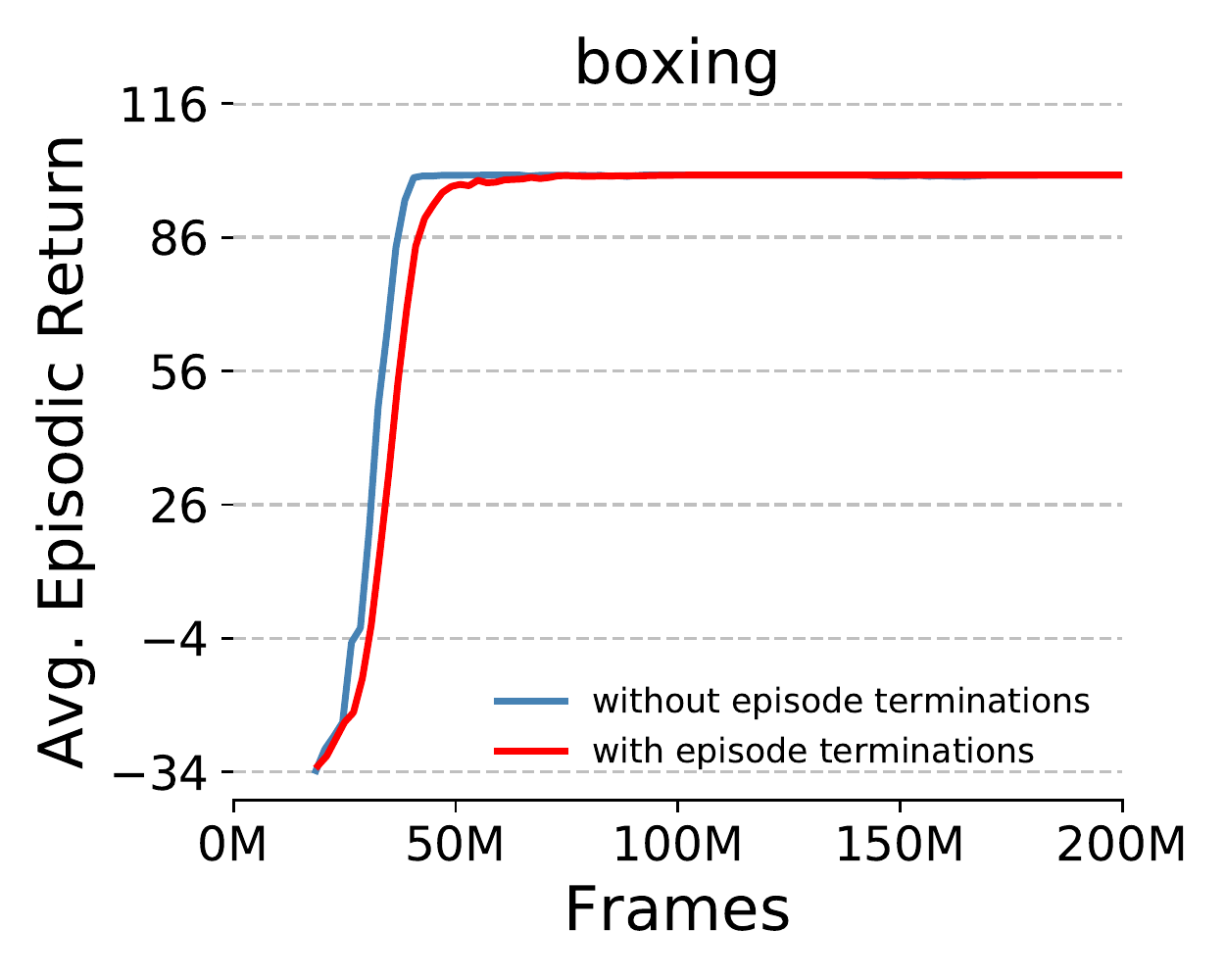}
  } 
  \ 
  \subfigure[][]{%
    \includegraphics[width=0.23\textwidth]{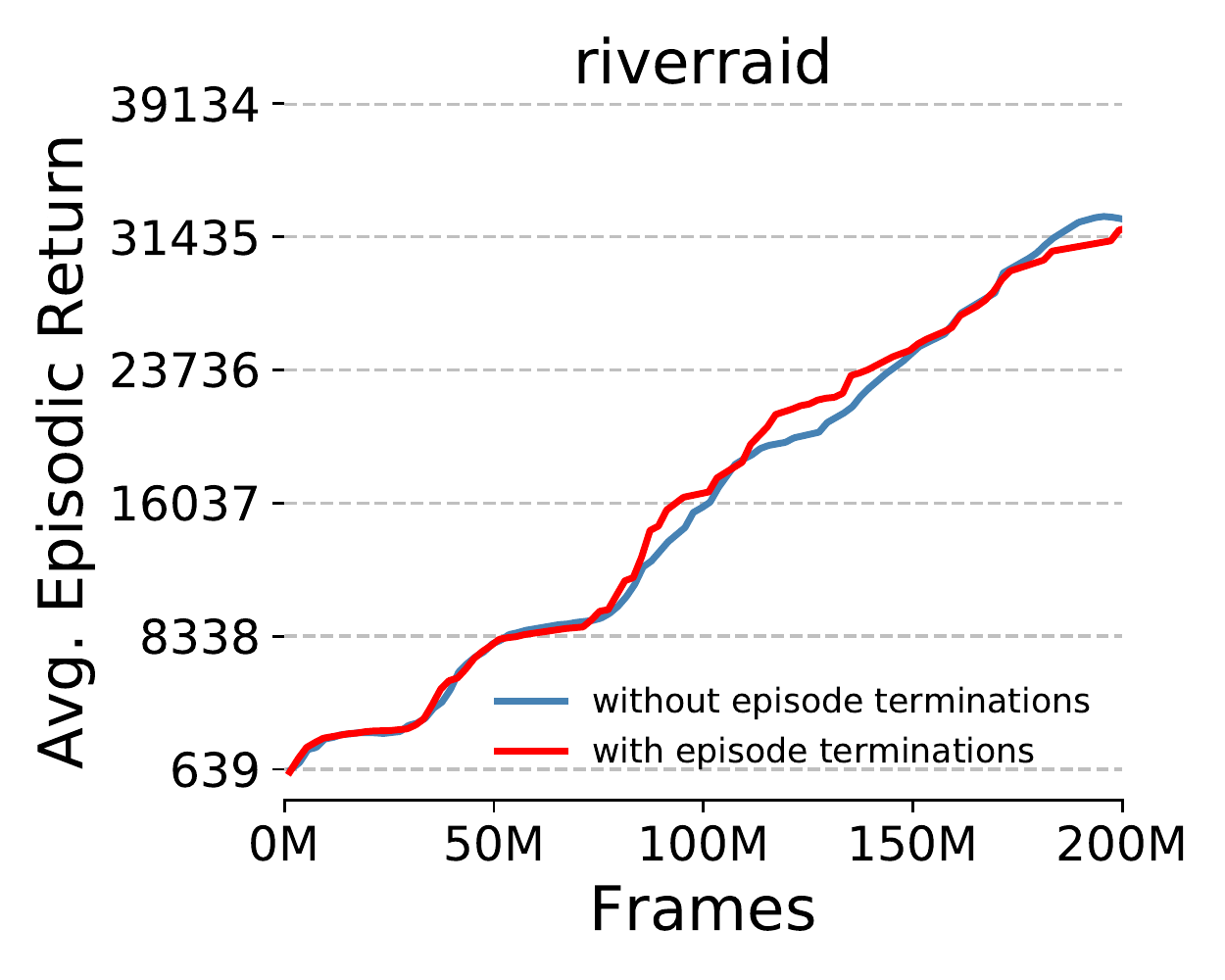}
  }
  \caption{Transfer experiments on Atari games. Figures show the performance of MODAC agents learning to maximise rewards from the main task while having access to options discovered from pixel-control tasks defined on the same game. The figure compares MODAC agents whose options were discovered by ignoring (in blue) and including (in red) the episode termination.} \label{fig:atari_ablation} 
\end{figure*}

In Figure~\ref{fig:atari_ablation}, we present learning curves of the MODAC agent with options discovered with and without episode terminations from the training task. From these, we can see that MODAC agent with options discovered by ignoring the episode terminations learns a little faster than its counterpart that included episode terminations in $2$ out of $4$ games (namely, {\it MsPacman, Boxing}). In {\it Hero}, ignoring the episode terminations seems to allow MODAC to achieve a stable asymptotic performance, and in {\it Riverraid}, there does not seem to be any visible difference in performance. 

\subsection{Objective and Update Equations for Discovering Option-Rewards and Terminations Using Meta-gradients}
The objective of MODAC is to discover the parameters of the option-rewards $ \{ \eta^{r^{o_i}} \} $ and terminations $ \{ \eta^{\beta^{o_i}} \} $ so as to maximise the hierarchical agent's performance $ G_t^M $ through the parameters of the option-policies $ \{ \theta^{o_i} \} $ that they induce. One way of accomplishing this objective is by measuring a change in the option-rewards and terminations on the hierarchical agent's performance through the change they induce in the option-policies. 



Recall that the parameters of option-policies are learned to maximise their corresponding option-rewards, with discounts applied with corresponding option-terminations, on the local trajectories that they produced; while the parameters of the manager's policy $ \theta^M $ are learned to maximise extrinsic rewards (see Eqns.~\ref{eqn:option_policy_inner_update},~\ref{eqn:manager_policy_inner_update} from main text).

The objective for option-rewards and option-terminations is to maximise the following:
\begin{align*}
    \max_{\{ \eta^{r^{o_i}} \}, \{ \eta^{\beta^{o_i}}\}} \mathbb{E}_{\{ \theta^{o_i} \}, \theta^M, \mathcal{G}} \big[ G_t^M \big]
\end{align*}
where the $n$-step return for the manager is defined as: $ G_t^M = \sum_{j=1}^{n}\gamma^j r_{t+j} - \gamma^n c + \gamma^{n+1}v^M(s_{t+n}) $, and $c$ is the switching cost. The expectation of the agent's performance is over the parameters of option-policies, manager's policy and set of training tasks $\mathcal{G}$.

Using the score-function estimator (similar to its use in the policy-gradient theorem~\citep{sutton2000policy}) and chain-rule, we can obtain the update equation for the option-rewards and terminations as follows: 
\begin{align*}
    \forall i, {\eta^{r^{o_i}}}^{\prime} &= \eta^{r^{o_i}} + \alpha_{\eta} \nabla_{\eta^{r^{o_i}}} \mathbb{E}_{\{ \theta^{o_i} \}, \theta^M, \mathcal{G}} \big[ G_t^M \big]\\
    &\approx \eta^{r^{o_i}} + \alpha_{\eta} \mathbb{E}_{\{ \theta^{o_i} \}, \theta^M, \mathcal{G}} \big[ G_t^M . \nabla_{{\theta^{o_i}}^{\prime}} \log\pi^{o_i} . \nabla_{\eta^{r^{o_i}}} {\theta^{o_i}}^{\prime}\big] \\
    \forall i, {\eta^{\beta^{o_i}}}^{\prime} &= \eta^{\beta^{o_i}} + \alpha_{\eta} \nabla_{\eta^{\beta^{o_i}}} \mathbb{E}_{\{ \theta^{o_i} \}, \theta^M, \mathcal{G}} \big[ G_t^M \big]\\
    &\approx \eta^{\beta^{o_i}} + \alpha_{\eta} \mathbb{E}_{\{ \theta^{o_i} \}, \theta^M, \mathcal{G}} \big[ G_t^M . \nabla_{{\theta^{o_i}}^{\prime}} \log\pi^{o_i} . \nabla_{\eta^{\beta^{o_i}}} {\theta^{o_i}}^{\prime}\big] 
\end{align*}
where ${\theta^{o_i}}^{\prime}$ refers to the inner-loop option-policy parameters obtained by making an inner-loop update, given produced by Eqn.~\ref{eqn:option_policy_inner_update} (thus, they are differentiable w.r.to option-rewards and terminations). The update equations are a stochastic gradient update and are computed over samples obtained from the environment. They are used according to how they are described in Alg.~\ref{alg:disc_options_algorithm}, where multiple inner-loop updates are performed per outer-loop update to option-rewards an terminations. Since the gradients for option-rewards and terminations are computed through the parameters of the option-policies, we call them meta-gradients. 

\subsection{Neural Network Architecture}
The MODAC agent, which consists of a manager, option-policy, option-reward and option-termination network, uses an identical torso architecture for all of them; and details about the torso are described below.

\textit{Gridworld:} The neural net torso consisted of a $2$-layer CNN each with $32$ filters (filter size $= 2 \times 2$, with stride length $=1$). The activations from the CNN were transformed by a single fully-connected layer of size $256$.

\textit{DeepMind Lab:} We use a Deep ResNet torso identical to the one from \citet{espeholt2018impala}, with an additional LSTM layer (with $256$ hidden units) after the feed-forward torso. 

\textit{Atari:} We use a Deep ResNet torso identical to the one from \citet{espeholt2018impala}.

All layers of the neural network use a ReLU activation function in the intermediate layers. The output layers of the option-reward and option-termination function use a $arctan$ and $sigmoid$ activations respectively. 

For the training phase in the gridworld, the task information was added as an additional channel to the input image, which was given as input to the manager network. In DeepMind Lab and Atari, during training, we obtain the task information as a one-hot vector from the environment and is passed through an embedding network to produce a 128-dimensional vector. This vector is concatenated with the feed-forward produced by the Deep ResNet torso of the manager network, which is then used for subsequent computations to produce the manager's policy.

We also used identical architecture choices for the hierarchical baselines.

\subsection{Preprocessing}
For both DeepMind Lab and Atari domains, the input to the learning agent consists of $ 4 $ consecutively stacked frames where each frame is a result of repeating the previous action for $ 4 $ time-steps, greyscaling and downsampling the resulting frames to $84 \times 84$ images, and max-pooling the last $2$. These are fairly canonical preprocessing pipeline applied to DeepMind Lab and Atari environments, and additionally, rewards are clipped to the [-1, 1] range.
 
\subsection{Hyperparameters}
For both hierarchical and flat agents (MODAC, Option-Critic, MLSH, actor-critic), we tuned the following hyperparameters: entropy weight and learning rate. In the case of the hierarchical agents, we tied the entropy weights for manager and option-policies to take the same value. Similarly, we also tied the learning rates used for training the parameters of the manager and option-policies. For MODAC, we used a single learning rate for learning the parameters of option-reward and option-termination. 

The hierarchical agents also include a switching cost hyperparameter, which is separately tuned for each agent.

We tuned the hyperparameters for each agent separately and then used a single set of hyperparameters across all DeepMind Lab and Atari environments, for each agent (which is usually the norm in many Deep RL work). The hyperparameters that we found for MODAC after tuning are reported in Table~\ref{tab:modac_hyperparam}.

We considered the following set of values $ \{ 0.0001, 0.001, 0.01, 0.03 \} $ for tuning the entropy weights and correspondingly $ \{ 0.0001, 0.0003, 0.0006, 0.001, 0.003 \} $ for the learning rates. For switching cost, we searched over $ \{ 0, 0.0001, 0.001, 0.01, 0.03, 0.05, 0.07, 0.1 \} $. Furthermore, we used RMSProp as the optimiser for updating the parameters of the learning agents.

\subsection{Experimental Setup for DeepMind Lab}
For our experiments on DeepMind Lab, we evaluated our approach on $4$ different task sets, where each set corresponds to a different navigation problem. Each set consists of a training task and a test task; In all our task sets, the training task is simpler to learn for an actor-critic agent when compared to the test task. Below, we provide the names of the tasks from the four task sets and these are taken from the suite of DeepMind Lab tasks~\citep{Beattie2016}.
\begin{center}
 \begin{tabular}{c | l l} 
 \hline
 Set No. & Training Task & Test Task \\ 
 \hline
 1 & \textit{explore\_goal\_locations\_small} &  \textit{explore\_goal\_locations\_large} \\ 
 2 & \textit{explore\_object\_rewards\_few} & \textit{explore\_object\_rewards\_many} \\
 3 & \textit{explore\_object\_locations\_small} &  \textit{explore\_object\_locations\_large} \\
 4 & \textit{explore\_obstructed\_goals\_small} & \textit{explore\_obstructed\_goals\_large} \\
 \hline
\end{tabular}
\end{center}
In both these the training and test tasks, the layout of the maze is procedurally generated, for every episode. Furthermore, the agent's start state is randomly initialised; the goal locations (for Set 1, 4), number of objects (for Set 2) and object locations (for Set 3) are also procedurally generated.

\subsection{Baselines}
We compare MODAC with the following three baseline agents in all our experiments. The first two of them are hierarchical agents that discover options/skills using their respective approaches, while the third is a non-hierarchical flat actor-critic agent. 

{\it Meta-Learned Shared Hierarchies (MLSH)}~\citep{frans2017meta}: The manager and option-policies are independently trained using an actor-critic update. The manager learns its policy by maximising task rewards; workers learn option-policies by maximising task rewards on the local trajectories generated whenever they were picked. The time scale of the workers is a fixed hyper-parameter. We tuned this via a search, and it is set to $5$ in gridworld experiments and to $10$ in Atari and DeepMind Lab.

{\it Multi-task extension of the Option-Critic with Deliberation Cost}~\citep{harb2018waiting}: The original Option-Critic with deliberation cost was designed for a single-task setting. It uses a manager and a set of workers, which learn their policies by optimising task rewards. The workers also learn a termination through the task-value function. We extend this to our multi-task setting, mirroring the architectural choices of our agent: the manager learns a task-conditional policy, while the workers learn task-independent policies and terminations. 

{\it Non-Hierarchical Actor-Critic (Flat):} In addition to the two hierarchical baselines described above, we also compare against a vanilla, non-hierarchical, actor-critic agent.

\subsection{Resource Usage}
The average running time for each agent on the DeepMind Lab training tasks is reported below. For the hierarchical agents, the running time that during the training phase are reported is significantly higher than that of the flat actor-critic agent, as they are simultaneously learning to solve the training tasks {\it and} discover options. In the test phase, the hierarchical agents reuse their discovered options, and as a result, their running times are similar to that of the flat actor-critic agent.
\begin{center}
 \begin{tabular}{c | c} 
 \hline
 Agent & Running Time \\ 
 \hline
 Actor-Critic & $3$ hours $10$ mins \\ 
 MLSH & $4$ hours $31$ mins \\
 Option-Critic &   $4$ hours $46$ mins \\
 MODAC & $5$ hours $56$ mins \\
 \hline
\end{tabular}
\end{center}

\subsection{Computing Infrastructure}
We run our experiments using a distributed infrastructure implemented in JAX \citep{jax2018github}. The computing infrastructure is based on an actor-learner decomposition \citep{espeholt2018impala}, where multiple actors generate experience in parallel, and this experience is channelled into a learner via a small queue. Both the actors and learners are co-located on a single machine, where the host is equipped with 56 CPU cores and connected to 8 TPU cores \citep{TPUs}. To minimise the effect of Python's Global Interpreter Lock, each actor-thread interacts with a \textit{batched environment}; this is exposed to Python as a single special environment that takes a batch of actions and returns a batch of observations, but that behind the scenes steps each environment in the batch in C++. The actor threads share 2 of the 8 TPU cores (to perform inference on the network), and send batches of fixed size trajectories of length T to a queue. The learner threads takes these batches of trajectories and splits them across the remaining 6 TPU cores for computing the parameter update (these are averaged with an all reduce across the participating cores). Updated parameters are sent to the actor TPU devices via a fast device to device channel as soon as the new parameters are available. This minimal unit can be replicates across multiple hosts, each connected to its own 56 CPU cores and 8 TPU cores, in which case the learner updates are synced and averaged across all learner cores (again via fast device to device communication).

\begin{table}
\vspace{3cm}
\centering
\renewcommand{\arraystretch}{1.3}
\begin{tabular}{ l c } 
 {\bf General Hyperparameters} & {\bf Value} \\ 
 \hline
 Number of environment steps & 200M \\
 $n$-step return & 20 \\
 Batch size & 32 \\
 Number of learners & 1 \\
 Number of parallel actors & 200 \\
 Learning rate schedule & Constant \\ \\
 
 {\bf Manager, Option-Policies} & {\bf Value} \\ 
 \hline
 Value loss coefficient & 0.5 \\
 Entropy coefficient & 0.01 \\
 Learning rate & 0.0006 (Atari), 0.0001 (DeepMind Lab) \\
 Switching cost & 0.1 (Atari), 0.03 (DeepMind Lab)\\
 Number of Options & 5 \\
 RMSProp momentum & 0.0 \\
 RMSProp decay & 0.99 \\
 RMSProp $ \epsilon $ & 0.01 \\
 Global gradient norm clip & 40 \\ \\
 
 {\bf Option-Rewards, Option-Terminations} & {\bf Value} \\ 
 \hline
 Meta-gradient norm clip & 1 \\
 Learning rate & 0.0001 \\
 RMSProp momentum & 0.0 \\
 RMSProp decay & 0.99 \\
 RMSProp $ \epsilon $ & 0.01 \\ 
 Inner update steps & 5 \\ 
 
 \end{tabular}
\vspace{25pt}
\caption{Detailed hyperparameters used by MODAC.}
\label{tab:modac_hyperparam}
\end{table}

\end{document}